\documentclass[10pt,twocolumn,letterpaper]{article}
\usepackage[pagenumbers]{cvpr} 
\usepackage{graphicx}
\usepackage{amsmath}
\usepackage{amssymb}
\usepackage{booktabs}
\usepackage{enumitem}
\usepackage{bm}                                      
\usepackage{amsmath}
\usepackage{graphicx}
\usepackage{amsfonts}
\usepackage{algorithm}
\usepackage{physics}
\usepackage{algpseudocode}
\usepackage{booktabs}
\usepackage{subcaption}
\usepackage{amssymb,amsmath,amsthm}
\usepackage{multirow}
\usepackage[english]{babel}

\usepackage[accsupp]{axessibility}
\usepackage[pagebackref,breaklinks,colorlinks]{hyperref}
\usepackage[title]{appendix}
\usepackage{multicol}
\usepackage{cuted}
\usepackage{appendix}

\usepackage{lipsum}
\usepackage{microtype}
\newcommand\blfootnote[1]{%
  \begingroup
  \renewcommand\thefootnote{}\footnote{#1}%
  \addtocounter{footnote}{-1}%
  \endgroup
}
\usepackage[accsupp]{axessibility}
\usepackage[capitalize]{cleveref}
\crefname{section}{Sec.}{Secs.}
\Crefname{section}{Section}{Sections}
\Crefname{table}{Table}{Tables}
\crefname{table}{Tab.}{Tabs.}

\begin{document}
\title{\vspace{-12pt}
Steered Diffusion: A Generalized Framework for Plug-and-Play\\ Conditional Image Synthesis \vspace{-5mm}}
\author{
Nithin Gopalakrishnan Nair$^{1*}$\quad Anoop Cherian$^{2}$\quad Suhas Lohit$^{2}$\quad Ye Wang$^{2}$\quad \\ Toshiaki Koike-Akino$^{2}$\quad  Vishal M. Patel$^{1}$\quad  Tim K. Marks$^{2}$\quad 
\\
\small{
$^{1}$~Johns Hopkins University\quad
$^{2}$~Mitsubishi Electric Research Laboratories (MERL)
}
\\
{\scriptsize \texttt{\{ngopala2,vpatel36\}@jhu.edu} \quad \texttt{\{acherian,slohit,yewang,koike,tmarks\}@merl.com}} \\
\small{\url{https://merl.com/demos/steered-diffusion}}
\vspace{-0.21cm}
}
\twocolumn[{%
\renewcommand\twocolumn[1][]{#1}%
\maketitle
\vspace{1.5cm}
\vspace{-3\baselineskip}
\vspace{-3\baselineskip}
\begin{center}
\centering
\setlength{\tabcolsep}{0.5pt}
\captionsetup{type=figure}
{\footnotesize
\renewcommand{\arraystretch}{0.5} 
\begin{tabular}{c c c c c c c c  }
 \tabularnewline
        {(a) Inpainting \hspace{120pt}(b)
         Colorization \hspace{120pt}(c)
        Super-resolution}\\
    \tabularnewline   

 \includegraphics[width=0.102\linewidth]{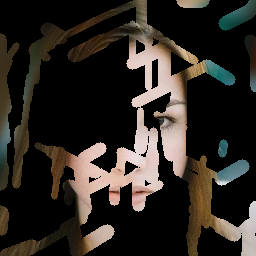}
 \hspace{0.5mm}
 \includegraphics[width=0.102\linewidth]{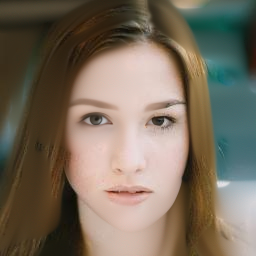}  \includegraphics[width=0.102\linewidth]{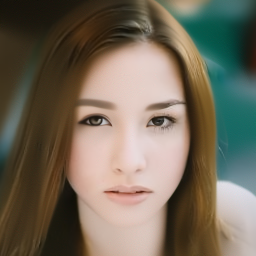}
   \hspace{2mm}
 \includegraphics[width=0.102\linewidth]{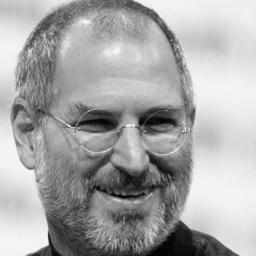} 
  \hspace{0.5mm}
 \includegraphics[width=0.102\linewidth]{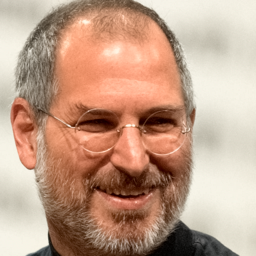}
 \includegraphics[width=0.102\linewidth]{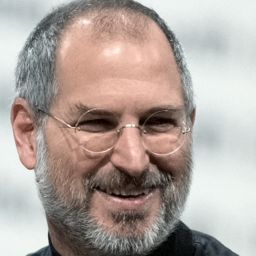}
\hspace{2mm}
 \includegraphics[width=0.102\linewidth]{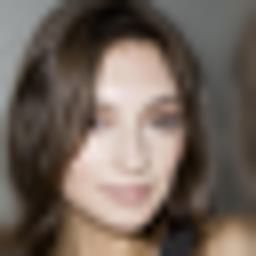}
  \hspace{0.5mm}
 \includegraphics[width=0.102\linewidth]{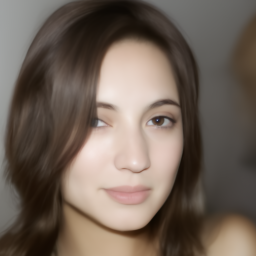}  \includegraphics[width=0.102\linewidth]{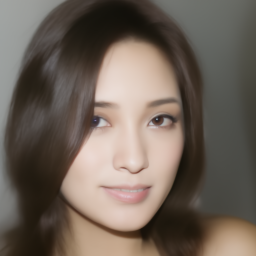}
    \tabularnewline
    \hspace{-0.6mm}
 \includegraphics[width=0.102\linewidth]{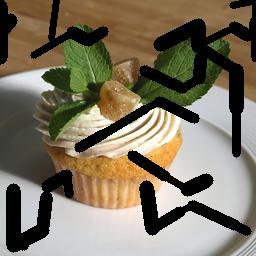}
  \hspace{0.5mm}
  \includegraphics[width=0.102\linewidth]{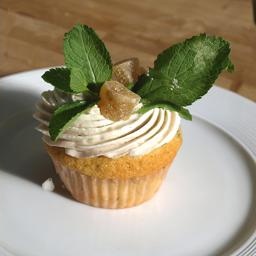}
    \includegraphics[width=0.102\linewidth]{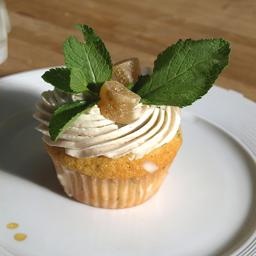}
 
  \hspace{2mm}
 \includegraphics[width=0.102\linewidth]{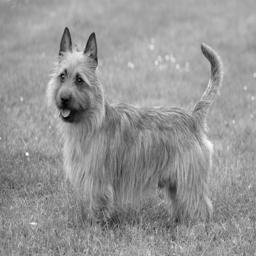}
  \hspace{0.5mm}
 \includegraphics[width=0.102\linewidth]{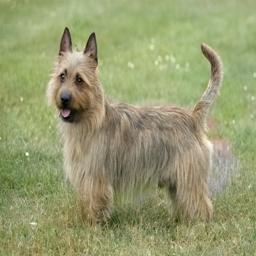}  \includegraphics[width=0.102\linewidth]{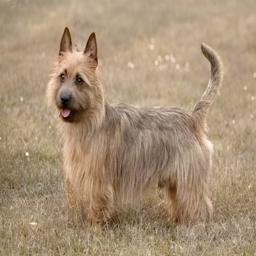}  
\hspace{2mm}
 \includegraphics[width=0.102\linewidth]{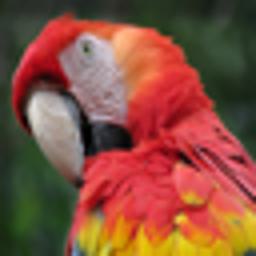}
  \hspace{0.5mm}
 \includegraphics[width=0.102\linewidth]{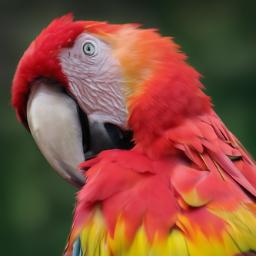}  \includegraphics[width=0.102\linewidth]{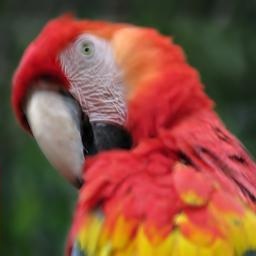}
\tabularnewline
        { Input  \hspace{35pt} Generated Samples\hspace{50pt }Input  \hspace{35pt} Generated Samples\hspace{50pt }Input  \hspace{35pt} Generated Samples
        }\\

\vspace{1mm}
\tabularnewline
             {(d) Semantic Generation \hspace{90pt}(e)
        Identity Replication \hspace{90pt}(f)
        Text-based editing}\\
    \tabularnewline   
 \includegraphics[width=0.102\linewidth]{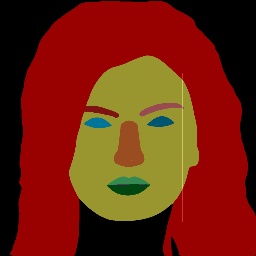} 
 \hspace{0.5mm}
  \includegraphics[width=0.102\linewidth]{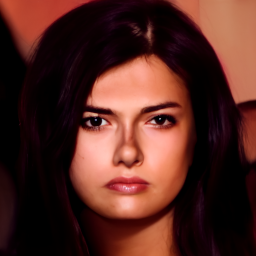}  \includegraphics[width=0.102\linewidth]{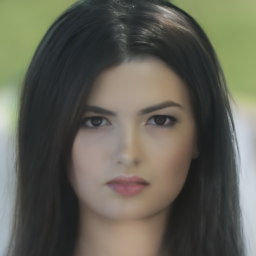}
\hspace{2mm}
  \includegraphics[width=0.102\linewidth]{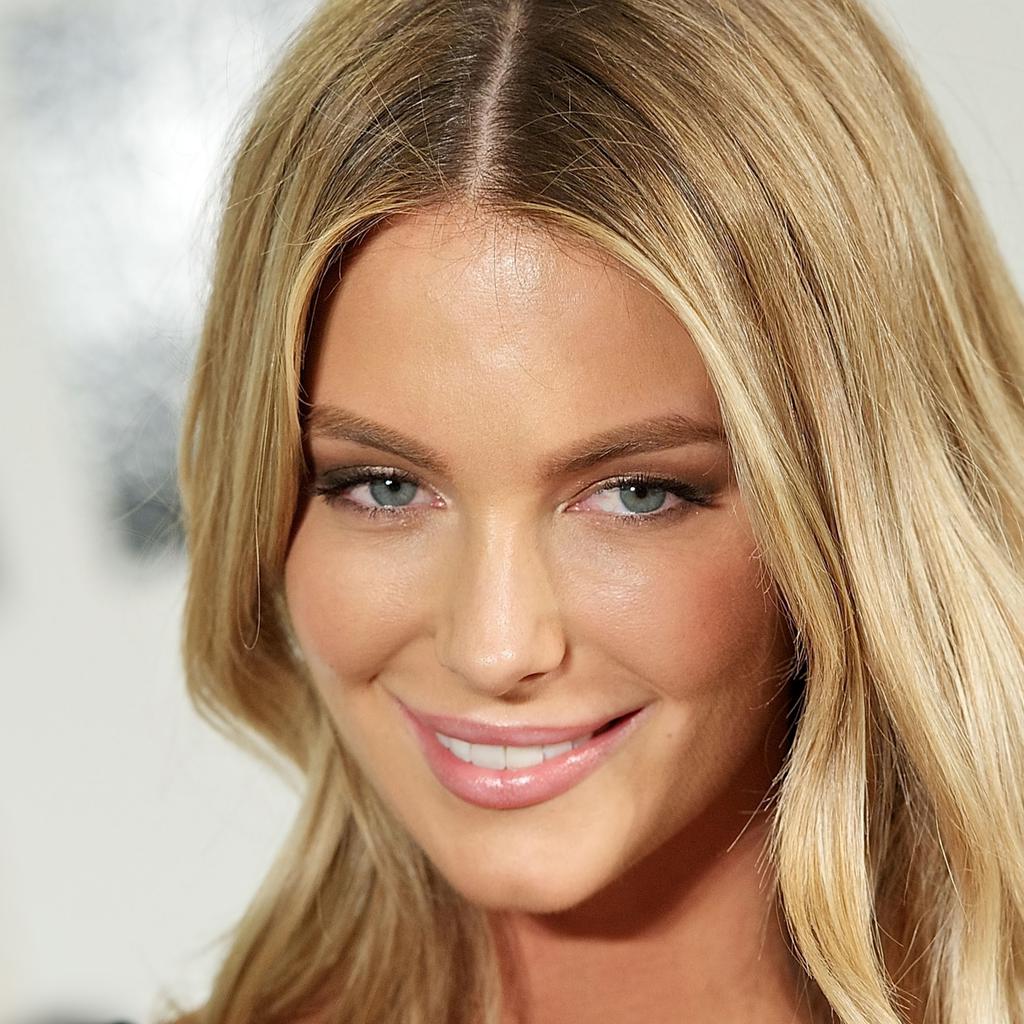}
  \hspace{0.5mm}
 \includegraphics[width=0.102\linewidth]{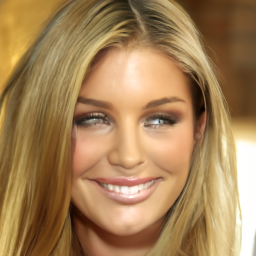}  \includegraphics[width=0.102\linewidth]{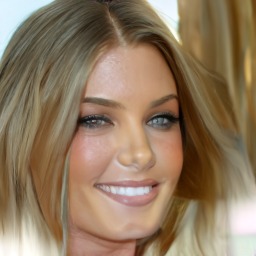} 
 \hspace{2mm}
\includegraphics[width=0.102\linewidth]{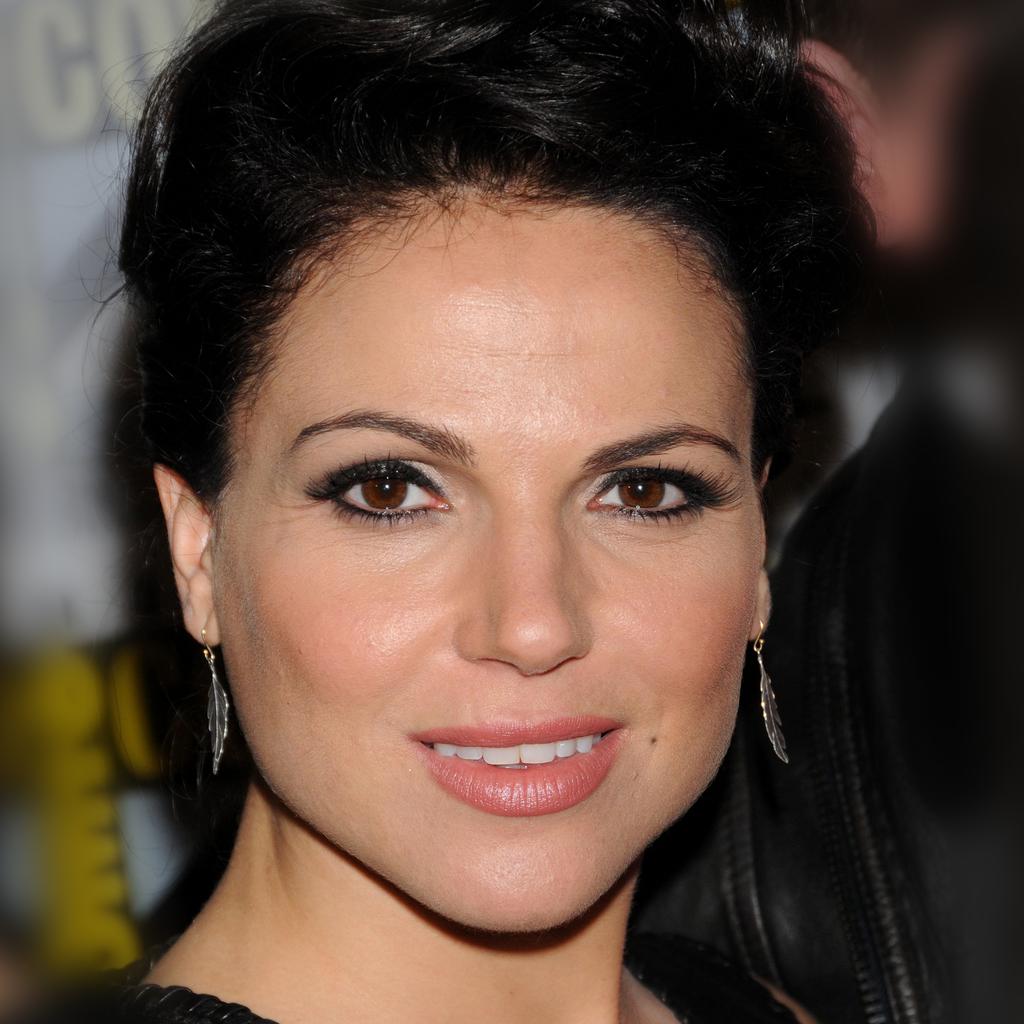}
  \hspace{0.5mm}
 \includegraphics[width=0.102\linewidth]{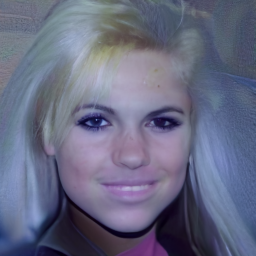}  \includegraphics[width=0.102\linewidth]{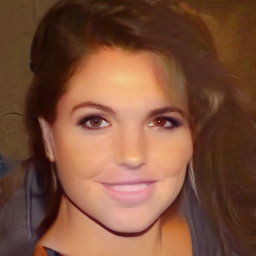} 
    \tabularnewline
\hspace{-0.8mm}
 \includegraphics[width=0.102\linewidth]{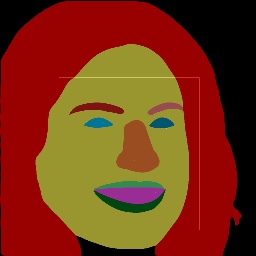}
  \hspace{0.5mm}
 \includegraphics[width=0.102\linewidth]{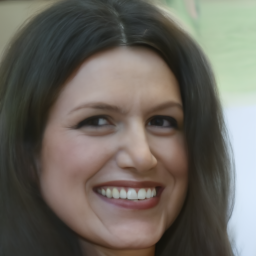}  \includegraphics[width=0.102\linewidth]{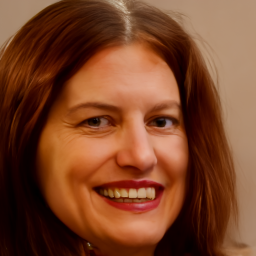}

\hspace{2mm}
   \includegraphics[width=0.102\linewidth]{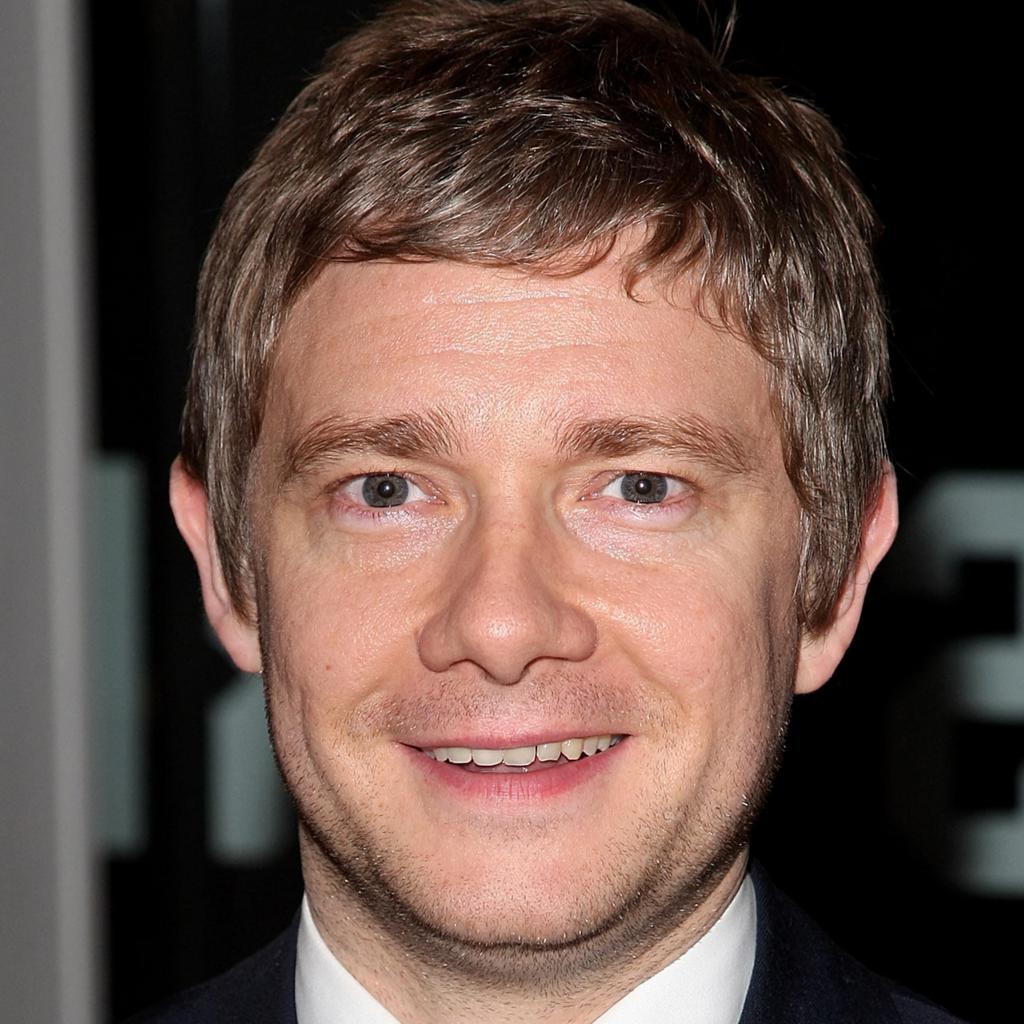}
  \hspace{0.5mm}
  \includegraphics[width=0.102\linewidth]{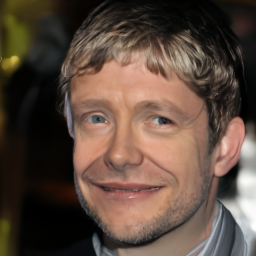}  \includegraphics[width=0.102\linewidth]{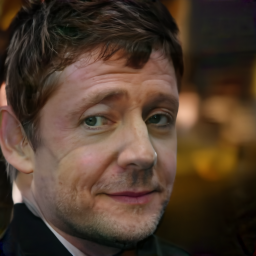}  
 
\hspace{2mm}
 \includegraphics[width=0.102\linewidth]{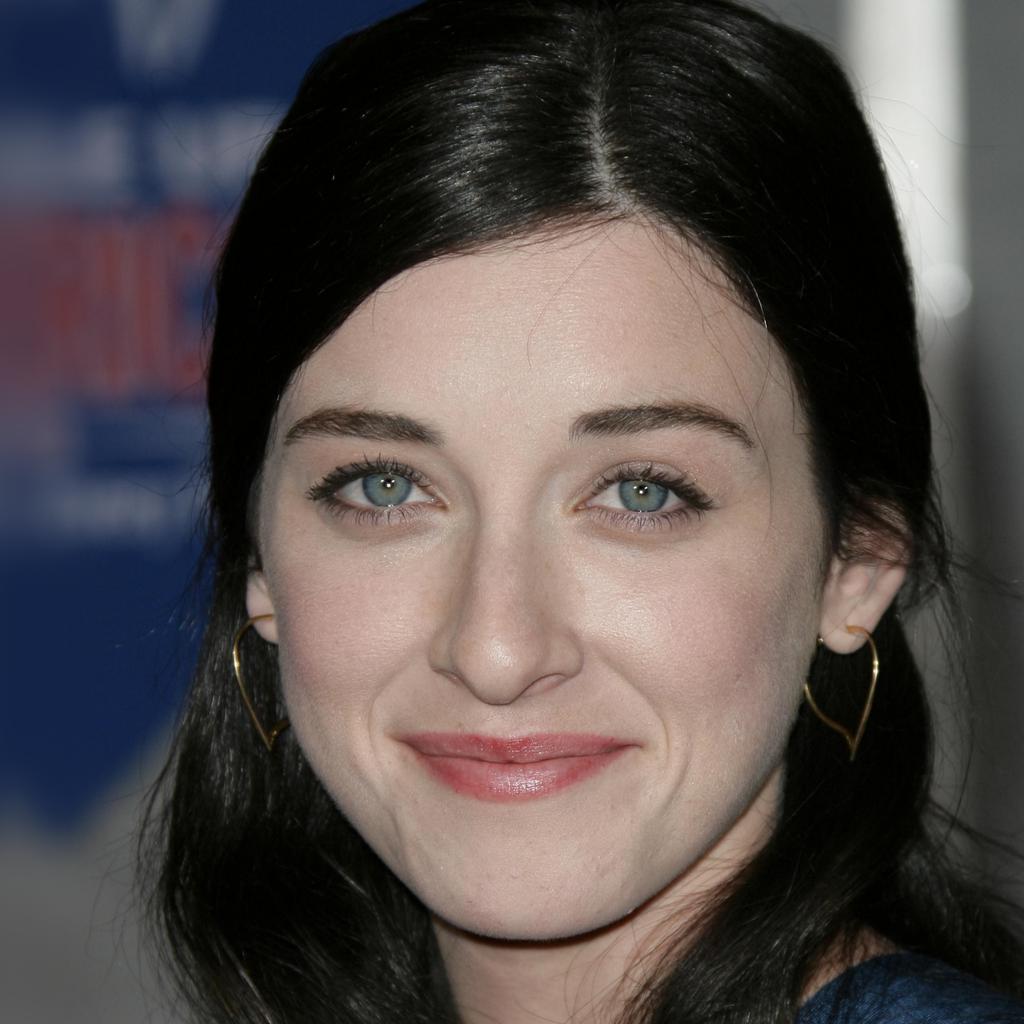}
  \hspace{0.5mm}
 \includegraphics[width=0.102\linewidth]{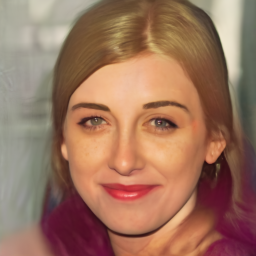}  \includegraphics[width=0.102\linewidth]{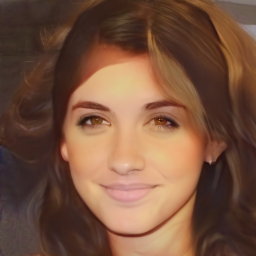} 
\tabularnewline
        { Input  \hspace{35pt} Generated Samples\hspace{50pt }Input  \hspace{35pt} Generated Samples\hspace{50pt }Input  \hspace{35pt} Generated Samples
        }\\
    \tabularnewline  %
\tabularnewline
\end{tabular}}
\vspace{-0.5\baselineskip}
\vspace{-3mm}
\hspace{20pt}\captionof{figure}{An illustration of various applications of our method. We use a diffusion model trained unconditionally and condition using our proposed algorithm only during the test time. We present the results on six tasks: (a) image inpainting, (b) colorization, (c) image super-resolution, (d) semantic generation, (e) identity replication, and (f) text-based image editing. In part (f), the text prompts for the the first and second columns, respectively, are ``This person has blonde hair'' and ``This person has wavy hair.''}
\label{fig:introfig}

\end{center}%
}]
\thispagestyle{empty}
\vspace{-5mm}

\blfootnote{* Work done during internship at MERL.}
\begin{abstract}
  Conditional generative models typically demand large annotated training sets to achieve high-quality synthesis. As a result, there has been significant interest in designing models that perform plug-and-play generation, i.e., to use a predefined or pretrained model, which is not explicitly trained on the generative task, to guide the generative process (e.g., using language). However, such guidance is typically useful only towards synthesizing high-level semantics rather than editing fine-grained details as in image-to-image translation tasks. To this end, and capitalizing on the powerful fine-grained generative control offered by the recent diffusion-based generative models, we introduce Steered Diffusion, a generalized framework for photorealistic zero-shot conditional image generation using a diffusion model trained for unconditional generation. The key idea is to steer the image generation of the diffusion model at inference time via designing a loss using a pre-trained inverse model that characterizes the conditional task. This loss modulates the sampling trajectory of the diffusion process. Our framework allows for easy incorporation of multiple conditions during inference. We present experiments using steered diffusion on several tasks including inpainting, colorization, text-guided semantic editing, and image super-resolution. Our results demonstrate clear qualitative and quantitative improvements over state-of-the-art diffusion-based plug-and-play models while adding negligible additional computational cost.

\end{abstract}
\vspace{-24pt}
\section{Introduction}
Deep diffusion-based probabilistic generative models\cite{ho2020denoising, sohl2015deep, dhariwal2021diffusion} are quickly emerging as one of the most powerful methods to synthesize high-quality content and have shown the potential to revolutionize content creation not only in computer vision, but also in many other areas including speech, audio, and language. Such models (e.g., ImagGen\cite{saharia2022photorealistic}, Stable Diffusion\cite{rombach2021highresolution}) have demonstrated outstanding synthesis results in conditional generation tasks, such as text-conditioned image synthesis \cite{balaji2022ediffi, ramesh2022hierarchical} and image reconstruction \cite{saharia2022palette, saharia2022image, preechakul2022diffusion}. However, these models do not typically possess zero-shot conditional generative abilities when used directly (zero-shot capabilities as are commonly seen in language foundation models such as GPT-3\cite{brown2020language}), and often demand large amounts of annotated and paired (multimodal) data for conditional generation, which may be challenging to obtain~\cite{huang2021multimodal}.

One way to circumvent this need for large annotated training sets is to leverage predefined models as plug-and-play modules \cite{nguyen2017plug, graikos2022diffusion,nair2023unite} in an otherwise unconditionally trained diffusion model. Specifically, in such plug-and-play models, a model is first trained in an unconditional setting (without labels). During inference, the plug-and-play modules (networks separately trained for a particular conditional task, e.g., image captioning) are incorporated in the reverse diffusion process to produce intermediate samples guided in the Markov chain in specific directions to satisfy the desired condition. Prior works, such as \cite{nguyen2017plug,nguyen2016synthesizing}, have proposed similar methods in which the authors derive text- or class-conditioned samples from Generative Adversarial Networks (GANs)\cite{goodfellow2020generative} that were trained without labels. To achieve this, they iteratively refine the noise input of the GAN until the desired sample satisfies the condition. Very recently, Grakios et al. \cite{graikos2022diffusion} proposed a diffusion-based plug-and-play method that enables using unconditional diffusion models for conditional generation utilizing class labels. Both these methods are specifically designed for tasks involving label-level semantics. However, these methods do not address the usage of unconditional models for general image-to-image translation tasks, which require synthesizing visual content conditioned on fine-grained details in the source image. There are also works that propose diffusion models for image-to-image translation, such as for image super-resolution and inpainting \cite{choi2021ilvr,lugmayr2022repaint }; however, these methods are task-specific and do not generalize well to new tasks or new types of inverse problems (as demonstrated in Section~\ref{sec:expts}). In this work, we present a generic framework that can generalize to any image-to-image translation task.

\begin{figure}[tb]
    \centering
    \includegraphics[width=\linewidth]{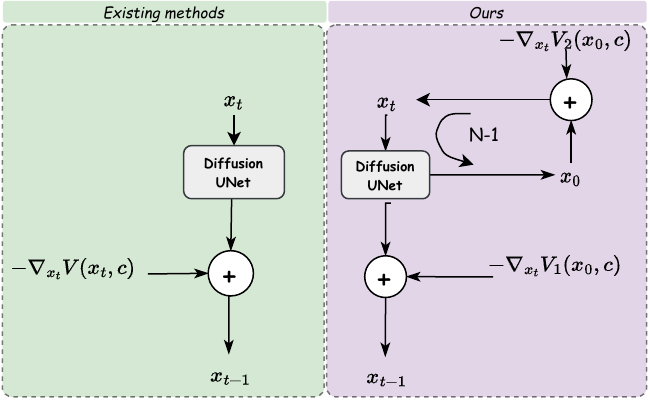}
    \vspace{-6mm}
  \caption{{An illustration of the difference between existing plug-and-play generation approaches (e.g.,~\cite{graikos2022diffusion}) and the proposed approach.} Existing plug-and-play works operate with an energy function $V$ of the noisy latent $x_t$. In contrast, our model uses the implicit prediction of the diffusion model (i.e., a coarse estimate of the clean image $x_0$) in its energy function $V_1$, which allows the use of any pre-trained network for steering. In addition, our model provides a looping mechanism $V_2$, which iterates $N$ times at each timestep $t$ to enhance generation quality.}
    \label{fig:intro2}
    \vspace{-5mm}
\end{figure}

  In this paper, we derive the necessary theory and formulate an algorithm, which we call \emph{Steered Diffusion}, for diffusion-based image editing and image-to-image translation; our model is subsequently validated on a wide range of tasks. Steered diffusion is motivated by the energy-based formulation of diffusion probabilistic models~\cite{gao2020learning}. In general, inference in a generative model can be thought of as deriving samples from a learned distribution. Recall that every probability density function can be formulated as an energy field that describes an unnormalized estimate of how the distribution density varies in space\cite{grathwohl2019your, nguyen2017plug}. 
  If one needs to find points in space that are the closest match to a given condition,  one can utilize gradient-based optimization algorithms to find points in the field that have the highest density value for the condition. The gradient-based optimization scheme can be viewed as a modulation of the energy toward the desired direction. Previous work has utilized this idea on GANs\cite{nguyen2017plug,nguyen2016synthesizing} and obtained reasonable results for label-based generation tasks. Due to their model structure, diffusion models are ideal candidates for such an energy modulation. One key challenge remains to design a good energy estimator that is robust to all noise levels. Previously, classifier-based guidance \cite{nichol2021glide,dhariwal2021diffusion} has been proposed and thought of as an energy modulation utilizing a pretrained classifier trained on noisy images. This poses a limitation that the guiding function should be noise-robust. In this work, we propose an alternative solution that does not need noise-robust networks but could use any network by utilizing the diffusion model as an implicit denoiser. Figure \ref{fig:intro2} gives a brief overview of how our approach is different from existing methods.

We present experiments using steered diffusion on multiple conditional generative tasks on faces as well as generic images as portrayed in Figure \ref{fig:introfig}, We present results on (i) identity replication \cite{deng2019arcface}, (ii) semantic image generation~\cite{park2019semantic}, (iii) linear inverse problems\cite{mei2022bi}, and (iv) text-conditioned image editing. Although our method is generic, for evaluations we perform experiments on faces. Before presenting our framework in detail, we now summarize the key contributions of our work:
\begin{itemize}[noitemsep]
    \item We propose {\it steered diffusion}, a general plug-and-play framework that can utilize various pre-existing models to steer an unconditional diffusion model.
    \item We present the first work applicable to both label-level synthesis and image-to-image translation tasks and demonstrate its effectiveness for various applications.
    \item We propose an implicit conditioning-based sampling strategy that significantly boosts the performance of conditional sampling from unconditional diffusion models compared with previous methods.
    \item We introduce a new strategy that uses multiple steps of projected gradient descent to improve sample quality.
 
\end{itemize}

\section{Background}

\subsection{Related Work}

Early works on unpaired image-to-image translation utilize a cycle consistency loss between the input and the target domains \cite{CycleGAN2017,FuCVPR19-GcGAN}. Newer works, such as~\cite{isola2017image}, have introduced a contrastive learning-based approach where a contrastive loss between corresponding patches of the input and target domains is minimized. The consistency-based method often fails to generate photorealistic images; hence, conditional generative models are preferred when labeled data are available. A few works \cite{saharia2021image,saharia2022image} utilize diffusion models for conditional image-to-image translation because of their photorealistic generation quality. 

Guiding diffusion models during inference time has been explored by several works, such as~\cite{nair2023unite}. The first method that proposed inference-time conditioning \cite{dhariwal2021diffusion} uses a pretrained noise-robust classifier to guide the inference of an unconditional model. GLIDE~\cite{nichol2021glide} proposed a method for conditioning using text. 
Earlier work in plug-and-play modelling for generative models utilized GANs and performed iterative refinement on the latent space of GANs~\cite{nguyen2017plug}. This method uses a predefined classifier or text captioning network to estimate a loss between the desired label output or text caption and the one generated from the GAN generator. This loss is backpropagated to refine the noise input of the GAN iteratively until the generator predicts the desired output. Recently,\cite{graikos2022diffusion} proposed a method that uses diffusion models as a plug-and-play prior for class-conditioned generation. Several works have addressed the task of image-to-image translation using unconditional diffusion models \cite{kawar2022denoising, choi2021ilvr, lugmayr2022repaint,avrahami2022blended}, but each of these proposes a task-specific inference scheme. For example, ILVR~\cite{choi2021ilvr} performs image super-resolution, and Reinpaint~\cite{lugmayr2022repaint} performs image inpainting. Blended diffusion~\cite{avrahami2022blended} proposes a method for text-conditioned image editing. DDRM~\cite{kawar2022denoising} proposes an inference-time scheme offering a general solution for linear inverse problems such as colorization and super-resolution.
\subsection{Concurrent Work}
In concurrent work to ours that also explores zero-shot conditional generation using diffusion models, \cite{bansal2023universal} used a text to image model\cite{rombach2021highresolution} and a two-step forward and backward universal guidance process, but it works well only after heavy optimization on network-based inverse problems such as semantic generation and identity generation. Another concurrent work\cite{song2022pseudoinverse} explored the usage of unconditional diffusion models for linear inverse problems using a pseudo-inverse model. In contrast to all these prior works, our steered diffusion algorithm generalizes well to both image-to-image translation tasks and high-level label-based generation tasks.

\subsection{Denoising Diffusion Probabilistic Models} 

Denoising diffusion probabilistic models (DDPMs) \cite{sohl2015deep,ho2020denoising} belong to a class of generative models in which the model learns the distribution of data through a Markovian sampling process. DDPMs consist of a forward process and a reverse process. Let $x_t$ denote the latent state of an input image at timestep $t$ in a diffusion process. The sampling operation $q(\cdot)$ for the forward process in DDPM is defined as:
\begin{equation}
    q(x_t|x_{t-1}) := \mathcal{N}(x_t; \sqrt{1-\beta_t} \, x_{t-1}, \beta_t I),
    \label{eq:q_sample}
\end{equation}
where $\{\beta_t\}$ is a predefined variance schedule and $I$ is the identity matrix. 
 
The forward process can be considered as a \emph{noising} operation, where the next state $x_t$ is obtained from the current state $x_{t-1}$ by adding a small amount of Gaussian noise according to the sampled timestep. The state $x_t$ at timestep $t$ can also be sampled \emph{directly} from the initial state $x_0$, using: 
\begin{equation}
    q(x_t|x_{0}) := \mathcal{N}\big(x_t; \sqrt{\bar{\alpha}_t} x_0, (1-\bar{\alpha}_t) I\big),
    \label{eq:q_sample0}
\end{equation}
\noindent or equivalently,
\begin{equation}
    x_t = x_0 \sqrt{\bar{\alpha}_t}  + \epsilon \sqrt{1-\bar{\alpha}_t}, \quad \epsilon \sim \mathcal{N}(0, I),
\end{equation}
where $\bar{\alpha}_t=\prod_{s=1}^t\alpha_s$ and $\alpha_t=1-\beta_t$.

In~\cite{sohl2015deep}, it is shown that if the number of time steps is large and the increment in $\{\beta_i\}$ is small, then each step in the reverse sampling process can also be approximated by a Gaussian. If $\mu_\theta$ and $\Sigma_\theta$ respectively denote the mean and the covariance of this Gaussian, modelled via neural networks with parameters $\theta$, then each reverse step samples the state $x_{t-1}$ according to:
\begin{equation}
    p_{\theta}(x_{t-1}|x_t) := \mathcal{N}\big(x_{t-1}; \mu_{\theta}(x_t,t),\Sigma_{\theta}(x_t,t)\big). 
\end{equation}
The parameters $\theta$ are obtained by minimizing the variational lower bound of the negative log-likelihood of the data distribution.

\section{Proposed Method}
\begin{figure*}[tb]
    \centering
    \includegraphics[width=\linewidth]{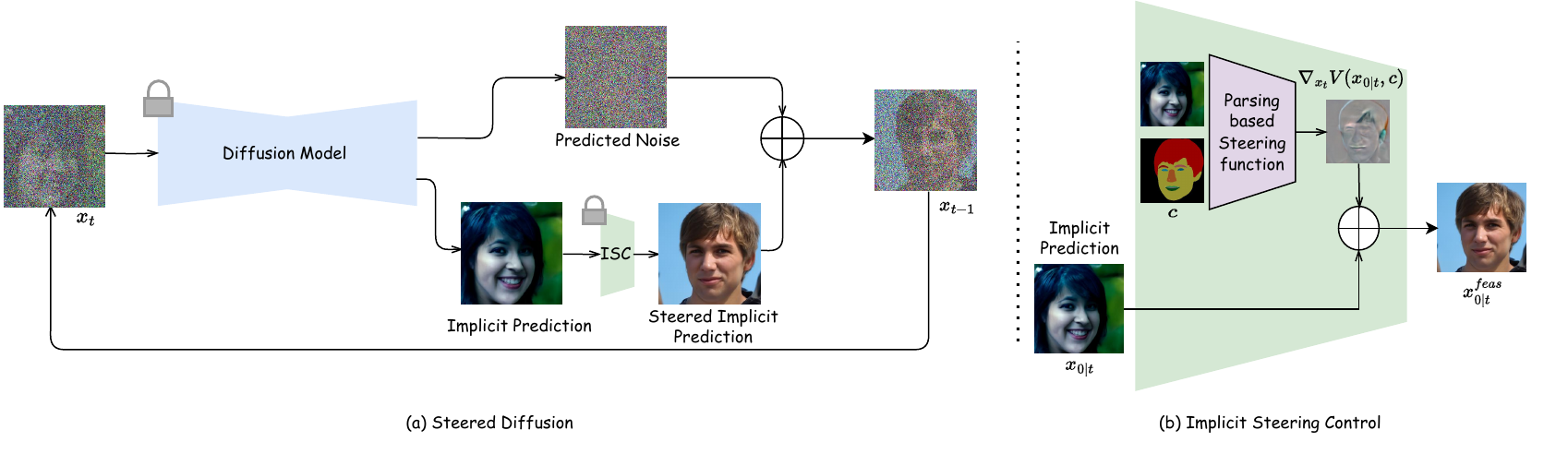}
    \vspace{-9mm}
  \caption{{An illustration of Steered Diffusion. During each step of the sampling process the implicit prediction is steered to the direction of the condition using a steering network or predefined function. Note that this figure is only for illustrating the idea and does not show the actual sampled images; in the actual sampling, the steering process is much more gradual, not sudden as potrayed in this image.}}
    \label{fig:steered}
    \vspace{-5mm}
\end{figure*}
\subsection{Steered Diffusion at Inference Time}
\label{sec:theory}
Our work is motivated by the energy-based formulation of diffusion models. 
For any probability density function, the corresponding energy-based model (EBM) is defined by:
\begin{align}
    p_{\theta}(x) &=\frac{\exp\big( -V(x)\big)}{Z}, 
    \label{eq:ebm}
\end{align}

\noindent where $V(x)$ denotes the corresponding energy function across states $x$, and $Z$ denotes a normalization constant. To derive samples from this distribution, one can utilize the Langevin equation~\cite{sekimoto1998langevin} describing the state transition of a particle in the presence of an energy field. For diffusion models the sampling step is

\begin{align}
x_{t-1} \!= x_t - \nabla_{x_t}\log p_{\theta}(x_{t-1}|x_t) + \epsilon, \; \epsilon \!\sim\! \mathcal{N}(0,I).
\label{eq:reverse_langevin}
\end{align}
The term $\nabla_{x_t} \text{log }p_{\theta}(x_{t-1}|x_t)$ is called the \emph{score function} of the density $p_{\theta}(x_{t-1}|x_t)$.  One key advantage of the energy-based formulation is that it allows modulation of the energy function to satisfy given criteria. This was initially introduced as classifier guidance \cite{dhariwal2021diffusion}, which allows label conditional sampling from an unconditionally trained diffusion model utilizing a noise-robust classifier. In the remaining part of this section, we motivate how we can extend the functionality of unconditional diffusion models to conditional tasks. 
Consider a conditional sampling scenario based on a condition $c$, for sampling from a state $x_t$ to state $x_{t-1}$. The conditional transition probability $p_{\theta}(x_{t-1}|x_{t},c)$ can be decomposed as 

\begin{align}
    \label{eq:diffeq}
   p_{\theta}(x_{t-1}|x_{t},c)\propto \frac{p_{\theta}(x_{t-1}|x_{t})
   p(c|x_{t-1})}{p(c|x_{t})}.
\end{align}

 Hence, for any timestep $t$, the effective score for conditional transition can be found by utilizing using the log of probability density in the EBM formulations (\ref{eq:ebm}) of the individual densities and can be represented as
\begin{align}
   &\nabla_{x_t} \text{log }p_{\theta}(x_{t-1}|x_{t},c) = \notag\\
   &\nabla_{x_t} \text{log }p_{\theta}(x_{t-1}|x_{t})-\nabla_{x_t} V_1(x_t,c)+\nabla_{x_t} V_2(x_{t-1},c),
   \label{eq:energy}
\end{align}
 where $V_1$ and $V_2$ are the corresponding energy functions that model the conditional distributions of $x_t$ and $x_{t-1}$ given a condition $c$. Specifically, they project the higher dimensional $x_t$ to the lower dimensional space of $c$ and measure the distance between the mapped value and $c$. The better this particular measure, the more effectively it can be used to generate conditional samples from an unconditional model. Using Eq.~\eqref{eq:energy}, the the conditional sampling equation for the reverse process is
\begin{align}\label{conditional_sampling}
x_{t-1} = & \frac{1}{\sqrt{\alpha_t}}\bigg( x_t - \frac{1-\alpha_t}{\sqrt{1-\bar\alpha_t}} \epsilon^{(t)}_\theta(x_t, t) \bigg)\notag\\
&-\nabla_{x_t} V_1(x_t,c)+\nabla_{x_t} {V_2}(x_{t-1},c)+ \sigma_t \epsilon.
\end{align}
Here $\epsilon^{(t)}_\theta(x_t, t)$ is the network prediction at a timestep $t$, and $\sigma_t$ is the corresponding variance of the reverse step. The formulation in Eq. \eqref{conditional_sampling} shows that the energy function requires a functional mapping from a noisy $x_t$ to $c$.

\label{sec:energyfn}
In many applications of interest, the mapping function from $x_t$ to $c$ is complex and can be modeled effectively using deep networks. For example, in image-to-image translation tasks such as image generation from semantic maps\cite{park2019semantic}, the image is $x_t$ and the semantic map is the condition $c$. Similarly, for text to image generation, the image is $x_t$ and $c$ corresponds to the text. Ideally, we would like to employ an existing (pre-trained) deep neural network for the mapping from $x_t$ to $c$. However, deep networks are usually trained on clean images, which limits the usability of existing pre-trained deep networks for mapping directly from the noisy image $x_t$ to $c$. One workaround would be to use noise-robust networks to map from $x_t$ to $c$, but training noise-robust networks for conditional mapping can be computationally expensive. Moreover, a network that is trained with multiple different noise levels often results in lower mapping performance, as it cannot denoise all noise levels accurately; we validate this claim experimentally in Section~\ref{sec:ablation}. Alternatively, one could include two mapping functions: a first that denoises $x_t$, and a second that maps from the denoised image to $c$. Rather than training a seperate denoising network, however, we realized that diffusion models are inherently trained as denoisers, and reconstruction quality improves as time proceeds in the reverse sampling of the diffusion process. Because of this capacity, we can use a reverse sampling step to make coarse predictions of the denoised image from any time step $t$. 

Hence, we modify our original energy expression (\ref{eq:energy}) to: 
\begin{align}
   \nabla_{x_t} &\text{log } p_{\theta}(x_{t-1}|x_{t},c) =
   \nabla_{x_t} \text{log } p_{\theta}(x_{t-1}|x_{t})\notag\\ &-\nabla_{x_t} V_1(x_{0|t},c) -\delta_1 + \nabla_{x_t} V_2(x_{0|t-1},c)+\delta_2,
\end{align}
where, we define the implicit step prediction $x_{0|t}$ as:
\begin{align}
    x_{0|t} &=\frac{x_t - \sqrt{1 - \bar{\alpha}_t} \, \epsilon_\theta^{(t)}(x_t)}{\sqrt{\bar{\alpha_t}}}. 
    \label{eq:implicit}
\end{align}
Here, we assume $x_t$ and $x_{t-1}$ are first denoised to  $x_{0|t}$ and $x_{0|t-1}$, respectively.
 The terms $\delta_1$ and $\delta_2$ capture the errors arising from the shift in the domain from $x_t$ to $x_0$ and from $x_{t-1}$ to $x_0$; for large $t$, $\delta_1 \approx \delta_2$ as the implicit predictions at nearby steps tend to be similar.

As shown in our experiments, the energy function should be selected according to the task. An easy way to choose the energy function is by looking at the training loss of the mapping network. For example, in the case of semantic generation, a good energy function is the cross-entropy loss between the predicted semantic map at any timestep and the input semantic map. In the case of identity replication, a good choice of regularization would be the negative cosine similarity score between the embeddings from a recognition network for the input and target image. In the case of text-to-image generation, it would be  CLIP loss \cite{radford2021learning}. As a rule of thumb, an energy function could be chosen easily by looking at the loss function used to train the pre-trained network (or an inverse function) that maps from the image $x$ to the condition $c$.

\begin{algorithm}
\caption{Steered Diffusion}
\label{ref:algo1}
\begin{algorithmic}[1]
\renewcommand{\algorithmicrequire}{\textbf{Input:}}
\renewcommand{\algorithmicensure}{\textbf{Initialize:}}
\Require Energy function $V$, condition $c$ 
\State $x_T \sim \mathcal{N}(x_T; 0,I)$
\For{$t = {T-1}, \ldots, 1$}
    \For{$n = N, \ldots, 1$}
         \State $\epsilon \sim \mathcal{N}(\epsilon; 0,I)$
    \State     $ x_{0|t} =\frac{x_t - \sqrt{1 - \bar{\alpha}_t} \, \epsilon_\theta^{(t)}(x_t)}{\sqrt{\bar{\alpha_t}}}$
    \State Compute $ x_{0|t}^\mathrm{feas}$ with $V,c$ using Eq. (\ref{feas_eq})  

        \If{$(n>1)$}
            \State  Compute $x_{t}^{uc}$ using Eq. (\ref{eq:uc})
        \Else
            \State  
            Compute $x_{t-1}^{uc}$ using Eq. (\ref{eq:uc})
        \EndIf
\EndFor
\EndFor

\\
\Return $x_0$
\end{algorithmic}
\label{test algo}
\end{algorithm}
\subsection{Revisiting Sampling in Diffusion Models}
To obtain a closed-form expression for plugging the energy-based formulation into the reverse sampling process efficiently, we take inspiration from DDIM\cite{song2020denoising} and revisit the reverse sampling operation of diffusion models. From $p_\theta(x_{1:T})$, one can generate a sample $x_{t-1}$ from a sample $x_{t}$ by:

\begin{align}
    x_{t-1}^{\text{uc}}  &= \sqrt{\bar{\alpha}_{t-1}}\cdot \underbrace{\frac{x_t - \sqrt{1 - \bar{\alpha}_t} \, \epsilon_\theta^{(t)}(x_t)}{\sqrt{\bar{\alpha_t}}}}_{\text{`` predicted } x_0 \text{''}} \notag\\
    + &\underbrace{\sqrt{1 - \bar{\alpha}_{t-1} - \sigma_t^2} \cdot \epsilon_\theta^{(t)}(x_t)}_{\text{``direction pointing to } x_t \text{''}} + \underbrace{\sigma_t \epsilon}_{\text{random noise}}, \label{eq:sample-eq-gen}
\end{align}
as in Song et al. \cite{song2020denoising}. 
Using Eq.~\eqref{eq:implicit}, we can rewrite the unconditional sampling step Eq.~\eqref{eq:sample-eq-gen} in terms of $x_{0|t}$:
\begin{align}
\label{eq:uc}
    x_{t-1}^{\text{uc}} =  &\sqrt{\bar{\alpha}_{t-1}} x_{0|t} \;+ \notag\\ 
    &\sqrt{1 - \bar{\alpha}_{t-1} - \sigma_t^2} \cdot \frac{ x_t - \sqrt{\bar{\alpha_t}} x_{0|t} }{\sqrt{1 - \bar{\alpha}_t}}+ \sigma_t \epsilon,
\end{align}
Here the superscript uc denotes the unconditional sample, which is obtained without any steering while transitioning from $x_t$ to $x_{t-1}$. The conditional sampling step~\eqref{eq:energy} can then be rewritten as
\begin{align}
    x_{t-1}  =  x_{t-1}^{\text{uc}}
   -\nabla_{x_t} V_1(x_{0|t},c)  + \nabla_{x_t} V_2(x_{0|t-1},c).
 \label{eq:cond_sample}
\end{align}
Through this, we can modulate $x_0$ directly, as $x_{0|t}$ is also a function of $\epsilon_{\theta}^{(t)}(x_t)$. Following Eq.~\eqref{eq:cond_sample}, a rough estimate of the desired $x_0$ for conditional sampling, represented by $x_{0|t}^\mathrm{feas}$, can be obtained using 
\begin{align}
    &x_{0|t}^\mathrm{feas} = x_{0|t}  -k(t) \nabla_{x_t}\left( V_1(x_{0|t},c) -V_2(x_{0|t-1},c)\right),
    \label{feas_eq}
\end{align}
where $k(t)$ is a scaling factor defining the strength of regularization.  We call the process of finding $x_{0|t}^\mathrm{feas}$ from $x_{0|t}$ {\it Implicit Steering Control} (ISC), and we call the new sampling process {\it steered diffusion}. The exact algorithm is illustrated in Fig.~\ref{fig:steered} and explained in Algorithm~\ref{ref:algo1}.

\section{Tips for Improved Performance}
\subsection{Linear Inverse Problems}

\begin{figure}[!tb]
    \centering
    \includegraphics[width=0.8\linewidth]{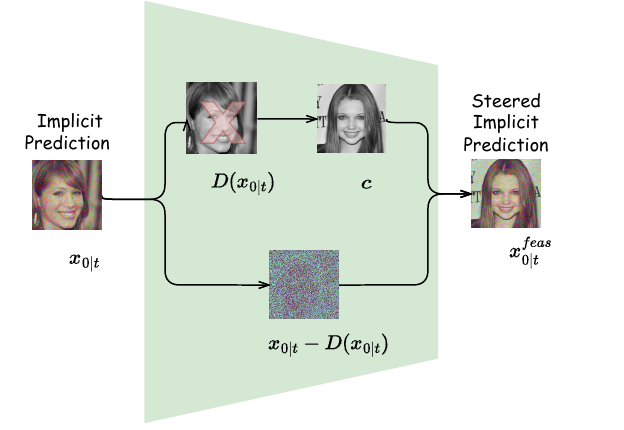}
    \vspace{-3mm}
  \caption{{An illustration of the steering function for linear inverse problems. For linear inverse problems, the component of the implicit prediction along the degradation direction can be replaced by the ground truth condition.  }}
    \label{fig:isclinear}
    \vspace{-5mm}
\end{figure}

For optimization-based inverse problems such as text-to-image generation and semantic map to natural image generation, the exact mapping function is not always available. On the other hand, for linear inverse problems such as colorization, super-resolution, and image inpainting, the mapping is simply a linear function, and we can write Eq.~\eqref{feas_eq} more simply.  In these cases, the exact mapping function to the latent space of the condition is known. Hence, one can decompose the implicit prediction at each timestep along the direction of the condition and simply replace this component by its desired ideal condition, i.e., if our predicted sample needs to map to a condition c, then the modified implicit prediction step becomes 
\begin{align}
  x^\mathrm{feas}_{0|t} = x_{0|t} +k(t)(D(y)-D(x_{0|t} )), \quad
   \text{where } c=D(y).
\end{align}
Here, $y$ is the clean image and $D$ is the known degradation model. Our sampling procedure ensures that the series of operations preserve the consistency of domains of $x_{t-1}$ and $x_{0|t}^\mathrm{feas}$ to the original data distribution at the corresponding timesteps. An illustration is shown in Figure \ref{fig:isclinear}.

\subsection{Multi-Step Implicit Modulation}
\label{sec:multisample}
 Our experiments show (see Fig.~\ref{fig:color}) that performing refinement using Eq.~(\ref{feas_eq}) on the implicit step prediction multiple times for each timestep significantly boosts the conditioning quality for more ill-posed conditions such as image inpainting and colorization. A similar observation was also found by~\cite{lugmayr2022repaint}. Specifically, at a particular timestep $t$, we iterate the procedure of steering towards the next sampling step $x_{t-1}$ and then adding noise to return to $x_t$.
We present the corresponding algorithm in Algorithm~\ref{test algo}.  An example is shown in Fig.~\ref{fig:color}, where more realistic images are generated using the multiple-step sampling scheme row labeled ``OURS multi." Effectively, the $V_2$ term in Eq.~\eqref{eq:sample-eq-gen} can be thought of as enabling a multistep sampling in which we modulate the current step by looking ahead to the next sampling step. On a careful analysis of Eq.~\eqref{eq:diffeq}, i.e. from the score contribution due to the different regularization functions, we can see that the term for $\nabla_{x_t} V_1(x_t,c)$ modulates $x_t$ based on its current state, and the term from $\nabla_{x_{t}} V_2(x_{t-1},c)$ is a look-ahead correction where the derivative with respect to the future prediction is found. This is exactly what happens in the case of looping back from  $x_{t-1}$ to $x_{t}$, where $x_t$ is modulated iteratively by looking forward to what the future prediction would be. 

\subsection{Choosing the Scaling Factor ${k(t)}$}
In Eq.~\eqref{feas_eq}, $k(t)$ denotes the strength of the regularization constraint. A very small $k(t)$ would denote no effective regularization, and a large $k$ would lead to the diffusion process going out of the latent space manifold. Since the derivative of the regularization function by itself is a score value, similar to the normal scaling value of the score function, the appropriate time-varying normalization factor is $\sqrt{1-\bar{\alpha_t}}$. The exact value for k(t) for each task is defined in Table~\ref{table:paramterset}. For linear inverse problems we use a constant $k(t)=1$, which provided the best results.

\section{Experiments}
\label{sec:expts}
We evaluate the performance of our network qualitatively and quantitatively using four image-to-image translation tasks---semantic layout to face image translation, face inpainting, face colorization, and face super-resolution---as well as two high-level vision tasks: identity-based image generation and text-guided image editing. Unlike existing approaches, our method is completely zero-shot and applies to a wide variety of tasks. We compare with other diffusion-based approaches best applicable for each task for a fair evaluation. We also compare the semantic layout to image translation performance with that of task-specific unsupervised methods. We choose the unconditional model 
released by \cite{choi2021ilvr} as the unconditional pretrained diffusion model in all of our experiments with faces. Note that the sampling scheme in ILVR\cite{choi2021ilvr} and Repaint\cite{lugmayr2022repaint} can be thought of as happening at time $t$ rather than at the implicit step as in our method. Hence, comparing these methods for super-resolution and inpainting in our experiments below can be considered an additional ablation study highlighting the improvement from our implicit sampling.

\begin{table*}[!t]
\begin{center}
\vspace{-2mm}
\scalebox{0.7}{
\begin{tabular}{c| c c c | c c c c }
\toprule
\multirow{2}{*}{Method}&\multicolumn{3}{c}{Linear inverse problems}&\multicolumn{3}{c}{Complex inverse problems}\\
\cmidrule(lr){2-4} \cmidrule(lr){5-7}
 & Colorization & Inpainting & Super-resolution & Semantic Generation & Identity Replication & Image Editing\\
\midrule

$V_1$ &$(D(x_{0|t})-c)^2$&$(D(x_{0|t})-c)^2$&$(D(x_{0|t})-c)^2$&$\text{BCE}(D(x_0),c)$&$1-\frac{D(x_{0|t}).D(c)}{|D(x_t)||D(c)|}$& $3 (D_1(x_{0|t}),D_1(c_1))^2+ 2000 CS(D_2(x_{0|t}), c_2)$\\
$V_2$ &$(D(x_{0|t-1})-c)^2$&$(D(x_{0|t-1})-c)^2$&$0$&0&0&0\\
$D$ &Grayscaling&Masking&Downscaling&FARL \cite{zheng2021farl}&FARL recognizer\cite{zheng2021farl}& VGG Face \cite{Parkhi15} \& FARL CLIP\cite{radford2021learning}\\
$N$ &3&3&1&1&1&1\\
$k(t)$ &1&1&1&20000$\sqrt{1-\bar{\alpha_t}}$&3000$\sqrt{1-\bar{\alpha_t}}$&10$\sqrt{1-\bar{\alpha_t}}$ \& 1500$\sqrt{1-\bar{\alpha_t}}$\\

\bottomrule
\end{tabular}}
\end{center}
\vspace{-1.5em}
\caption{Parameter set for each application.}
\label{table:paramterset}
\end{table*}
\subsection{Implementation Details}
For our experiments, we utilize pixel-level unconditional diffusion models. For faces, we utilize the model trained on the FFHQ dataset\cite{karras2019style} that was released by\cite{choi2021ilvr}. For generic images, we use the model trained on ImageNet\cite{deng2009imagenet} released in ADM\cite{dhariwal2021diffusion}. All our experiments use 100 steps of sampling.
\subsection{Semantic Face Generation}
\begin{figure}[t!]
    \centering
    \begin{subfigure}[t]{0.19\linewidth}
      \captionsetup{justification=centering, labelformat=empty, font=scriptsize}
      \includegraphics[width=1\linewidth]{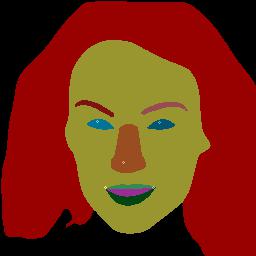}
      \includegraphics[width=1\linewidth]{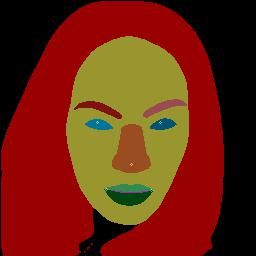}
      \includegraphics[width=1\linewidth]{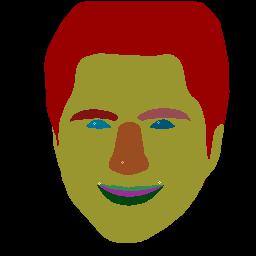}

      \caption{Labels}
    \end{subfigure}
    \begin{subfigure}[t]{0.19\linewidth}
      \captionsetup{justification=centering, labelformat=empty, font=scriptsize}

      \includegraphics[width=1\linewidth]{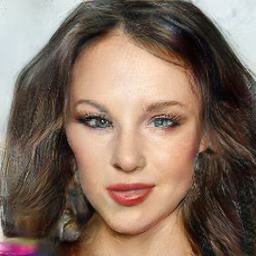}
      \includegraphics[width=1\linewidth]{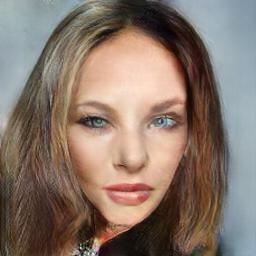}

      \includegraphics[width=1\linewidth]{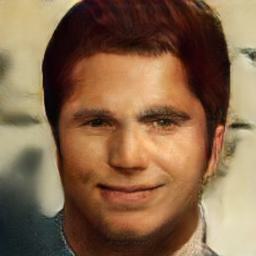}

      \caption{CycleGAN}
    \end{subfigure}
    \begin{subfigure}[t]{0.19\linewidth}
      \captionsetup{justification=centering, labelformat=empty, font=scriptsize}
      \includegraphics[width=1\linewidth]{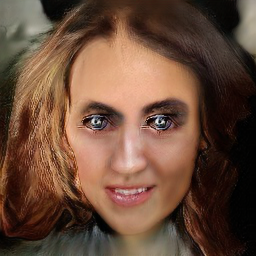}
      \includegraphics[width=1\linewidth]{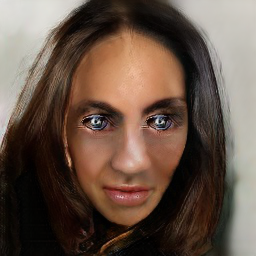}

      \includegraphics[width=1\linewidth]{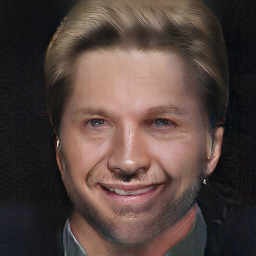}

      \caption{CUT}
    \end{subfigure}
    \begin{subfigure}[t]{0.19\linewidth}
      \captionsetup{justification=centering, labelformat=empty, font=scriptsize}
   
      \includegraphics[width=1\linewidth]{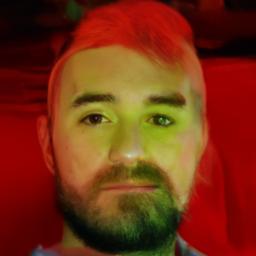}
      \includegraphics[width=1\linewidth]{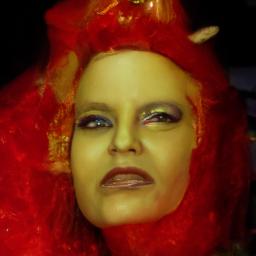}

      \includegraphics[width=1\linewidth]{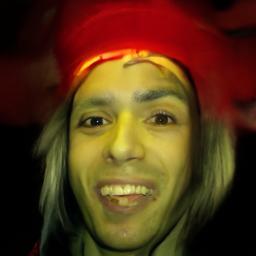}
      \caption{ILVR}
    \end{subfigure}
        \begin{subfigure}[t]{0.19\linewidth}
      \captionsetup{justification=centering, labelformat=empty, font=scriptsize}
   
      \includegraphics[width=1\linewidth]{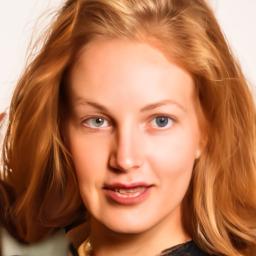}
      \includegraphics[width=1\linewidth]{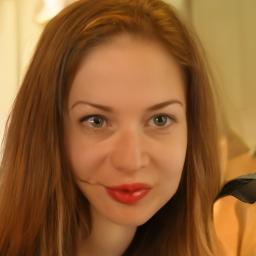}

      \includegraphics[width=1\linewidth]{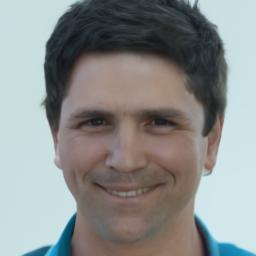}
      \caption{OURS}
    \end{subfigure}
    \vspace{-3mm}    
    \caption{Qualitative comparisons for semantic generation. }
    \label{fig:facesematic}
    \vspace{-2mm}
  \end{figure}

\begin{figure}[!tp]
    \centering
    \setlength{\tabcolsep}{1pt}
    {\small
    \renewcommand{\arraystretch}{0.5} 
    \begin{tabular}{c c c c c c}
    \captionsetup{type=figure, font=scriptsize}
    \raisebox{0.3in}{\rotatebox[origin=t]{90}{\scriptsize Low Res}}&
    \includegraphics[width=0.19\linewidth]{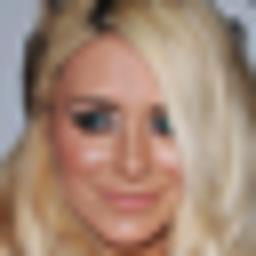}&
    \includegraphics[width=0.19\linewidth]{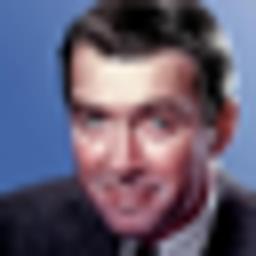}&
    \includegraphics[width=0.19\linewidth]{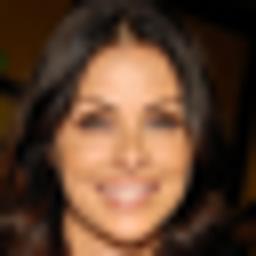}&
    \includegraphics[width=0.19\linewidth]{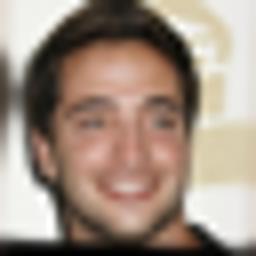}&
    \includegraphics[width=0.19\linewidth]{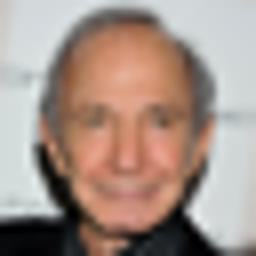}
    \tabularnewline
    \raisebox{0.2in}{\rotatebox[origin=t]{90}{\scriptsize ILVR}}&
    \includegraphics[width=0.19\linewidth]{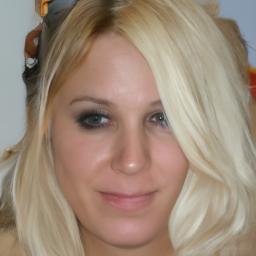}&
    \includegraphics[width=0.19\linewidth]{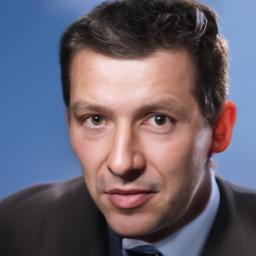}&
    \includegraphics[width=0.19\linewidth]{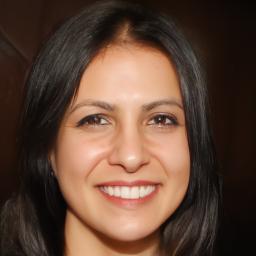}&
    \includegraphics[width=0.19\linewidth]{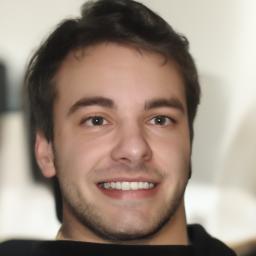}&
    \includegraphics[width=0.19\linewidth]{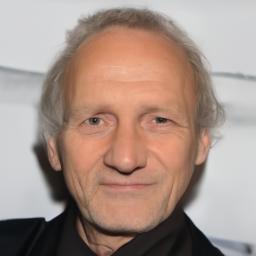}
    \tabularnewline
    \raisebox{0.2in}{\rotatebox[origin=t]{90}{\scriptsize PULSE}}&
    \includegraphics[width=0.19\linewidth]{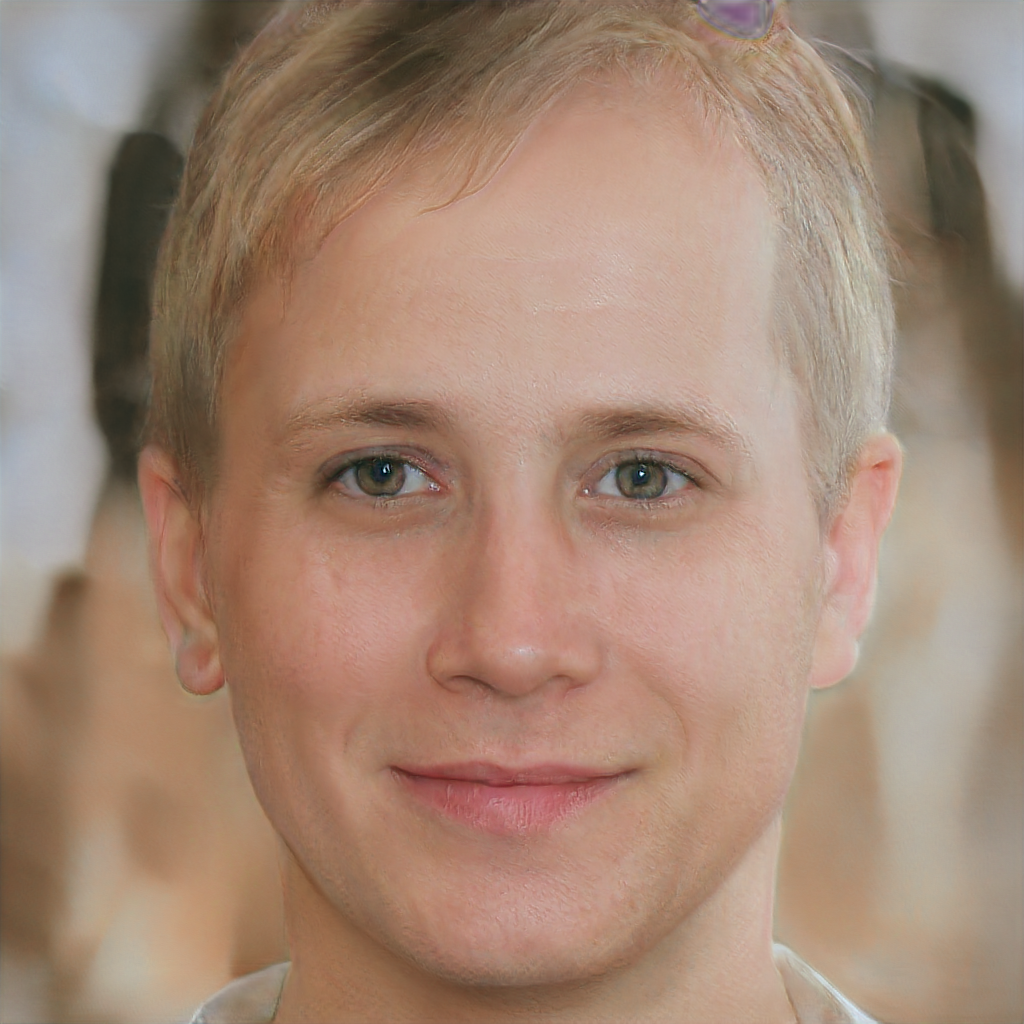}&
    \includegraphics[width=0.19\linewidth]{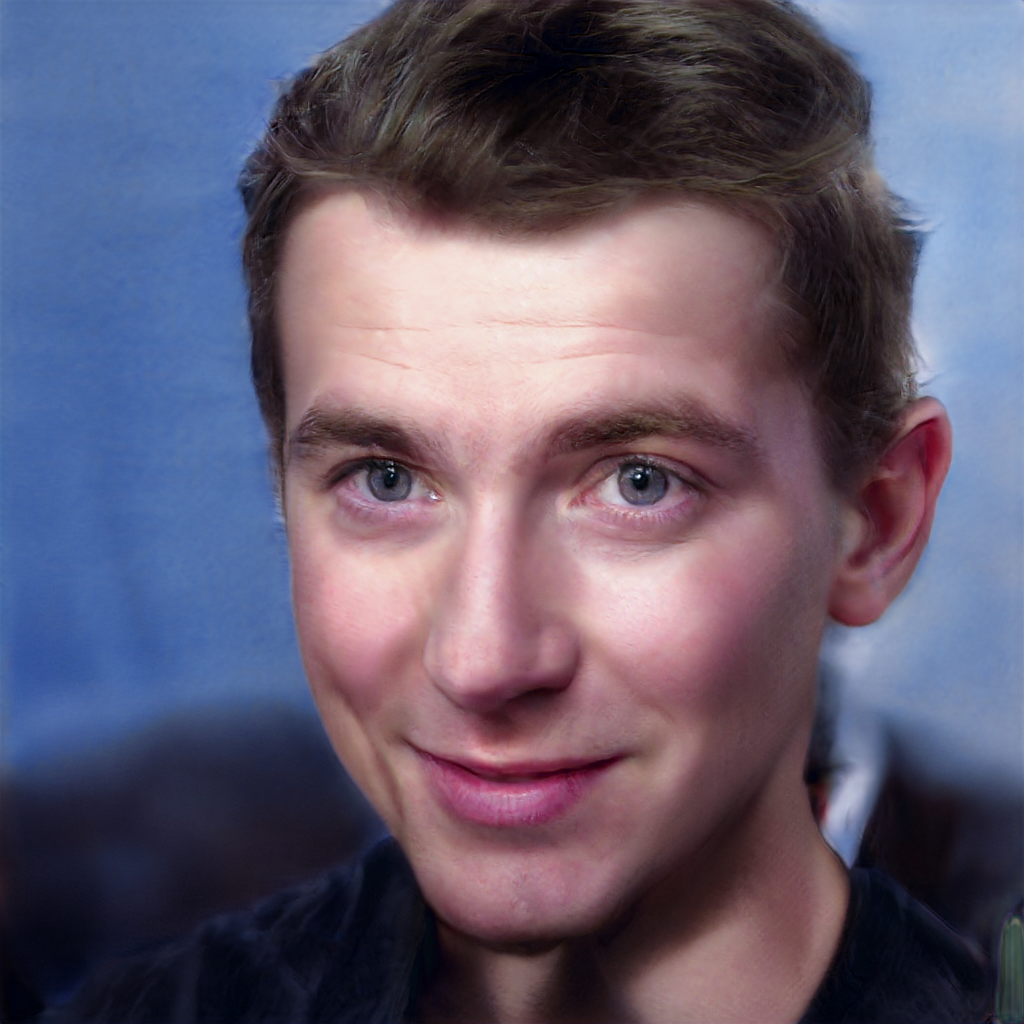}&
    \includegraphics[width=0.19\linewidth]{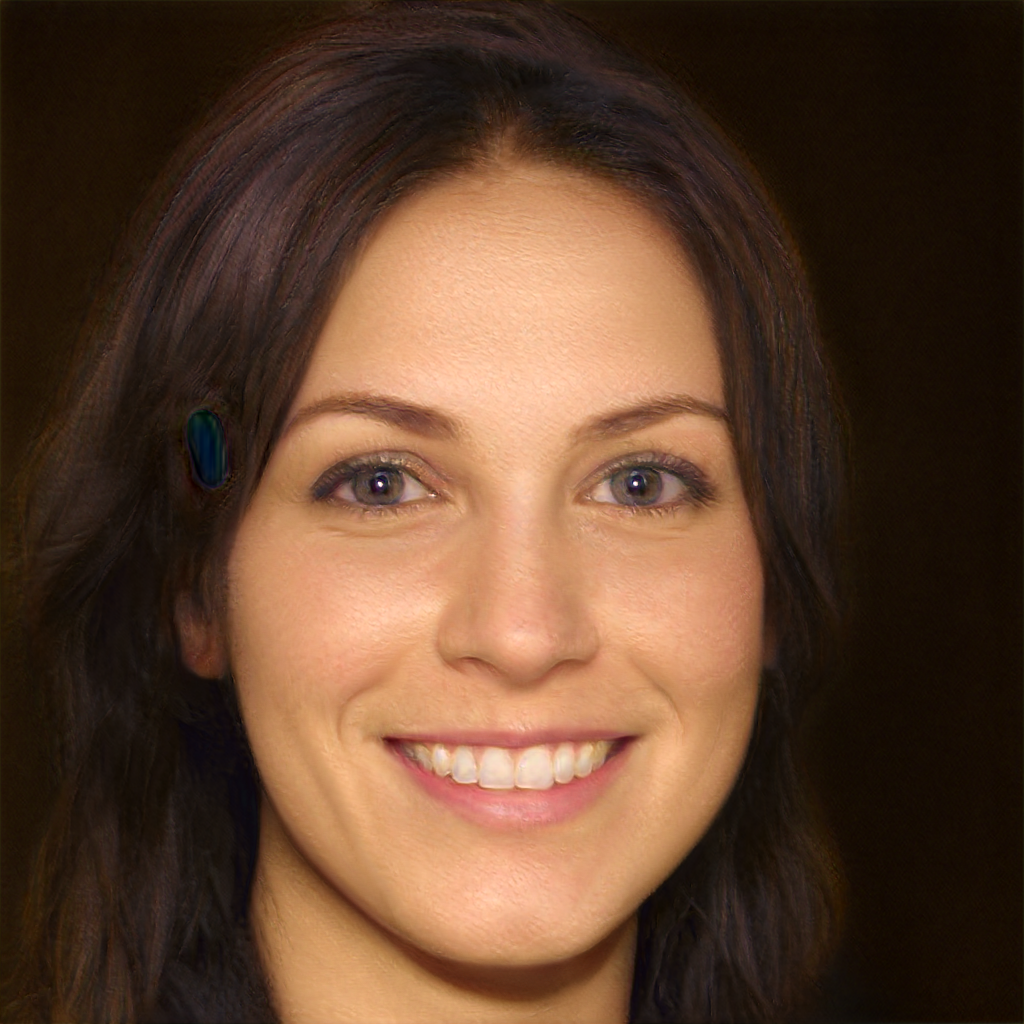}&
    \includegraphics[width=0.19\linewidth]{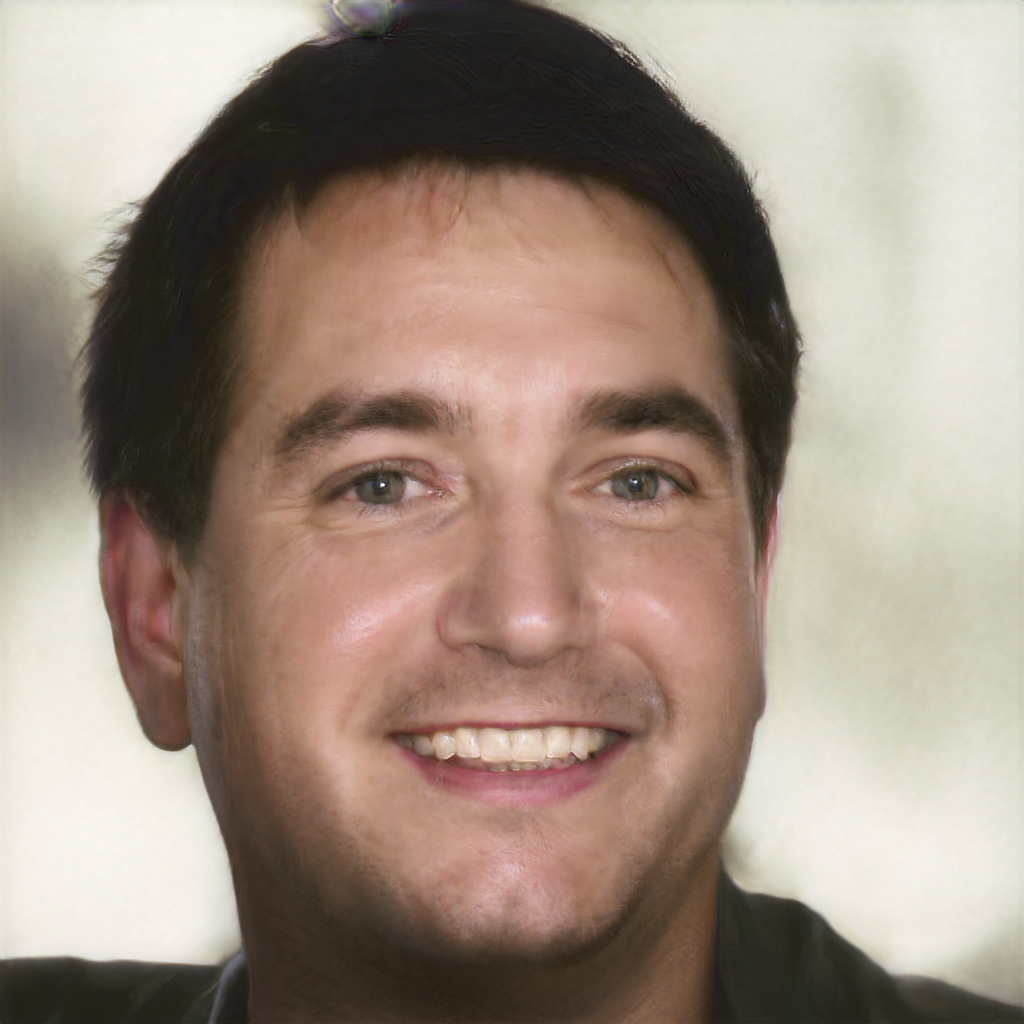}&
    \includegraphics[width=0.19\linewidth]{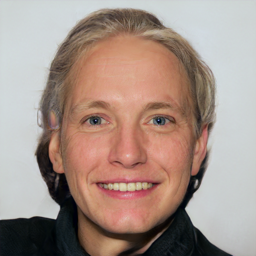}

    \tabularnewline
    \raisebox{0.2in}{\rotatebox[origin=t]{90}{\scriptsize OURS}}&
    \includegraphics[width=0.19\linewidth]{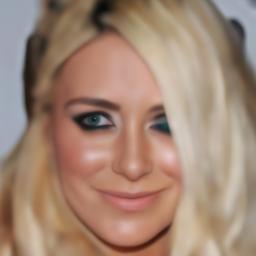}&
    \includegraphics[width=0.19\linewidth]{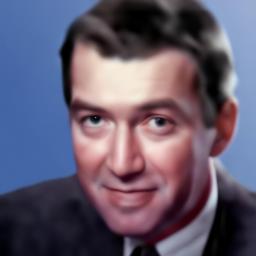}&
    \includegraphics[width=0.19\linewidth]{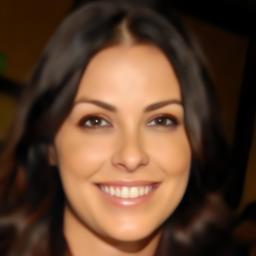}&
    \includegraphics[width=0.19\linewidth]{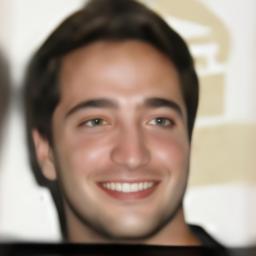}&
    \includegraphics[width=0.19\linewidth]{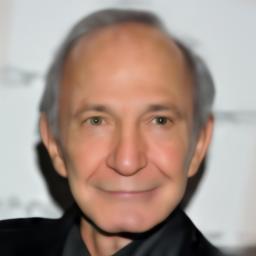}
    
     \tabularnewline

\end{tabular}}
\vspace{-.1in}
\caption{Results on $8 \times$ super resolution.}
\label{fig:face_sr}
\vspace{-6mm}
\end{figure}

To evaluate how our method performs in generic image-to-image translation tasks, we evaluate our method's performance for the task of semantic layout to face generation. We utilize the CelebA dataset for this. To generate the semantic labels, we use facer\cite{zheng2021farl} and create $11$ label classes for each face. Since there is no other unconditional model that can perform fully test-time semantic generation, to evaluate the performance of our method, we compare with fully unsupervised image translation methods: CycleGAN\cite{CycleGAN2017},  CUT\cite{park2020contrastive}, and ILVR \cite{choi2021ilvr}.

The corresponding qualitative results are shown in Fig.~\ref{fig:facesematic}. It is clear that CUT, CycleGAN, and ILVR produce unrealistic facial images with huge artifacts or create low-resolution faces. In contrast, our method always creates good-quality realistic faces. We present the quantitative results in Table~\ref{table:celeba_semantics}. The table shows that our method obtains the best FID scores of all methods and obtains the best mIOU score among the inference-time techniques.

\subsection{Face Super-Resolution}

\begin{figure}[tp]
    \centering
    \setlength{\tabcolsep}{1pt}
    {\small
    \renewcommand{\arraystretch}{0.5} 
    \begin{tabular}{c c c c c c}
    \captionsetup{type=figure, font=scriptsize}
    \raisebox{0.2in}{\rotatebox[origin=t]{90}{\scriptsize Grayscale}}&
    \includegraphics[width=0.19\linewidth]{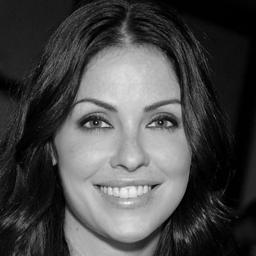}&
    \includegraphics[width=0.19\linewidth]{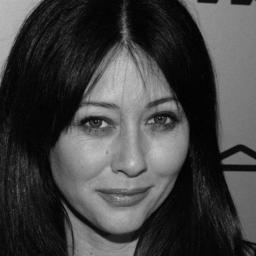}&
    \includegraphics[width=0.19\linewidth]{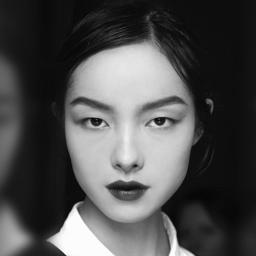}&
    \includegraphics[width=0.19\linewidth]{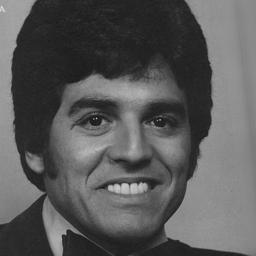}&
    \includegraphics[width=0.19\linewidth]{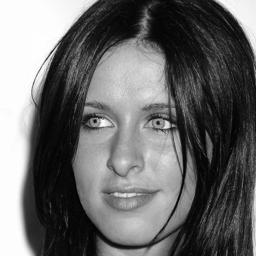}

    \tabularnewline
    \raisebox{0.2in}{\rotatebox[origin=t]{90}{\scriptsize ILVR}}&
    \includegraphics[width=0.19\linewidth]{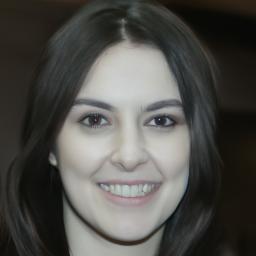}&
    \includegraphics[width=0.19\linewidth]{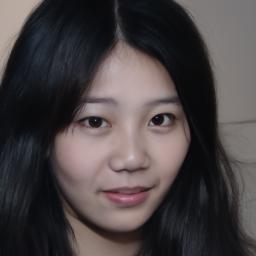}&
    \includegraphics[width=0.19\linewidth]{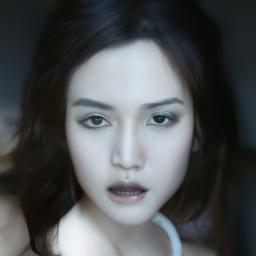}&
    \includegraphics[width=0.19\linewidth]{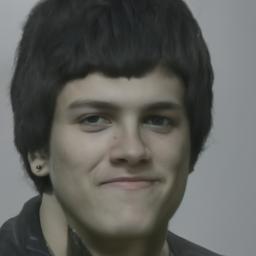}&
    \includegraphics[width=0.19\linewidth]{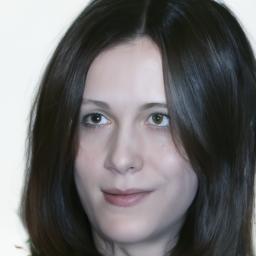}

    \tabularnewline
    \raisebox{0.2in}{\rotatebox[origin=t]{90}{\scriptsize OURS}}&
    \includegraphics[width=0.19\linewidth]{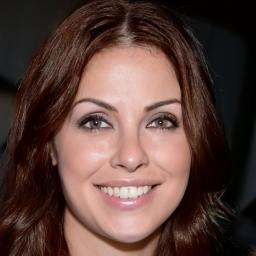}&
    \includegraphics[width=0.19\linewidth]{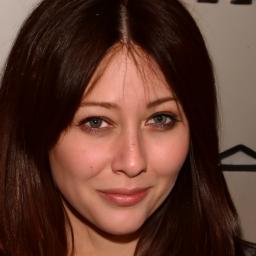}&
    \includegraphics[width=0.19\linewidth]{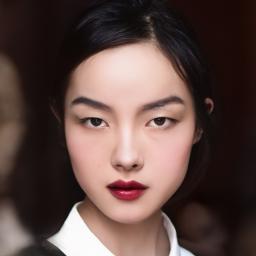}&
    \includegraphics[width=0.19\linewidth]{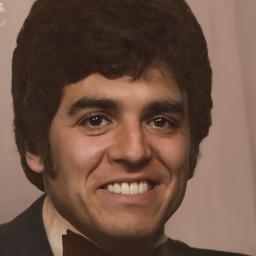}&
    \includegraphics[width=0.19\linewidth]{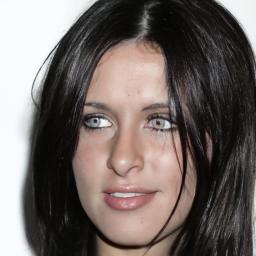}
    \tabularnewline
    \raisebox{0.3in}{\rotatebox[origin=t]{90}{\scriptsize OURS multi}}&
    \includegraphics[width=0.19\linewidth]{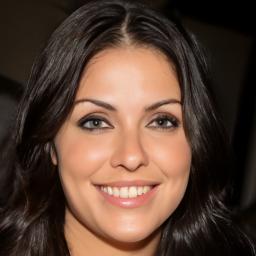}&
    \includegraphics[width=0.19\linewidth]{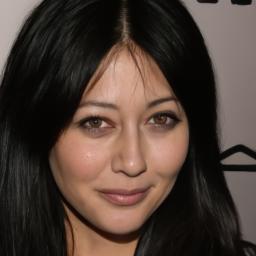}&
    \includegraphics[width=0.19\linewidth]{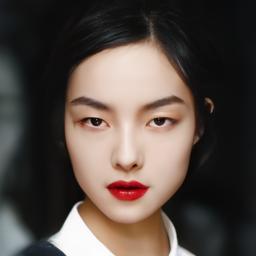}&
    \includegraphics[width=0.19\linewidth]{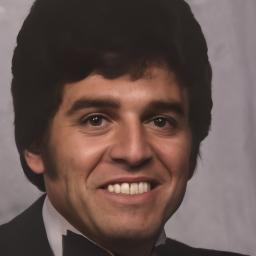}&
    \includegraphics[width=0.19\linewidth]{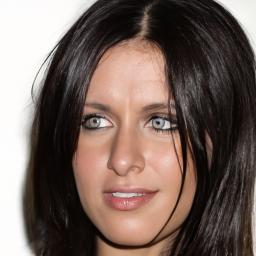}
\end{tabular}}
\caption{Qualitative comparisons for colorization. The row labeled ``OURS multi'' refers to the use of multi-step sampling, as described in Sec.~\ref{sec:multisample}.}
\label{fig:color}
\vspace{-.1in}
\end{figure}

\begin{figure*}[t!]
    \centering
    \begin{subfigure}[t]{0.10\linewidth}
      \captionsetup{justification=centering, labelformat=empty, font=scriptsize}
      \includegraphics[width=1\linewidth]{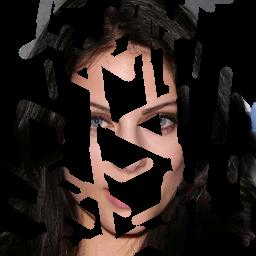}
      \includegraphics[width=1\linewidth]{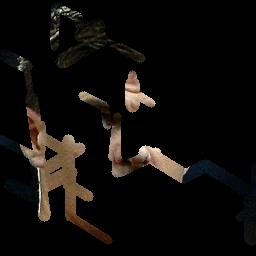}
      \includegraphics[width=1\linewidth]{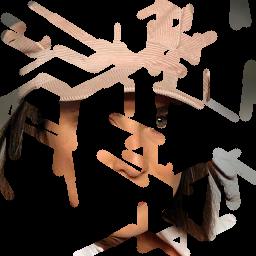}
      \caption{Degraded}
    \end{subfigure}
    \begin{subfigure}[t]{0.10\linewidth}
      \captionsetup{justification=centering, labelformat=empty, font=scriptsize}
      \includegraphics[width=1\linewidth]{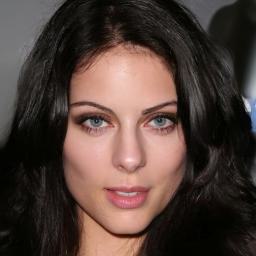}
      \includegraphics[width=1\linewidth]{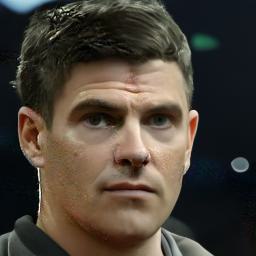}
      \includegraphics[width=1\linewidth]{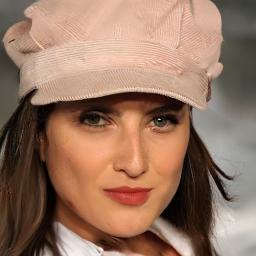}
      \caption{Reinpaint}
    \end{subfigure}
    \begin{subfigure}[t]{0.10\linewidth}
      \captionsetup{justification=centering, labelformat=empty, font=scriptsize}
      \includegraphics[width=1\linewidth]{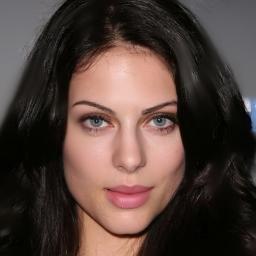}
      \includegraphics[width=1\linewidth]{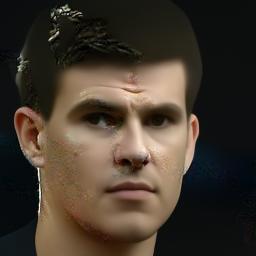}
      \includegraphics[width=1\linewidth]{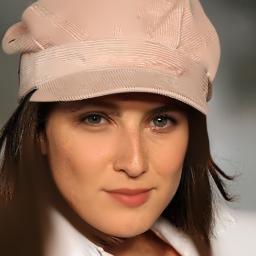}
      \caption{Ours}
    \end{subfigure}
    \hspace{2mm}
    \begin{subfigure}[t]{0.10\linewidth}
      \captionsetup{justification=centering, labelformat=empty, font=scriptsize}
      \includegraphics[width=1\linewidth]{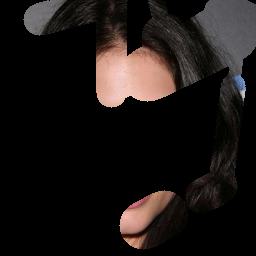}
      \includegraphics[width=1\linewidth]{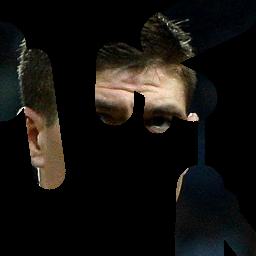}
      \includegraphics[width=1\linewidth]{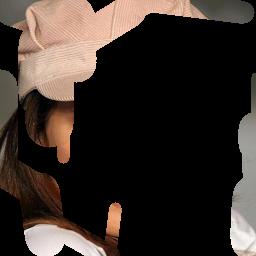}
      \caption{Degraded}
    \end{subfigure}
    \begin{subfigure}[t]{0.10\linewidth}
      \captionsetup{justification=centering, labelformat=empty, font=scriptsize}
      \includegraphics[width=1\linewidth]{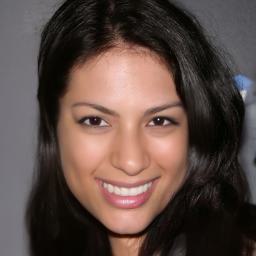}
      \includegraphics[width=1\linewidth]{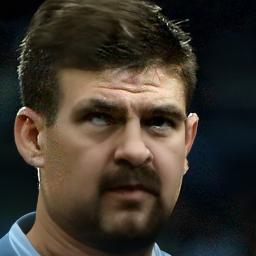}
      \includegraphics[width=1\linewidth]{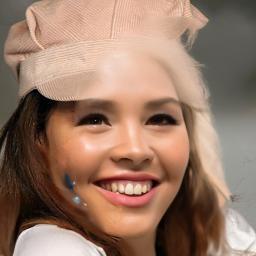}
      \caption{Reinpaint}
    \end{subfigure}
    \begin{subfigure}[t]{0.10\linewidth}
      \captionsetup{justification=centering, labelformat=empty, font=scriptsize}
      \includegraphics[width=1\linewidth]{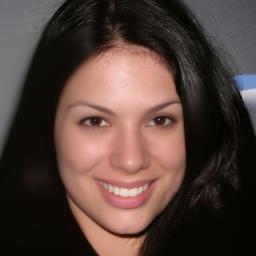}
      \includegraphics[width=1\linewidth]{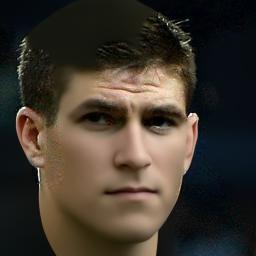}
      \includegraphics[width=1\linewidth]{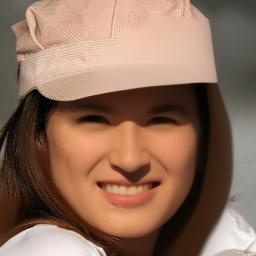}
      \caption{Ours}
    \end{subfigure}
    \hspace{2mm}
    \begin{subfigure}[t]{0.10\linewidth}
      \captionsetup{justification=centering, labelformat=empty, font=scriptsize}
      \includegraphics[width=1\linewidth]{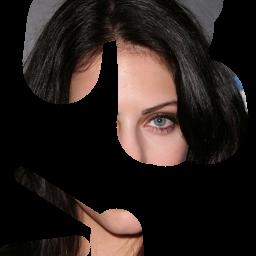}
      \includegraphics[width=1\linewidth]{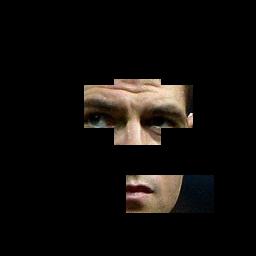}
      \includegraphics[width=1\linewidth]{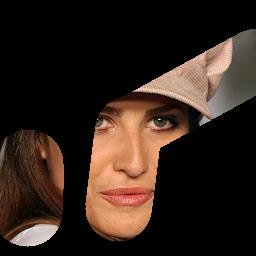}
      \caption{Degraded}
    \end{subfigure}
    \begin{subfigure}[t]{0.10\linewidth}
      \captionsetup{justification=centering, labelformat=empty, font=scriptsize}
      \includegraphics[width=1\linewidth]{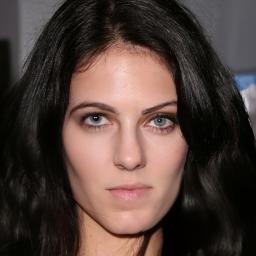}
      \includegraphics[width=1\linewidth]{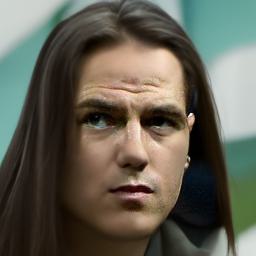}
      \includegraphics[width=1\linewidth]{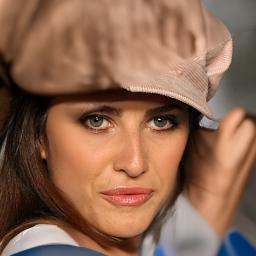}
      \caption{Reinpaint}
    \end{subfigure}
    \begin{subfigure}[t]{0.10\linewidth}
      \captionsetup{justification=centering, labelformat=empty, font=scriptsize}
      \includegraphics[width=1\linewidth]{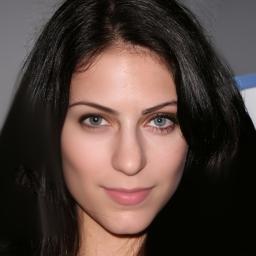}
      \includegraphics[width=1\linewidth]{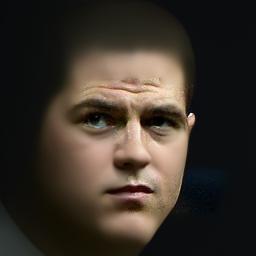}
      \includegraphics[width=1\linewidth]{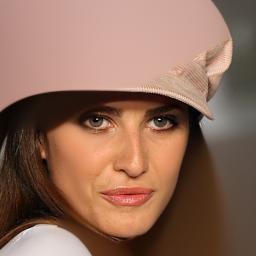}
      \caption{Ours}
    \end{subfigure}
    \vspace{-3mm}    
    \caption{Qualitative comparisons for inpainting for thin, medium and thick masks.}
    \label{fig:sup4inp}
    \vspace{-2mm}
  \end{figure*}

  \begin{figure*}[t!]
    \centering
    \begin{subfigure}[t]{0.135\linewidth}
      \captionsetup{justification=centering, labelformat=empty, font=scriptsize}
      \includegraphics[width=1\linewidth]{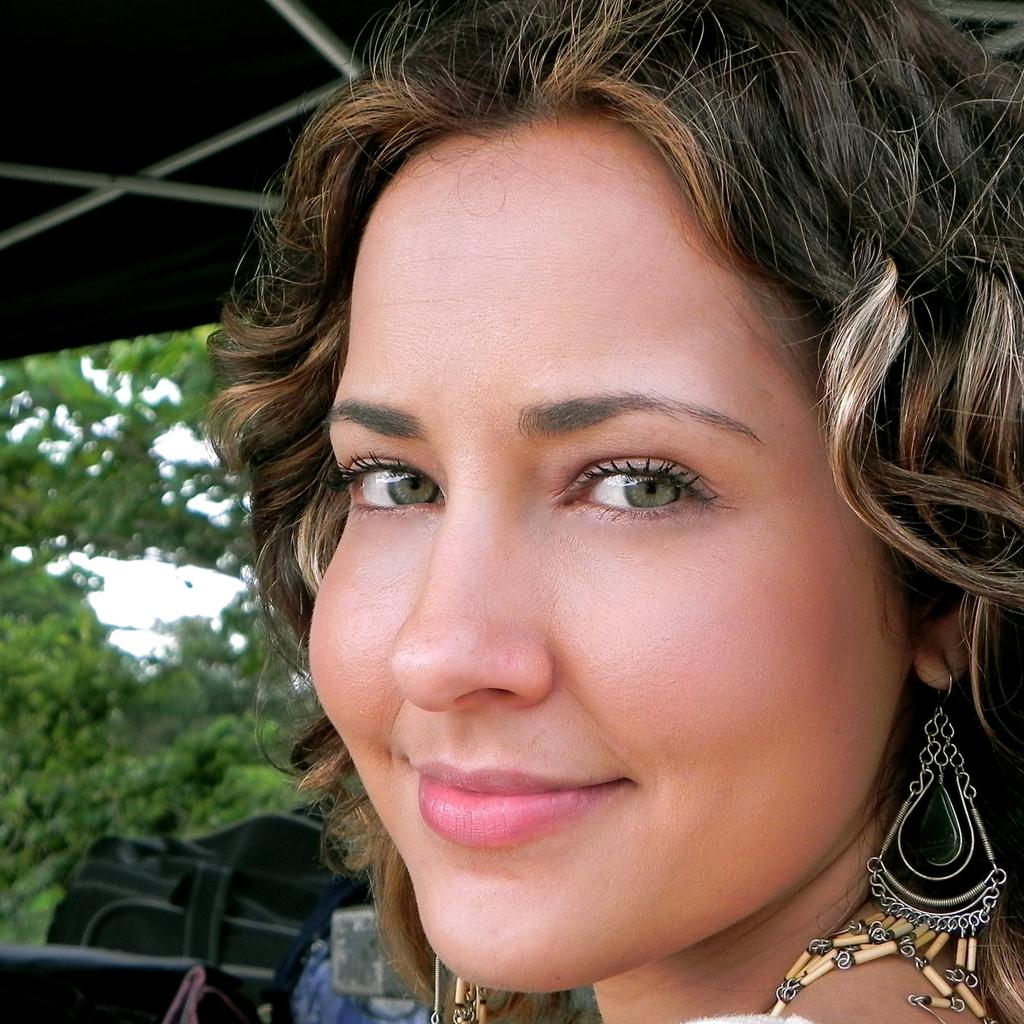}
      \caption{Original Image}
    \end{subfigure}
    \begin{subfigure}[t]{0.135\linewidth}
      \captionsetup{justification=centering, labelformat=empty, font=scriptsize}
      \includegraphics[width=1\linewidth]{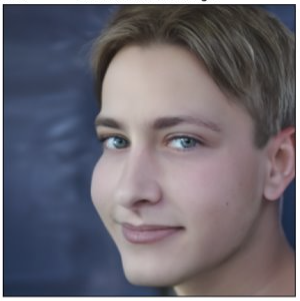}
      \caption{Photo of a young man}
    \end{subfigure}
        \begin{subfigure}[t]{0.135\linewidth}
      \captionsetup{justification=centering, labelformat=empty, font=scriptsize}
      \includegraphics[width=1\linewidth]{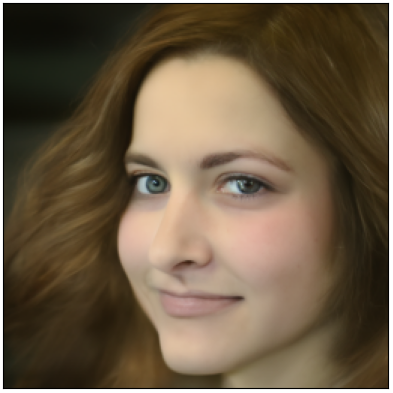}
      \caption{She has wavy hair}
    \end{subfigure}
       \begin{subfigure}[t]{0.135\linewidth}
      \captionsetup{justification=centering, labelformat=empty, font=scriptsize}
      \includegraphics[width=1\linewidth]{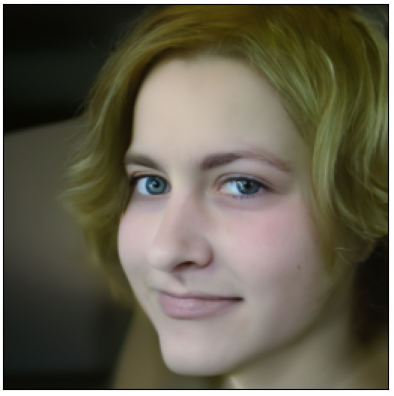}
      \caption{She has blonde hair}
    \end{subfigure}
       \begin{subfigure}[t]{0.135\linewidth}
      \captionsetup{justification=centering, labelformat=empty, font=scriptsize}
      \includegraphics[width=1\linewidth]{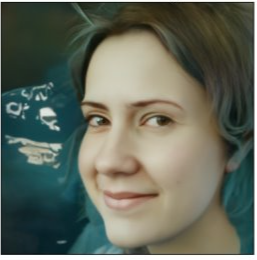}
      \caption{Photo of an old woman}
    \end{subfigure}
        \begin{subfigure}[t]{0.135\linewidth}
      \captionsetup{justification=centering, labelformat=empty, font=scriptsize}
      \includegraphics[width=1\linewidth]{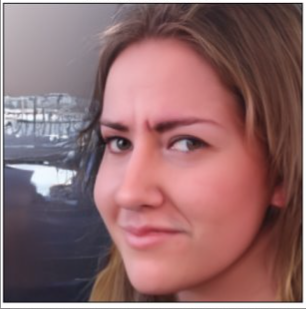}
      \caption{She is angry}
    \end{subfigure}
        \begin{subfigure}[t]{0.135\linewidth}
      \captionsetup{justification=centering, labelformat=empty, font=scriptsize}
      \includegraphics[width=1\linewidth]{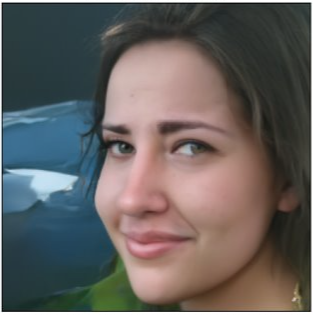}
      \caption{She is sad}
    \end{subfigure}   
    \caption{Qualitative comparisons for text-based image editing.}
    \label{fig:faceedit}
    \vspace{-2mm}
  \end{figure*}

We evaluate the performance for the face super-resolution task using the CelebA dataset~\cite{liu2015faceattributes}. As baselines, we utilize fully inference-time methods in which no task-specific training is used. As the first baseline, we choose PULSE\cite{menon2020pulse}, a self-supervised upsampling technique utilizing GANs. As the next comparison method, we choose ILVR\cite{choi2021ilvr} which, like our method, performs super-resolution utilizing an unconditional pre-trained diffusion model. However, in ILVR, sampling happens at timestep $t$ rather than at the implicit step in our algorithm. In total, we utilize 300 images for evaluations. We present some qualitative results in Fig.~\ref{fig:face_sr}. For ILVR\cite{choi2021ilvr}, we utilize 100 timesteps of sampling, the same as in our case. As we can see, PULSE\cite{menon2020pulse} and ILVR\cite{choi2021ilvr} are unable to restore the correct identity and also contain blur artifacts after restoration. On the other hand, steered diffusion (our method) is able to restore photorealistic facial images. The qualitative evaluations are presented in Table \ref{table:celeba_sr}; our method yields a 0.18 improvement in perceptual similarity, 6.95 dB improvement in PSNR, and 0.24 improvement in SSIM versus all of the other comparison methods.





\begin{table}[!t]

\begin{center}

\vspace{-2mm}
\scalebox{0.7}{
\begin{tabular}{ c | c c c |c c}
\toprule
  &\multicolumn{3}{c}{Trained Methods} & \multicolumn{2}{c}{Inference-Time Methods} \\	
\hline
Method &CUT\cite{isola2017image} & GCGAN\cite{FuCVPR19-GcGAN} &CycleGAN\cite{CycleGAN2017}&ILVR\cite{choi2021ilvr} & Ours \\
\hline
FID~$\textcolor{black}{(\downarrow)}$&49.34&58.80&52.70&107.46&41.38 \\
mIoU$\textcolor{black}{(\uparrow)}$&0.647&0.699&0.723&0.377&0.532\\
\bottomrule
\end{tabular}}
\end{center}
\vspace{-1.5em}
\caption{Quantitative results for semantic generation
\label{table:celeba_semantics}
}
\end{table}

\begin{table}[!t]
\begin{center}
\scalebox{0.8}{
\begin{tabular}{c c c c c c}
\toprule
   Method& FID~$\textcolor{black}{\downarrow}$ &  LPIPS$\textcolor{black}{\downarrow}$ &NIQE$\textcolor{black}{\downarrow}$ &PSNR$\textcolor{black}{\uparrow}$ &SSIM$\textcolor{black}{\uparrow}$        \\
\midrule
Bicubic  &130.3&0.4419& 12.03 & 23.85 & 0.642\\
ILVR\cite{choi2021ilvr}  & 62.24& 0.4164 & 7.38 & 20.54 & 0.5527\\
PULSE\cite{menon2020pulse} & 84.67& 0.4365 & 5.04 & 21.08 & 0.5285\\
Ours & 51.19& 0.2593 & 9.24 & 26.02 & 0.711 \\

\bottomrule
\end{tabular}}
\end{center}
\vspace{-5mm}
\caption{Quantitative results for super-resolution}
\label{table:celeba_sr}
\vspace{-3mm}
\end{table}

\subsection{Face Colorization}

As a baseline method, we modify ILVR\cite{choi2021ilvr} to suit the task of colorization. For this, rather than performing the constraint at every step, we start the sampling process from a noised grayscale image and enforce consistency between the generated and original grayscale images. In total, we utilize 300 images for evaluation. The corresponding results can be seen in Fig.~\ref{fig:color}. As we can see, our method is able to reconstruct photorealistic faces with naturalistic colours compared to ILVR\cite{choi2021ilvr}. The corresponding quantitative metrics are presented in Table.~\ref{table:celebacolor}. We get a significant boost in performance, with an FID score of $19$, LPIPS\cite{zhang2018unreasonable} score of $0.19$, and NIQE\cite{mittal2012making} score of $1.5$ for our method.

\begin{table}[!t]
\begin{center}
\scalebox{1}{
\begin{tabular}{c c c c }
\toprule
   Method & FID~$\textcolor{black}{\downarrow}$ &  LPIPS$\textcolor{black}{\downarrow}$ &NIQE$\textcolor{black}{\downarrow}$        \\
\midrule
Grayscale  &69.27&0.2781 & 5.07\\
ILVR\cite{choi2021ilvr}  &67.66&0.5270 & 7.54\\
Ours  &49.72&0.3311& 5.91 \\

\bottomrule
\end{tabular}}
\caption{Quantitative results for colorization}
\label{table:celebacolor}
\end{center}
\vspace{-5mm}
\end{table}

\begin{table*}[!t]
\begin{center}
\vspace{-2mm}
\scalebox{0.9}{
\begin{tabular}{c c c c c c c c c c c c c}
\toprule
\multirow{2}{*}{Method}&\multicolumn{4}{c}{Thin}&\multicolumn{4}{c}{Medium}&\multicolumn{4}{c}{Thick}\\
\cmidrule(lr){2-5} \cmidrule(lr){6-9} \cmidrule(lr){10-13}
   & FID~$\textcolor{black}{\downarrow}$ &  LPIPS$\textcolor{black}{\downarrow}$ &PSNR$\textcolor{black}{\uparrow}$ &SSIM$\textcolor{black}{\uparrow}$ &FID~$\textcolor{black}{\downarrow}$ &  LPIPS$\textcolor{black}{\downarrow}$ &PSNR$\textcolor{black}{\uparrow}$ &SSIM$\textcolor{black}{\uparrow}$ &FID~$\textcolor{black}{\downarrow}$ &  LPIPS$\textcolor{black}{\downarrow}$ &PSNR$\textcolor{black}{\uparrow}$ &SSIM$\textcolor{black}{\uparrow}$       \\

\midrule
Degraded  &371.03&0.676 & 7.847&0.201&258.86&0.624 & 7.527&0.228&231.93&0.585 & 7.606&0.249\\
Reinpaint\cite{lugmayr2022repaint} &43.35 &0.304&17.99&0.671& 53.71&0.399 & 13.71&0.558& 52.40&0.407 & 12.78&0.530\\
Ours  &30.85&0.183& 24.92&0.833&35.39&0.220& 21.55&0.786 &40.12&0.242& 18.87&0.705\\
\bottomrule
\end{tabular}}
\end{center}
\vspace{-1.5em}
\caption{Quantitative results for inpainting.}
\label{table:inpaint}
\vspace{-2mm}
\end{table*}

  \begin{figure*}[t!]
    \centering
        \begin{subfigure}[t]{0.130\linewidth}
      \captionsetup{justification=centering, labelformat=empty, font=scriptsize}
      \includegraphics[width=1\linewidth]{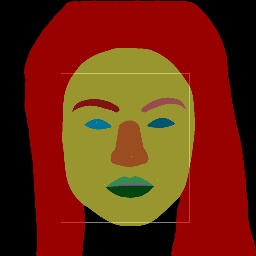}
      \caption{Semantic Map}
    \end{subfigure}
    \begin{subfigure}[t]{0.130\linewidth}
      \captionsetup{justification=centering, labelformat=empty, font=scriptsize}
      \includegraphics[width=1\linewidth]{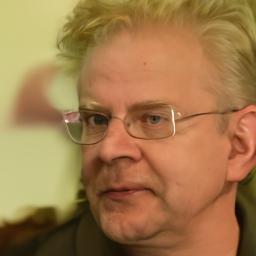}
      \caption{$K=0$}
    \end{subfigure}
    \begin{subfigure}[t]{0.130\linewidth}
      \captionsetup{justification=centering, labelformat=empty, font=scriptsize}
      \includegraphics[width=1\linewidth]{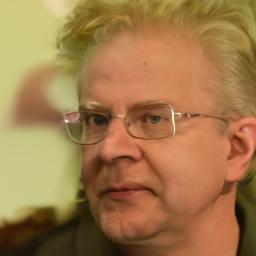}
      \caption{$K=200$}
    \end{subfigure}
    \begin{subfigure}[t]{0.130\linewidth}
      \captionsetup{justification=centering, labelformat=empty, font=scriptsize}
      \includegraphics[width=1\linewidth]{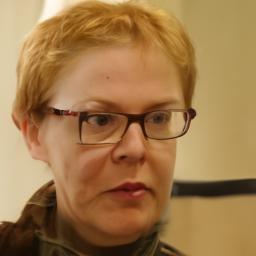}
      \caption{$K=2000$}
    \end{subfigure}
        \begin{subfigure}[t]{0.130\linewidth}
      \captionsetup{justification=centering, labelformat=empty, font=scriptsize}
      \includegraphics[width=1\linewidth]{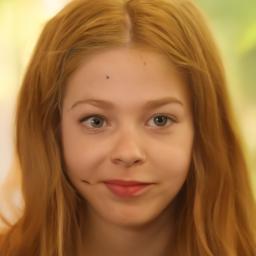}
      \caption{$K=20000$}
    \end{subfigure}
       \begin{subfigure}[t]{0.130\linewidth}
      \captionsetup{justification=centering, labelformat=empty, font=scriptsize}
      \includegraphics[width=1\linewidth]{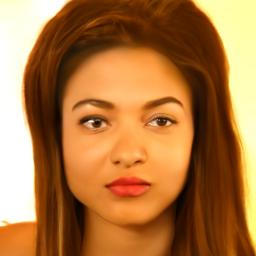}
    \caption{$K=2\times10^5$}    
    \end{subfigure}
        \begin{subfigure}[t]{0.130\linewidth}
      \captionsetup{justification=centering, labelformat=empty, font=scriptsize}
      \includegraphics[width=1\linewidth]{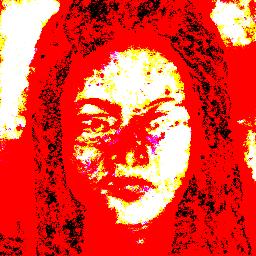}
    \caption{$K=2\times10^6$}    
    \end{subfigure}   
    \vspace{-3mm}    \caption{Figure showing sample variation with scaling factor $k(t) = K\sqrt{1-\bar{\alpha_t}}$}
    \label{fig:ablation}
    \vspace{-5mm}
  \end{figure*}

\begin{figure}[t!]
    \centering

      \begin{subfigure}[t]{0.23\linewidth}
      \captionsetup{justification=centering, labelformat=empty, font=scriptsize}
    \includegraphics[width=\linewidth]{results/colorization/grayscale/10014.jpg}
    \includegraphics[width=\linewidth]{results/colorization/grayscale/10013.jpg}
      \caption{Degraded}
    \end{subfigure}
    \begin{subfigure}[t]{0.23\linewidth}
      \captionsetup{justification=centering, labelformat=empty, font=scriptsize}
    \includegraphics[width=\linewidth]{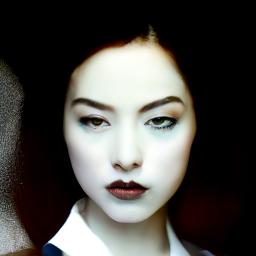}
    \includegraphics[width=\linewidth]{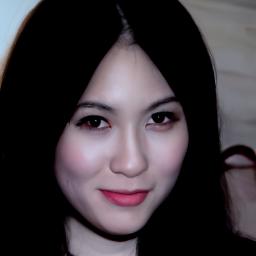}
      \caption{Noise robust}
    \end{subfigure}
    \begin{subfigure}[t]{0.23\linewidth}
      \captionsetup{justification=centering, labelformat=empty, font=scriptsize}
    \includegraphics[width=\linewidth]{results/colorization/ours_multi/10014_.jpg}
    \includegraphics[width=\linewidth]{results/colorization/ours_multi/10013_.jpg}
      \caption{OURS}
    \end{subfigure}
    \vspace{-3mm}    
    \caption{Qualitative comparisons for Colorization. }
    \label{fig:facenoiserobust}
    \vspace{-5mm}
  \end{figure}

\subsection{Inpainting, Image Editing, Identity replication}
Our method can also utilize multiple conditions simultaneously; we provide an illustration in Figure \ref{fig:faceedit} where we condition with an identity-preserving network and a text caption simultaneously. From the figure, one can see that our method can generalize well to a  diverse range of captions. For preserving identity, we used the VGGFace network\cite{Parkhi15}. We utilize FARL\cite{zheng2021farl}, which is pre-trained with face and corresponding text pairs to enforce the captions. For generic identity replication as in Figure \ref{fig:introfig} we use FARL face embedder.

For our image inpainting experiments, we use the subset released by \cite{suvorov2021resolution} and evaluate three different kinds of masks. Our method obtains better results than existing baselines across all mask variations. Qualitative and quantitative results are shown in Fig.~\ref{fig:sup4inp} and Table~\ref{table:inpaint}, respectively.

\subsection{Ablation Study}
\label{sec:ablation}
\begin{figure}[!tp]
    \centering
    \setlength{\tabcolsep}{1pt}
    {\small
    \renewcommand{\arraystretch}{0.5} 
    \begin{tabular}{c c c c c c}
    \captionsetup{type=figure, font=scriptsize}
    \raisebox{0.3in}{\rotatebox[origin=t]{90}{\scriptsize Semantic Map}}&
    \includegraphics[width=0.25\linewidth]{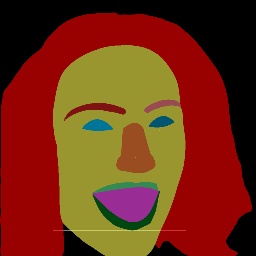}&
    \includegraphics[width=0.25\linewidth]{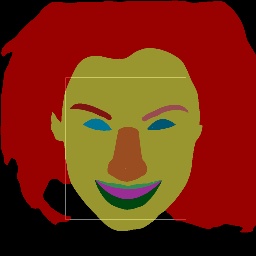}&
    \includegraphics[width=0.25\linewidth]{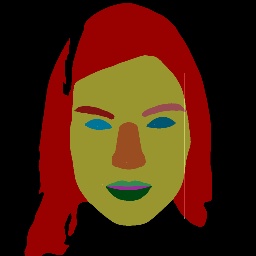}
    \tabularnewline
    \raisebox{0.2in}{\rotatebox[origin=t]{90}{\scriptsize constant}}&
    \includegraphics[width=0.25\linewidth]{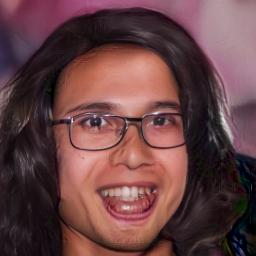}&
    \includegraphics[width=0.25\linewidth]{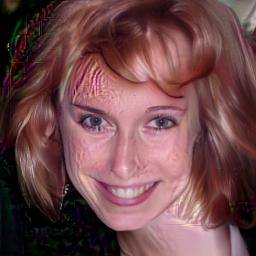}&
    \includegraphics[width=0.25\linewidth]{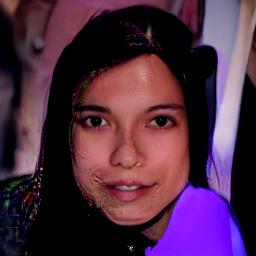}
    \tabularnewline
    \raisebox{0.2in}{\rotatebox[origin=t]{90}{\scriptsize varying}}&
    \includegraphics[width=0.25\linewidth]{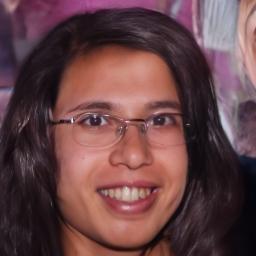}&
    \includegraphics[width=0.25\linewidth]{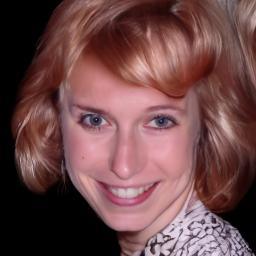}&
    \includegraphics[width=0.25\linewidth]{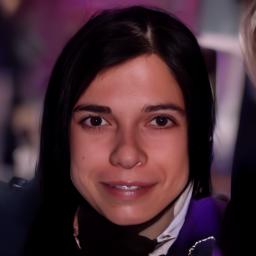}

\end{tabular}}
\vspace{-.1in}
\caption{A comparison of a time-varying scaling factor with non-time-varying.}
\label{fig:sup3}
\vspace{-6mm}
\end{figure}

\noindent\textbf{Effect of scaling factor $k(t)$ for semantic generation:} In this section, we analyze how the scaling factor affects the quality of the sample in the case of complex conditioning of semantic generation. Fig.~\ref{fig:ablation} shows the variation of sample quality starting from the same initial noise with different scaling factors $k(t) = K\sqrt{1-\bar{\alpha_t}}$. The sample quality is bad for very low scaling factors, and for very high scaling factors, the diffusion process escapes the manifold of natural face images. We show the variation in sample quality for a fixed scaling factor versus a time-varying scaling factor in Fig.~\ref{fig:sup3} , which demonstrates that a time-varying scale factor produces more realistic samples. This is because the effective variance of the noise scheduling, which effectively controls the amount of regularization possible at a particular timestep, reduces as the generation process proceeds. Hence a larger tweak is permissible at the early steps of diffusion, and only very small tweaks are permitted in the later steps.

\noindent\textbf{Noise-Robust classifier:} To validate the claims in section \ref{sec:energyfn}, we train a noise-robust inverse mapper for the task of colorization and show the output of the noise-robust classifier and the diffusion outputs for different noise levels in Fig.~\ref{fig:facenoiserobust}. The noise-robust classifier fails to preserve key details that our approach preserves.

\noindent\textbf{Limitations}
Although our method can generalize to a wide series of tasks, one limitation that persists is the value of the scaling factor $k(t)$. The value of $k(t)$ has to be empirically found based on the task, but once a few images are used to tune the value of $k(t)$, the model generalizes well to other conditioning images of the same task. Like any other conditional generation models that can perform image editing, our method also has societal impacts, and care must be taken in applying these methods.

\vspace{-2mm}
\section{Conclusion}
In this paper, we propose the first framework for plug-and-play conditional generation that can generalize well to both image-to-image translation tasks and label-based generation tasks. For this, we use the energy-based formulation of diffusion models and modulate the inference process using a task-specific predefined network or other preexisting function. Furthermore, we introduce a novel implicit sampling-based technique that improves the sampling quality across multiple tasks. We performed experiments on various tasks to show that our method can generalize across multiple tasks and outperforms existing methods that do not require additional training.
{\small
\bibliographystyle{ieee_fullname}
\bibliography{egbib}
}
\clearpage

\begin{strip} 
\section*{\centering Steered Diffusion: A Generalized Framework for Plug-and-Play\\ Conditional Image Synthesis } 
\section*{\centering Supplementary Material} 

\end{strip}
\begin{appendices}

\section*{A1. Visualization of diffusion steering process}
\begin{figure*}[tb!]
    \centering
    \setlength{\tabcolsep}{0.5pt}
    {\small
    \renewcommand{\arraystretch}{0.5} 
\begin{tabular}{c c c c c c c c c c c c }
    \captionsetup{type=figure, font=scriptsize}
  \includegraphics[width=0.095\linewidth]{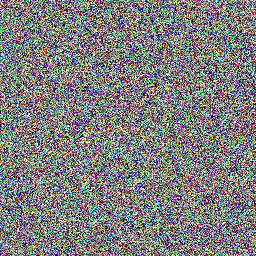}
  \includegraphics[width=0.095\linewidth]{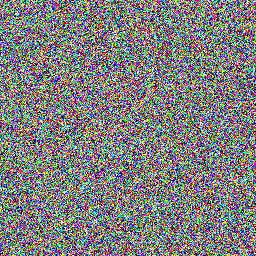}
  \includegraphics[width=0.095\linewidth]{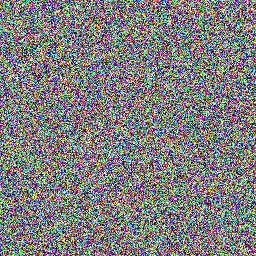}
  \includegraphics[width=0.095\linewidth]{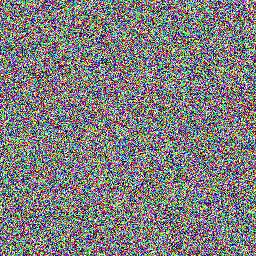}
  \includegraphics[width=0.095\linewidth]{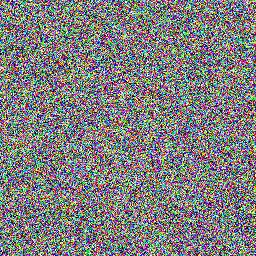}
  \hspace{5mm}
  \includegraphics[width=0.095\linewidth]{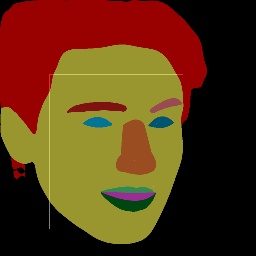}
  \includegraphics[width=0.095\linewidth]{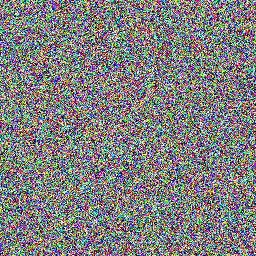}
  \includegraphics[width=0.095\linewidth]{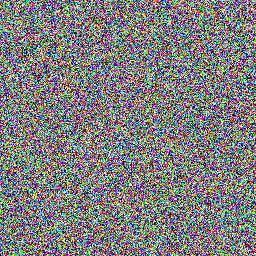}
  \includegraphics[width=0.095\linewidth]{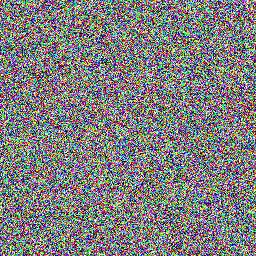}
  \includegraphics[width=0.095\linewidth]{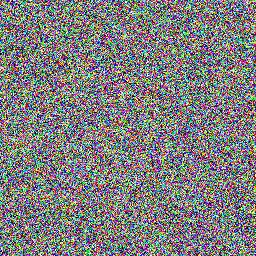}

 \tabularnewline
  \includegraphics[width=0.095\linewidth]{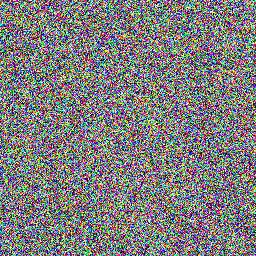}
  \includegraphics[width=0.095\linewidth]{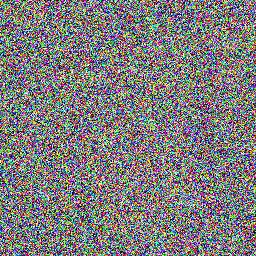}
  \includegraphics[width=0.095\linewidth]{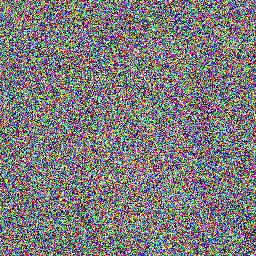}
  \includegraphics[width=0.095\linewidth]{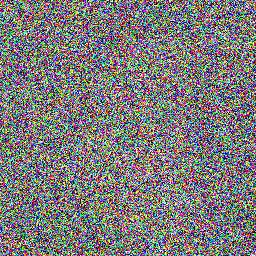}
  \includegraphics[width=0.095\linewidth]{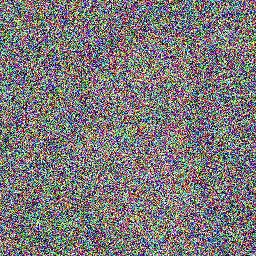}
  \hspace{5mm}
  \includegraphics[width=0.095\linewidth]{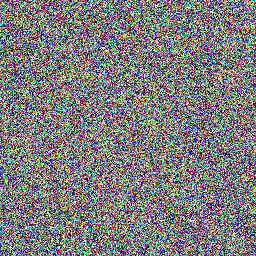}
  \includegraphics[width=0.095\linewidth]{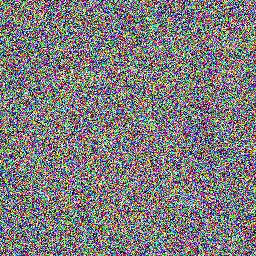}
  \includegraphics[width=0.095\linewidth]{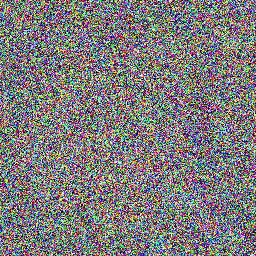}
  \includegraphics[width=0.095\linewidth]{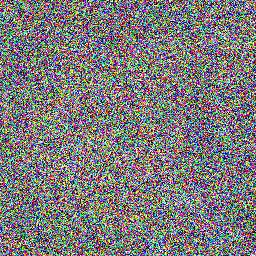}
  \includegraphics[width=0.095\linewidth]{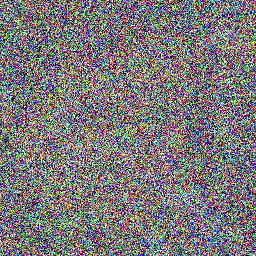}

\tabularnewline
\end{tabular}}
\vspace{-0.8cm}
\hspace{20pt}\captionof{figure}{Evolution of sampling process without and with guidance}
\label{fig:sup11}
\vspace{-8mm}
\end{figure*}

\begin{figure*}[tb!]
    \centering
    \setlength{\tabcolsep}{0.5pt}
    {\small
    \renewcommand{\arraystretch}{0.5} 
    \begin{tabular}{c c c c c c c c c c c c }
    \captionsetup{type=figure, font=scriptsize}
  \includegraphics[width=0.095\linewidth]{supplementary/True9_3/0.jpg}
  \includegraphics[width=0.095\linewidth]{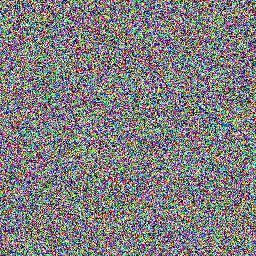}
  \includegraphics[width=0.095\linewidth]{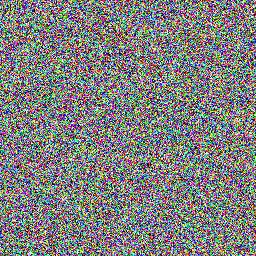}
  \includegraphics[width=0.095\linewidth]{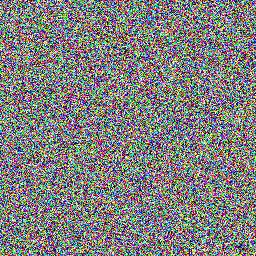}
  \includegraphics[width=0.095\linewidth]{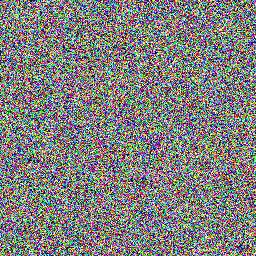}
  \hspace{5mm}
  \includegraphics[width=0.095\linewidth]{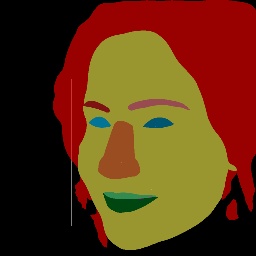}
  \includegraphics[width=0.095\linewidth]{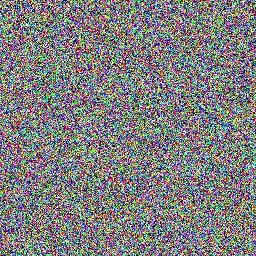}
  \includegraphics[width=0.095\linewidth]{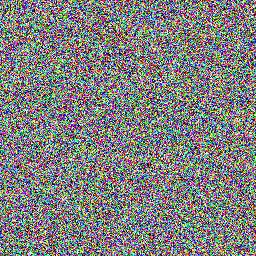}
  \includegraphics[width=0.095\linewidth]{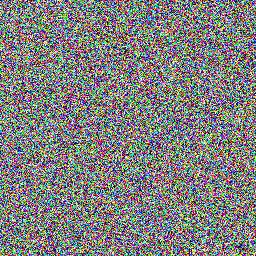}
  \includegraphics[width=0.095\linewidth]{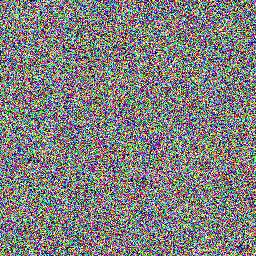}

 \tabularnewline
  \includegraphics[width=0.095\linewidth]{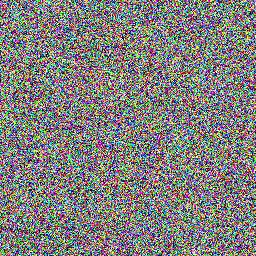}
  \includegraphics[width=0.095\linewidth]{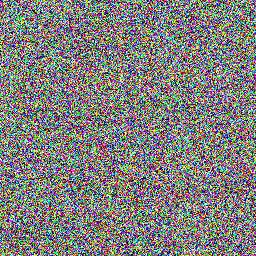}
  \includegraphics[width=0.095\linewidth]{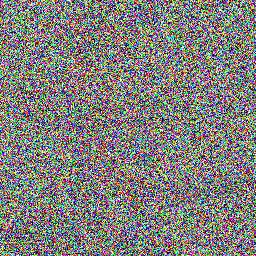}
  \includegraphics[width=0.095\linewidth]{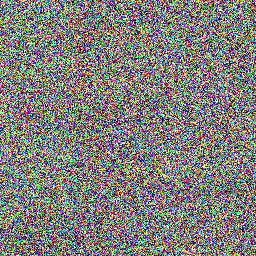}
  \includegraphics[width=0.095\linewidth]{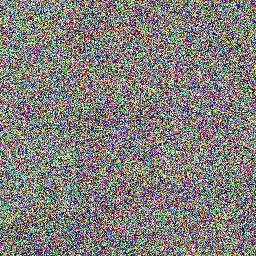}
  \hspace{5mm}
  \includegraphics[width=0.095\linewidth]{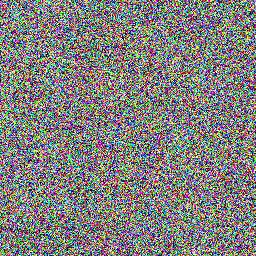}
  \includegraphics[width=0.095\linewidth]{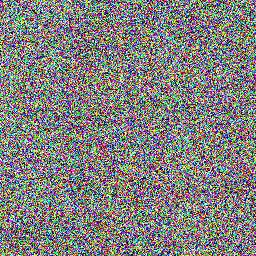}
  \includegraphics[width=0.095\linewidth]{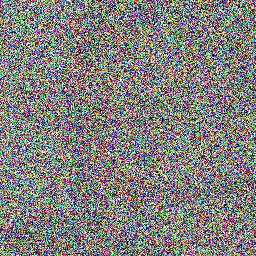}
  \includegraphics[width=0.095\linewidth]{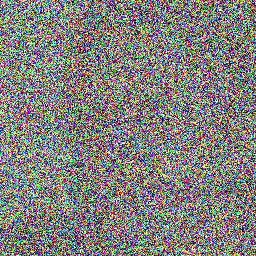}
  \includegraphics[width=0.095\linewidth]{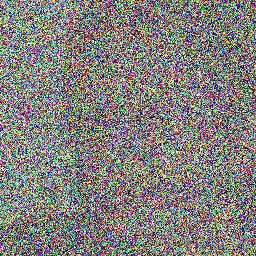}

\tabularnewline
\end{tabular}}
\vspace{-0.8cm}
\hspace{20pt}\captionof{figure}{Evolution of sampling process without and with guidance}
\label{fig:sup22}
\vspace{-2mm}
\end{figure*}

We present a visualization of intermediate outputs of the diffusion steering process in Figure~\ref{fig:sup11} and Figure~\ref{fig:sup22}. The left image denotes the case without the steering loss and the right denotes the case with a steering loss. As we can see, with the presence of the steering loss, the generated images are consistent with the semantic maps from an early stage and the results continue to improve. The consistency grows stronger as timesteps progress.

\section*{A2. Illustrating sample diversity using our method}
We present non-cherry-picked results for various conditional generation tasks to demonstrate our method's photorealistic image generation quality and the diversity of examples generated by our method. Here we use the same noise levels across different examples, and the generation condition is shown in the first image in each sequence of images.

\begin{figure*}[t!]
    \centering
    \begin{subfigure}[t]{0.105\linewidth}

      \captionsetup{justification=centering, labelformat=empty, font=scriptsize}

      \includegraphics[width=1\linewidth]{results/results_semantics/labels/1.jpg}
      \includegraphics[width=1\linewidth]{results/results_semantics/labels/103.jpg}
      \includegraphics[width=1\linewidth]{results/results_semantics/labels/1038.jpg}
      \caption{Labels}
    \end{subfigure}
    \begin{subfigure}[t]{0.105\linewidth}
      \captionsetup{justification=centering, labelformat=empty, font=scriptsize}

      \includegraphics[width=1\linewidth]{results/results_semantics/cyclegan/1.jpg}
      \includegraphics[width=1\linewidth]{results/results_semantics/cyclegan/103.jpg}
      \includegraphics[width=1\linewidth]{results/results_semantics/cyclegan/1038.jpg}

      \caption{CycleGAN}
    \end{subfigure}
    \begin{subfigure}[t]{0.105\linewidth}
      \captionsetup{justification=centering, labelformat=empty, font=scriptsize}

      \includegraphics[width=1\linewidth]{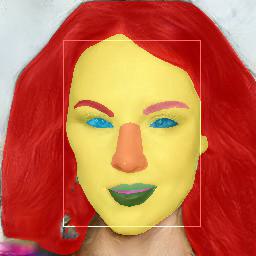}
      \includegraphics[width=1\linewidth]{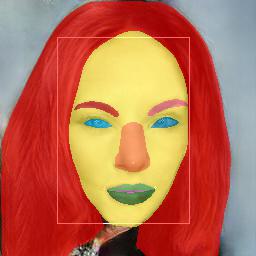}
      \includegraphics[width=1\linewidth]{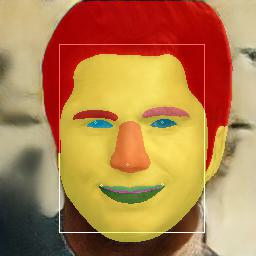}

      \caption{CycleGAN}
    \end{subfigure}
    \begin{subfigure}[t]{0.105\linewidth}
      \captionsetup{justification=centering, labelformat=empty, font=scriptsize}
      \includegraphics[width=1\linewidth]{results/results_semantics/cut/1.png}
      \includegraphics[width=1\linewidth]{results/results_semantics/cut/103.png}
      \includegraphics[width=1\linewidth]{results/results_semantics/cut/1038.png}

      \caption{CUT}
    \end{subfigure}
    \begin{subfigure}[t]{0.105\linewidth}
      \captionsetup{justification=centering, labelformat=empty, font=scriptsize}
      \includegraphics[width=1\linewidth]{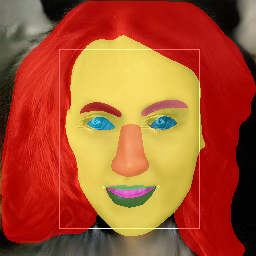}
      \includegraphics[width=1\linewidth]{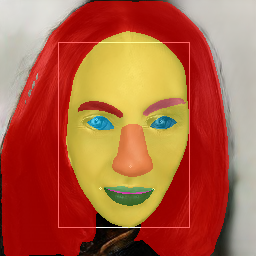}
      \includegraphics[width=1\linewidth]{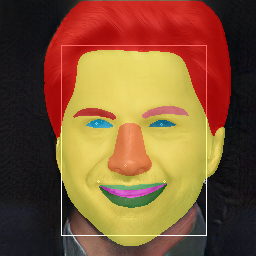}

      \caption{CUT}
    \end{subfigure}
    \begin{subfigure}[t]{0.105\linewidth}
      \captionsetup{justification=centering, labelformat=empty, font=scriptsize}
   
      \includegraphics[width=1\linewidth]{results/results_semantics/ilvr/1.jpg}
      \includegraphics[width=1\linewidth]{results/results_semantics/ilvr/103.jpg}
      \includegraphics[width=1\linewidth]{results/results_semantics/ilvr/1038.jpg}
      \caption{ILVR}
    \end{subfigure}
    \begin{subfigure}[t]{0.105\linewidth}
      \captionsetup{justification=centering, labelformat=empty, font=scriptsize}
   
      \includegraphics[width=1\linewidth]{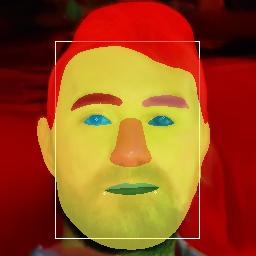}
      \includegraphics[width=1\linewidth]{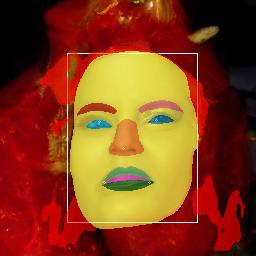}
      \includegraphics[width=1\linewidth]{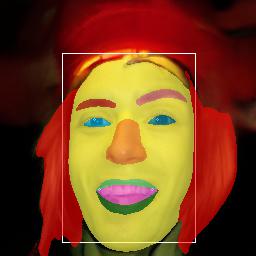}
      \caption{ILVR}
    \end{subfigure}
        \begin{subfigure}[t]{0.105\linewidth}
      \captionsetup{justification=centering, labelformat=empty, font=scriptsize}
   
      \includegraphics[width=1\linewidth]{results/results_semantics/ours_new/1.jpg}
      \includegraphics[width=1\linewidth]{results/results_semantics/ours_new/1_103.jpg}
      \includegraphics[width=1\linewidth]{results/results_semantics/ours_new/2_1038.jpg}
      \caption{OURS}
    \end{subfigure}
        \begin{subfigure}[t]{0.105\linewidth}
      \captionsetup{justification=centering, labelformat=empty, font=scriptsize}
      \includegraphics[width=1\linewidth]{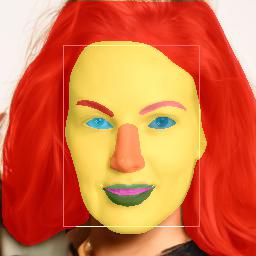}
      \includegraphics[width=1\linewidth]{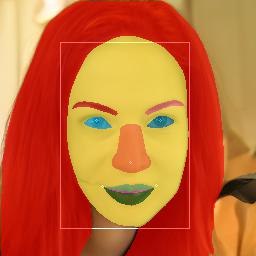}
      \includegraphics[width=1\linewidth]{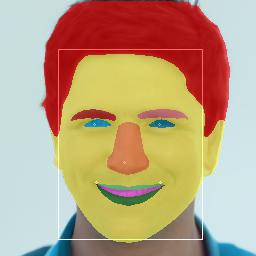}
      \caption{OURS}
    \end{subfigure}
    \vspace{-3mm}    
    \caption{Qualitative comparisons for segmentation labels from Figure 5 from original paper. }
    \label{fig:facesematicorig}
  \end{figure*}

\begin{figure*}[tb!]
    \centering
    \setlength{\tabcolsep}{0.5pt}
    {\small
    \renewcommand{\arraystretch}{0.5} 
    \begin{tabular}{c c c c c c c c c c c c }
    \captionsetup{type=figure, font=scriptsize}
  \includegraphics[width=0.095\linewidth]{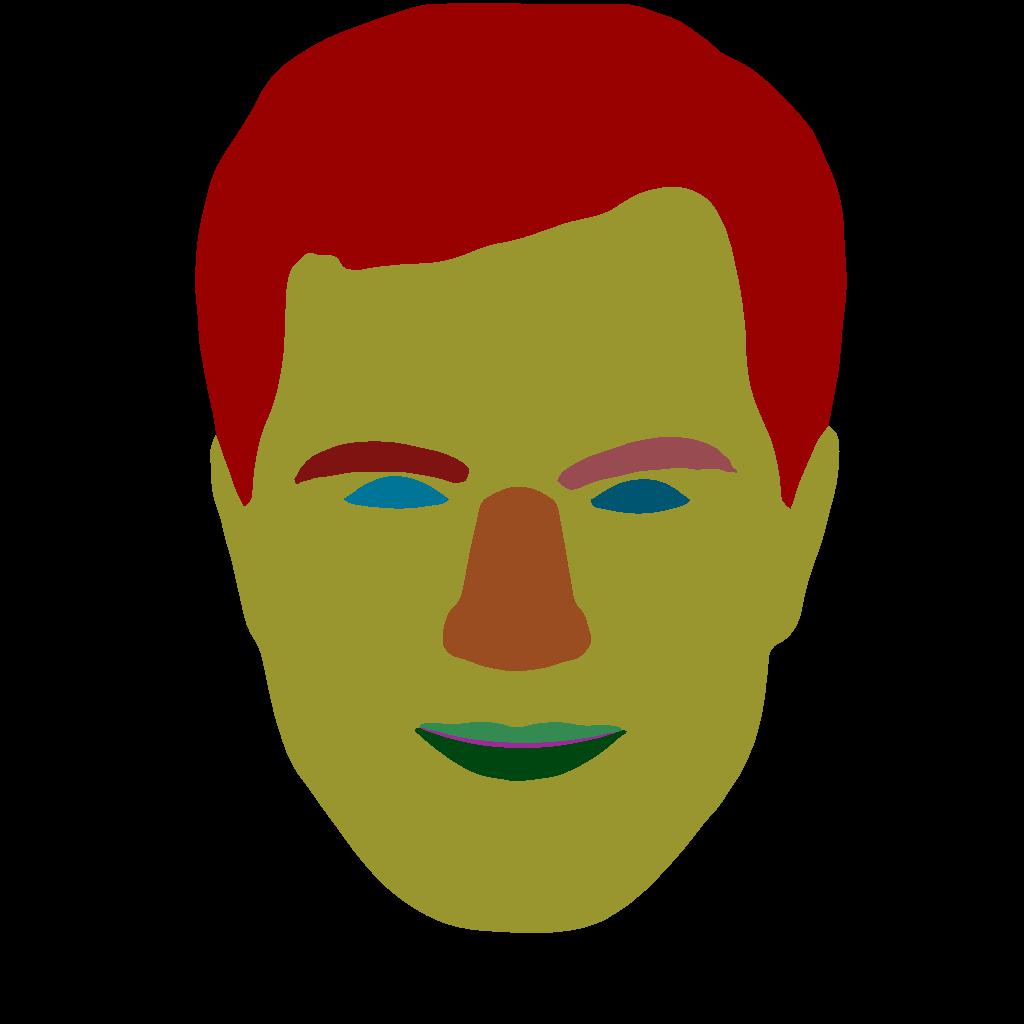}
  \includegraphics[width=0.095\linewidth]{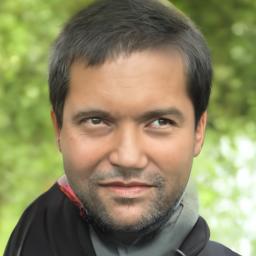}
  \includegraphics[width=0.095\linewidth]{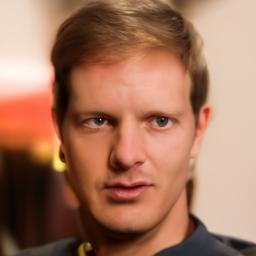}
  \includegraphics[width=0.095\linewidth]{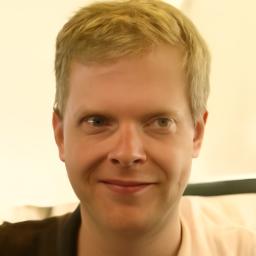}
  \includegraphics[width=0.095\linewidth]{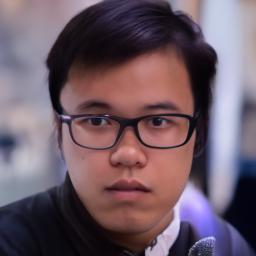}
    \hspace{5mm}
      \includegraphics[width=0.095\linewidth]{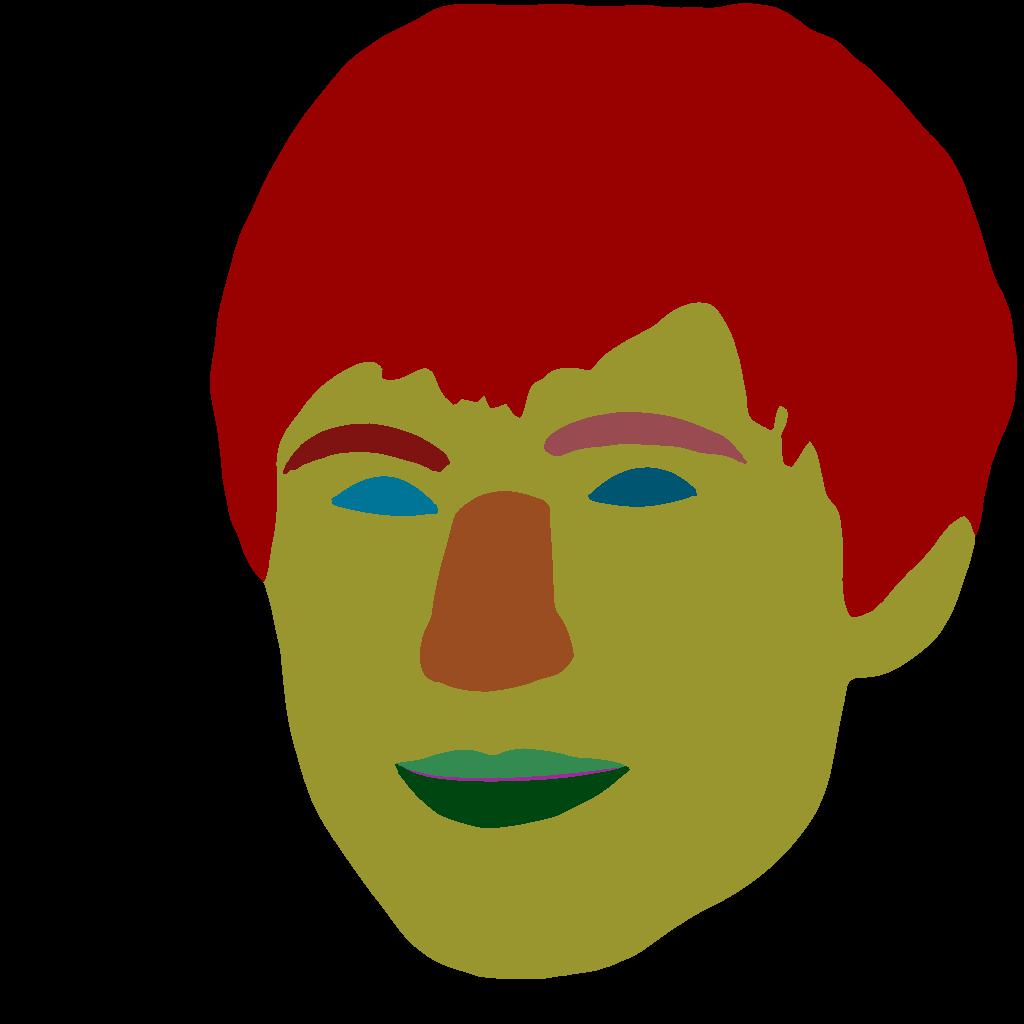}
  \includegraphics[width=0.095\linewidth]{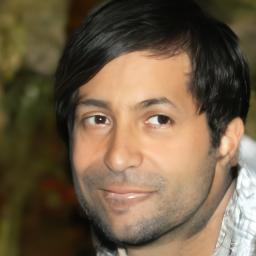}
  \includegraphics[width=0.095\linewidth]{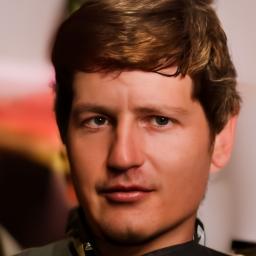}
  \includegraphics[width=0.095\linewidth]{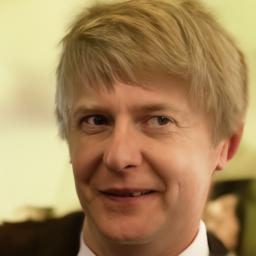}
  \includegraphics[width=0.095\linewidth]{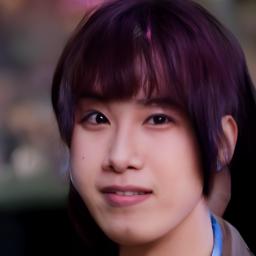}

 \tabularnewline
  \includegraphics[width=0.095\linewidth]{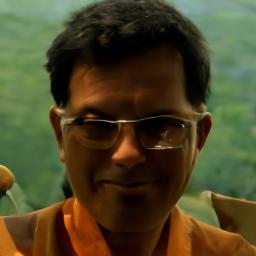}
  \includegraphics[width=0.095\linewidth]{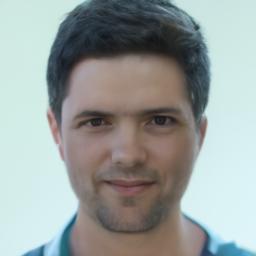}
  \includegraphics[width=0.095\linewidth]{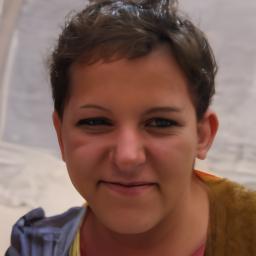}
  \includegraphics[width=0.095\linewidth]{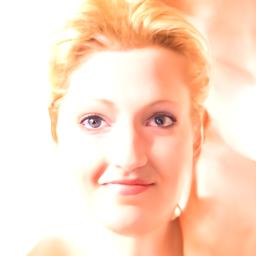}
  \includegraphics[width=0.095\linewidth]{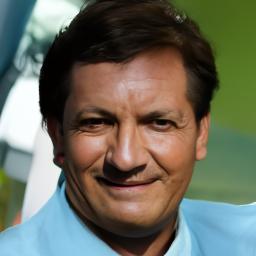}
    \hspace{5mm}
  \includegraphics[width=0.095\linewidth]{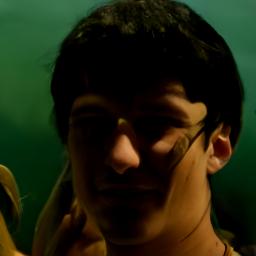}
  \includegraphics[width=0.095\linewidth]{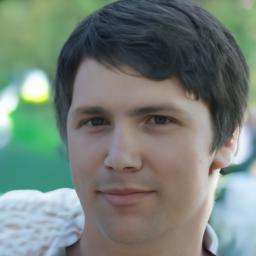}
  \includegraphics[width=0.095\linewidth]{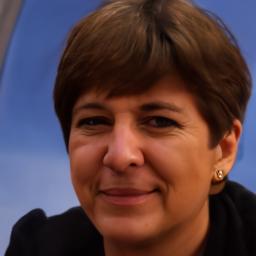}
  \includegraphics[width=0.095\linewidth]{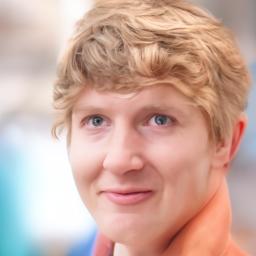}
  \includegraphics[width=0.095\linewidth]{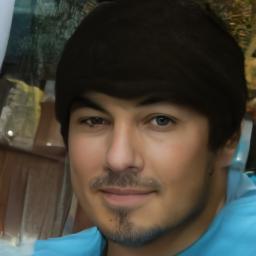}
\tabularnewline
\end{tabular}}
\vspace{-0.8cm}
\hspace{20pt}\captionof{figure}{Non-cherry-picked samples from our method corresponding to the same semantic map. The upper left figure shows the corresponding semantic map. Samples across different examples with same sampling location has same random seeds }
\label{fig:sup1v}
\vspace{-8mm}
\end{figure*}%
\begin{figure*}[tb!]
    \centering
    \setlength{\tabcolsep}{0.5pt}
    {\small
    \renewcommand{\arraystretch}{0.5} 
    \begin{tabular}{c c c c c c c c c c c c }
    \captionsetup{type=figure, font=scriptsize}
  \includegraphics[width=0.095\linewidth]{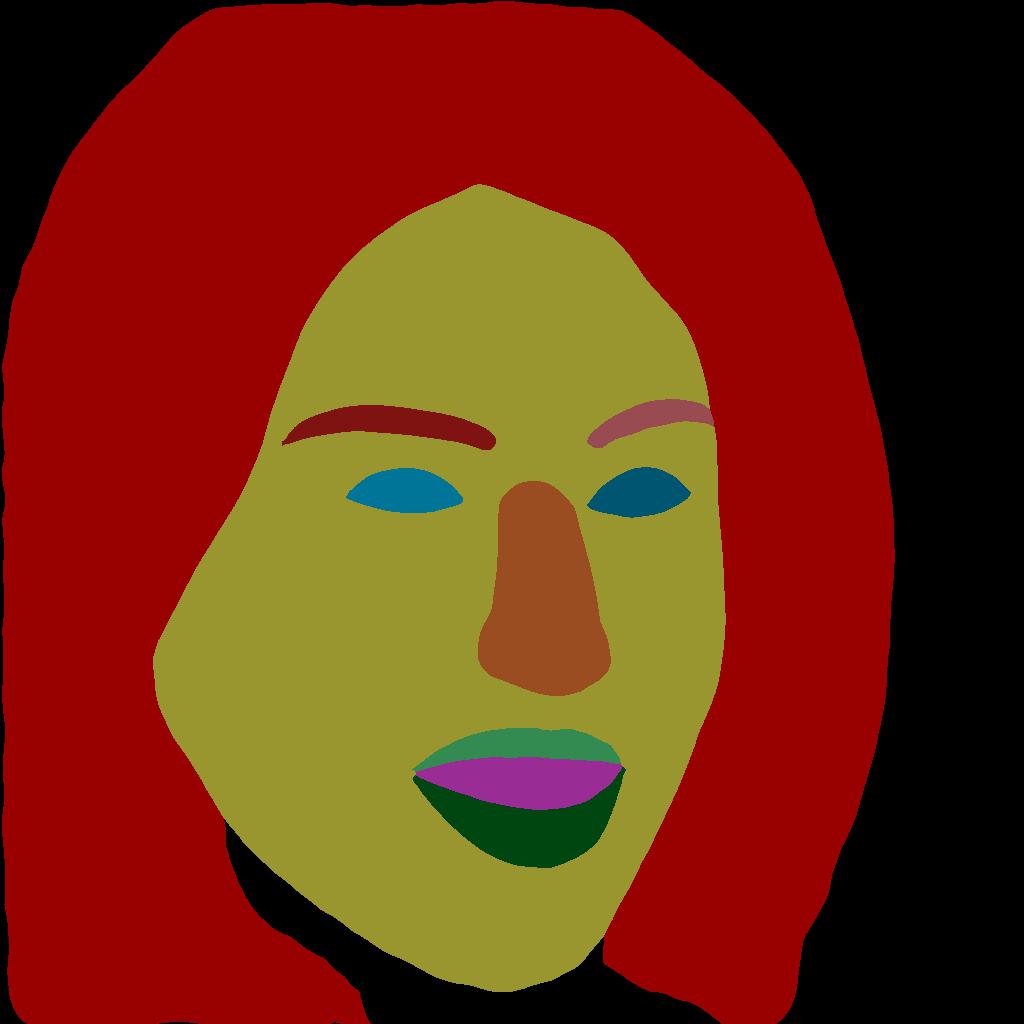}
  \includegraphics[width=0.095\linewidth]{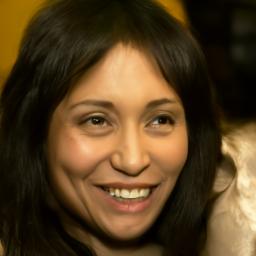}
  \includegraphics[width=0.095\linewidth]{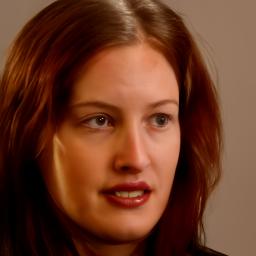}
  \includegraphics[width=0.095\linewidth]{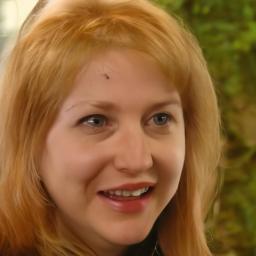}
  \includegraphics[width=0.095\linewidth]{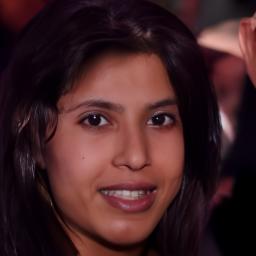}
    \hspace{5mm}
      \includegraphics[width=0.095\linewidth]{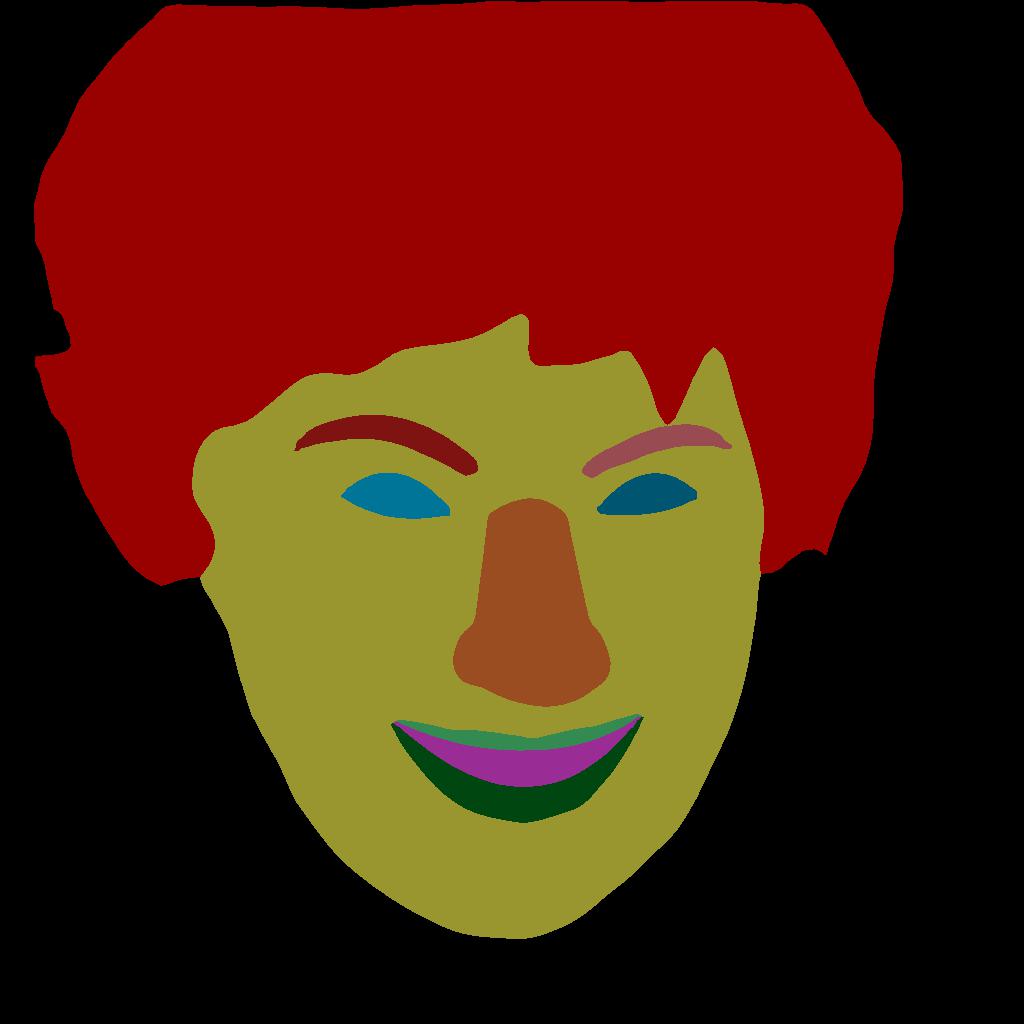}
  \includegraphics[width=0.095\linewidth]{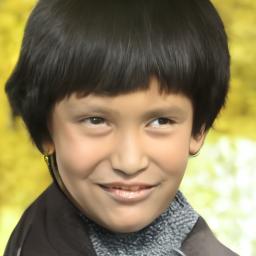}
  \includegraphics[width=0.095\linewidth]{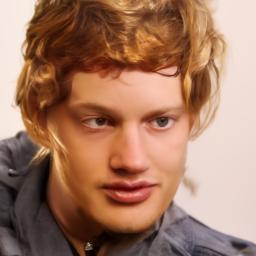}
  \includegraphics[width=0.095\linewidth]{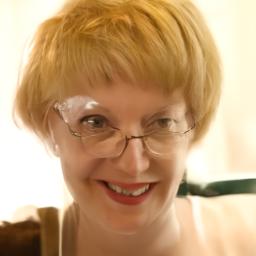}
  \includegraphics[width=0.095\linewidth]{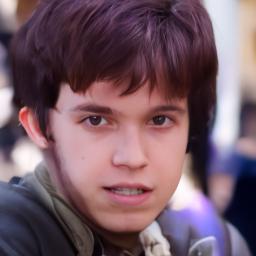}

 \tabularnewline
  \includegraphics[width=0.095\linewidth]{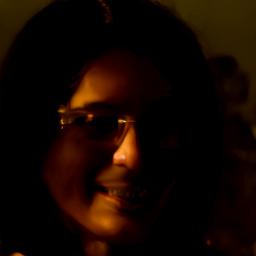}
  \includegraphics[width=0.095\linewidth]{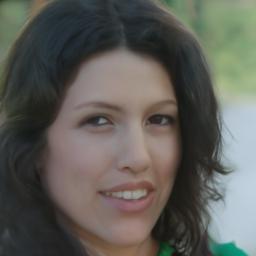}
  \includegraphics[width=0.095\linewidth]{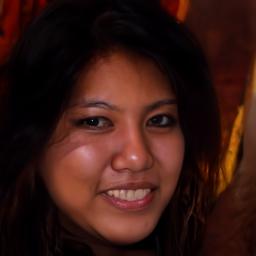}
  \includegraphics[width=0.095\linewidth]{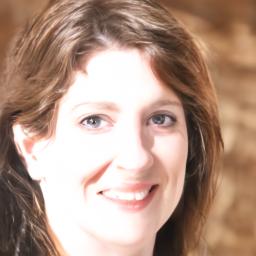}
  \includegraphics[width=0.095\linewidth]{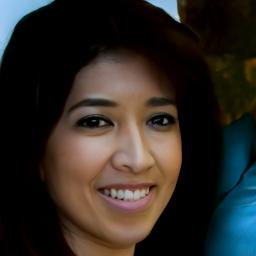}
    \hspace{5mm}
  \includegraphics[width=0.095\linewidth]{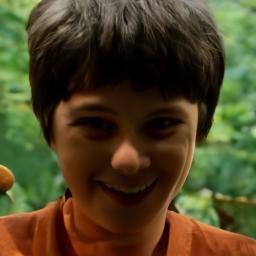}
  \includegraphics[width=0.095\linewidth]{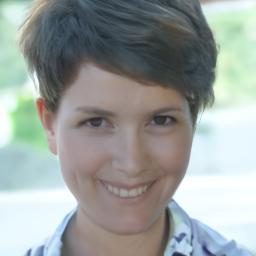}
  \includegraphics[width=0.095\linewidth]{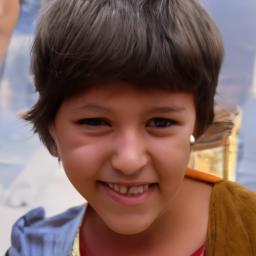}
  \includegraphics[width=0.095\linewidth]{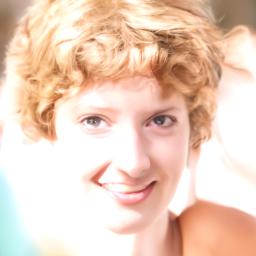}
  \includegraphics[width=0.095\linewidth]{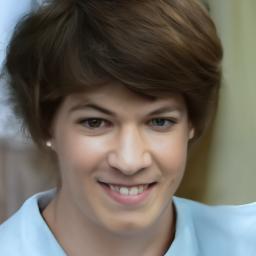}
\tabularnewline
\end{tabular}}
\vspace{-0.8cm}
\label{fig:sup1}
\vspace{-8mm}
\end{figure*}%
\begin{figure*}[tb!]
    \centering
    \setlength{\tabcolsep}{0.5pt}
    {\small
    \renewcommand{\arraystretch}{0.5} 
    \begin{tabular}{c c c c c c c c c c c c }
    \captionsetup{type=figure, font=scriptsize}
  \includegraphics[width=0.095\linewidth]{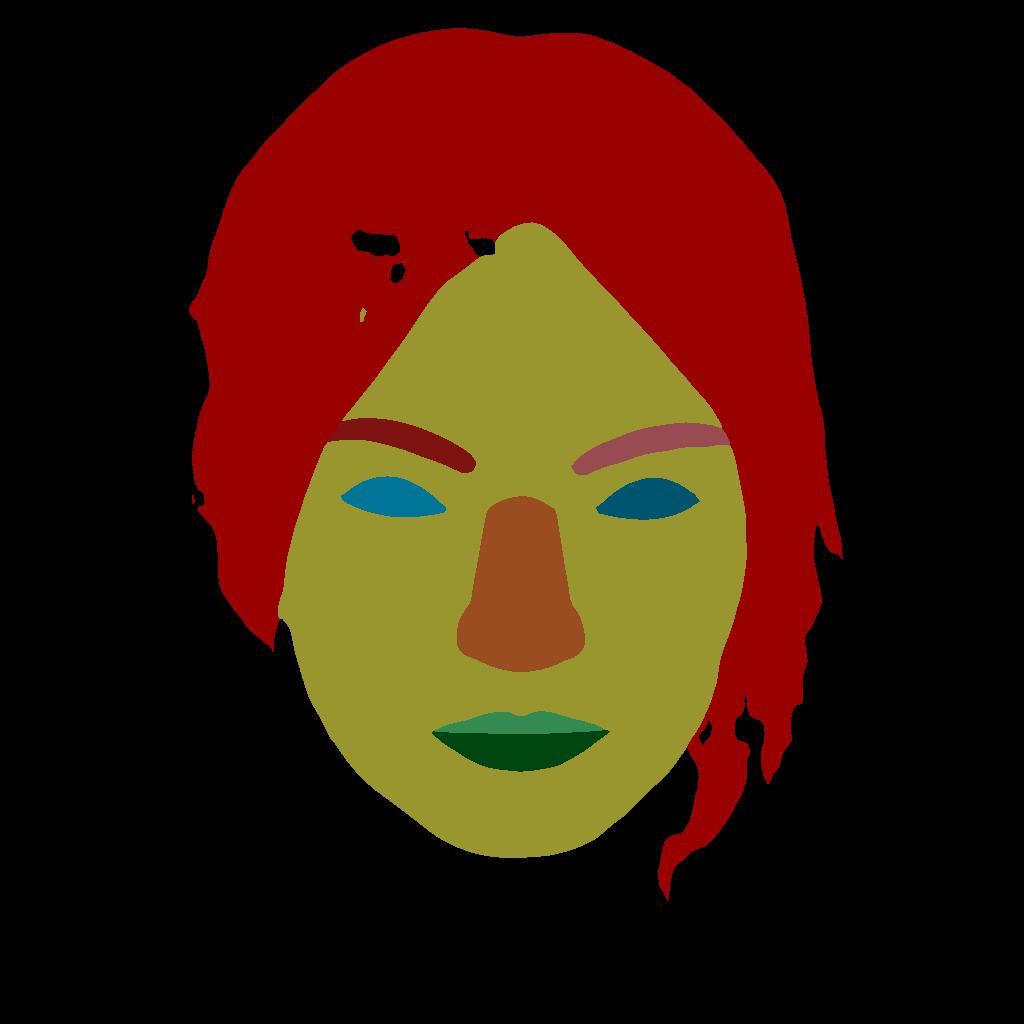}
  \includegraphics[width=0.095\linewidth]{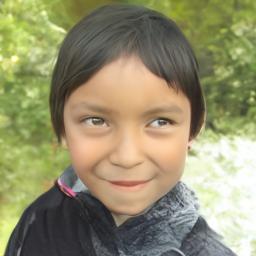}
  \includegraphics[width=0.095\linewidth]{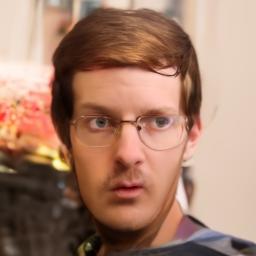}
  \includegraphics[width=0.095\linewidth]{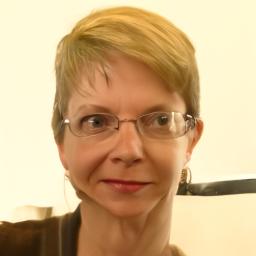}
  \includegraphics[width=0.095\linewidth]{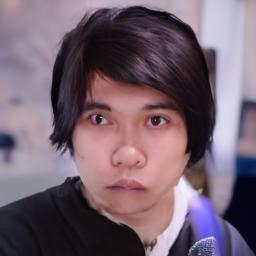}
    \hspace{5mm}
      \includegraphics[width=0.095\linewidth]{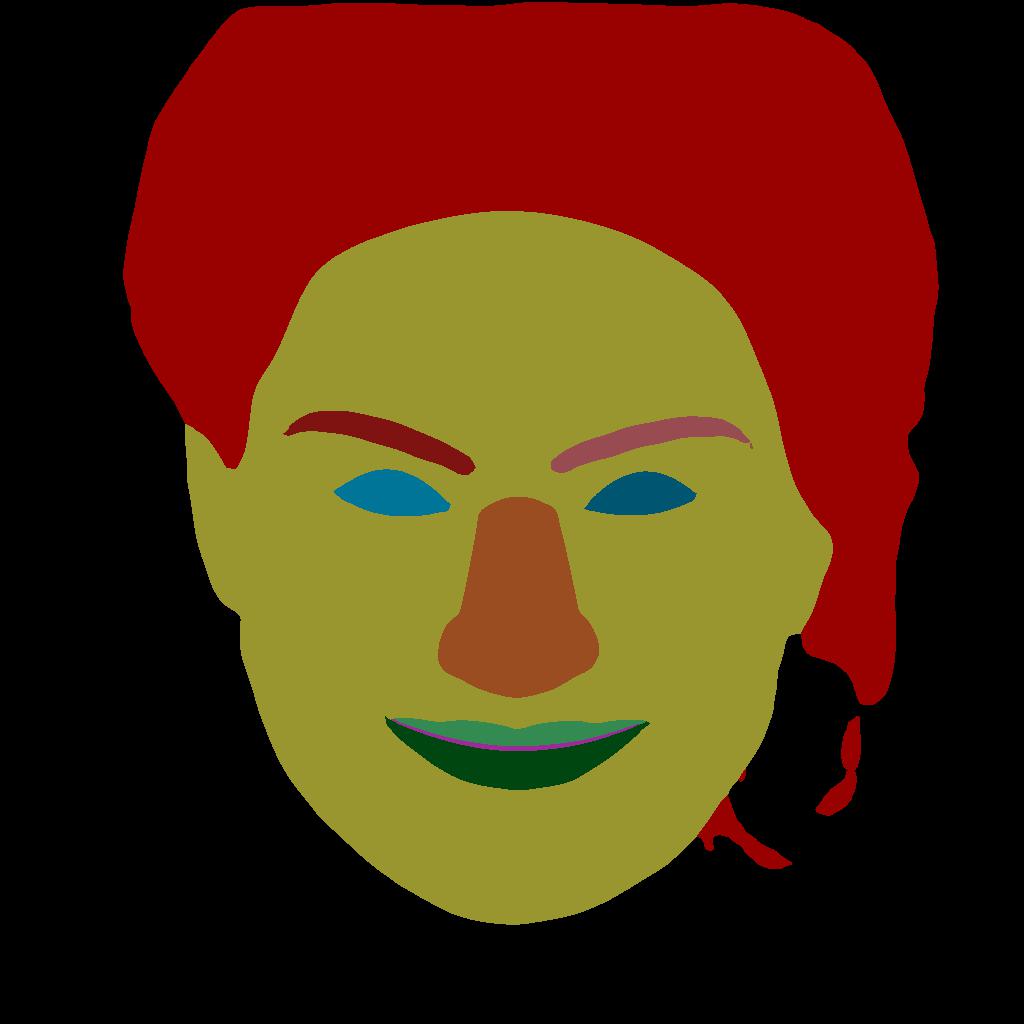}
  \includegraphics[width=0.095\linewidth]{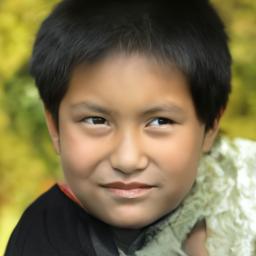}
  \includegraphics[width=0.095\linewidth]{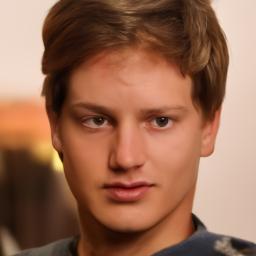}
  \includegraphics[width=0.095\linewidth]{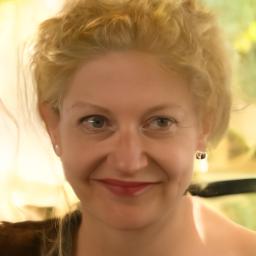}
  \includegraphics[width=0.095\linewidth]{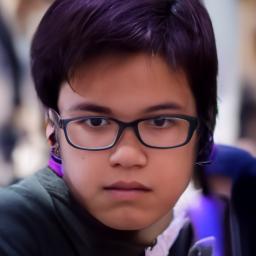}

 \tabularnewline
  \includegraphics[width=0.095\linewidth]{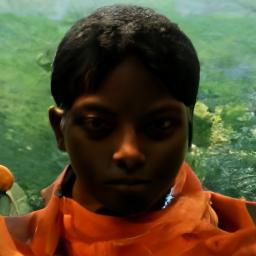}
  \includegraphics[width=0.095\linewidth]{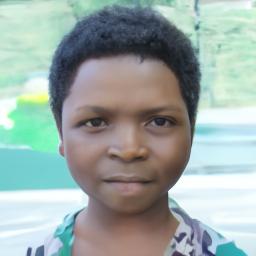}
  \includegraphics[width=0.095\linewidth]{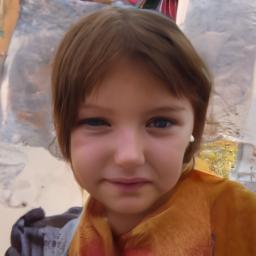}
  \includegraphics[width=0.095\linewidth]{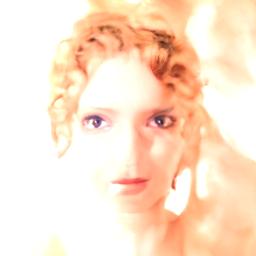}
  \includegraphics[width=0.095\linewidth]{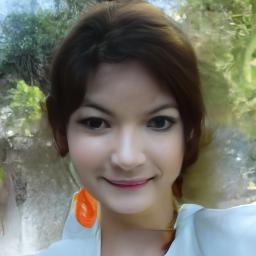}
    \hspace{5mm}
  \includegraphics[width=0.095\linewidth]{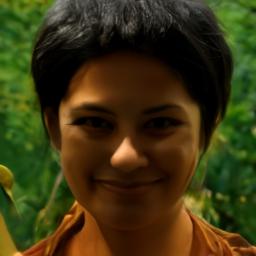}
  \includegraphics[width=0.095\linewidth]{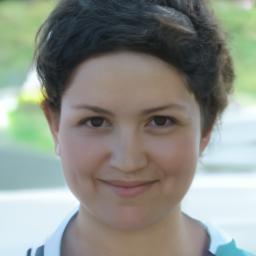}
  \includegraphics[width=0.095\linewidth]{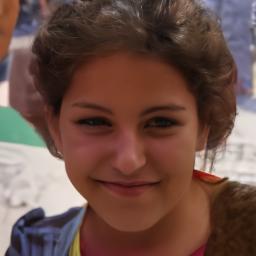}
  \includegraphics[width=0.095\linewidth]{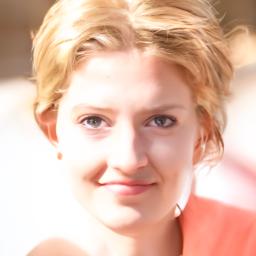}
  \includegraphics[width=0.095\linewidth]{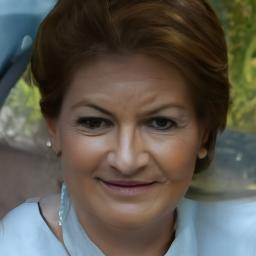}
\tabularnewline
\end{tabular}}
\vspace{-0.8cm}
\hspace{20pt}\captionof{figure}{Non-cherry-picked samples from our method corresponding to the same semantic map. The upper left figure shows the corresponding semantic map. Samples across different examples with same sampling location has same random seeds }
\label{fig:sup2}
\end{figure*}%
\begin{figure*}[tb!]
    \centering
    \setlength{\tabcolsep}{0.5pt}
    {\small
    \renewcommand{\arraystretch}{0.5} 
    \begin{tabular}{c c c c c c c c c c c c }
    \captionsetup{type=figure, font=scriptsize}
  \includegraphics[width=0.08\linewidth]{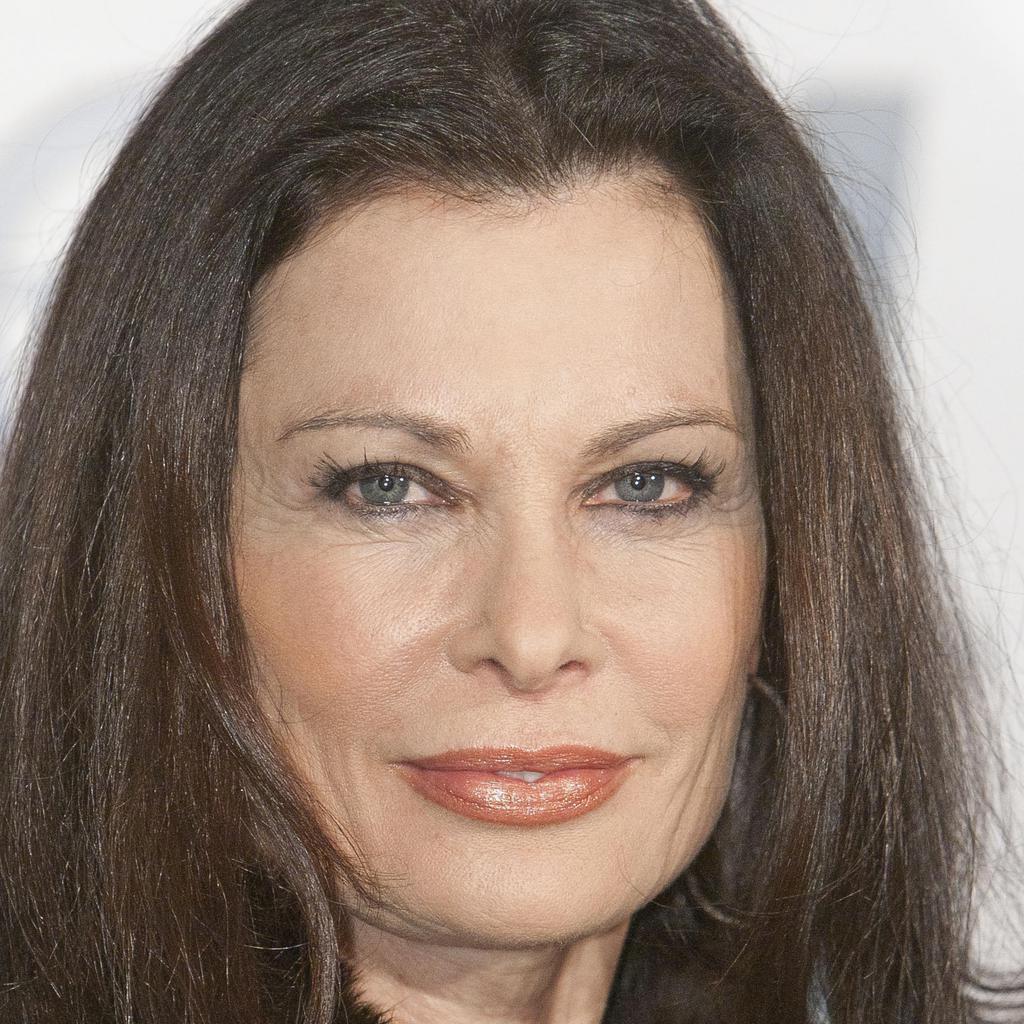}
  \includegraphics[width=0.08\linewidth]{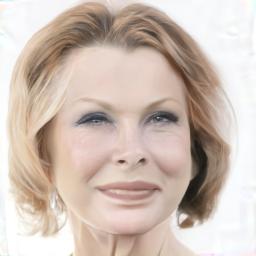}
  \includegraphics[width=0.08\linewidth]{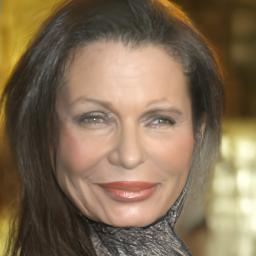}
  \includegraphics[width=0.08\linewidth]{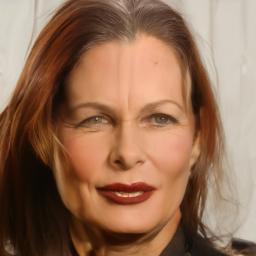}
  \includegraphics[width=0.08\linewidth]{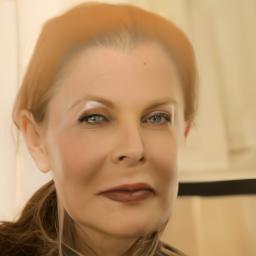}
  \includegraphics[width=0.08\linewidth]{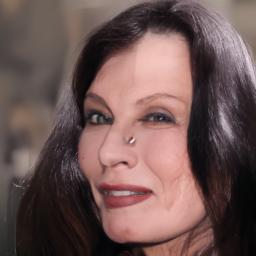}
    \hspace{5mm}
  \includegraphics[width=0.08\linewidth]{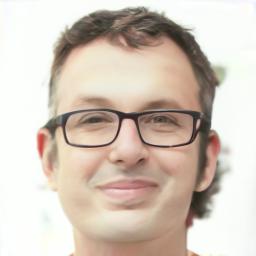}
  \includegraphics[width=0.08\linewidth]{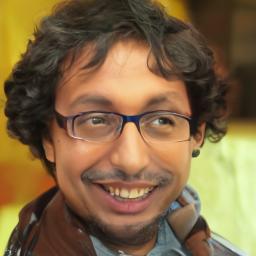}
  \includegraphics[width=0.08\linewidth]{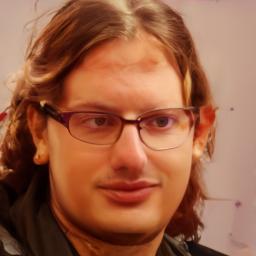}
  \includegraphics[width=0.08\linewidth]{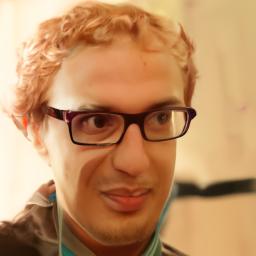}
  \includegraphics[width=0.08\linewidth]{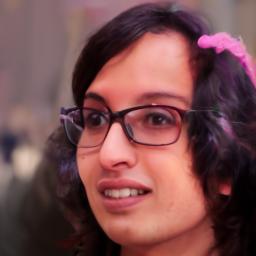}
  \includegraphics[width=0.08\linewidth]{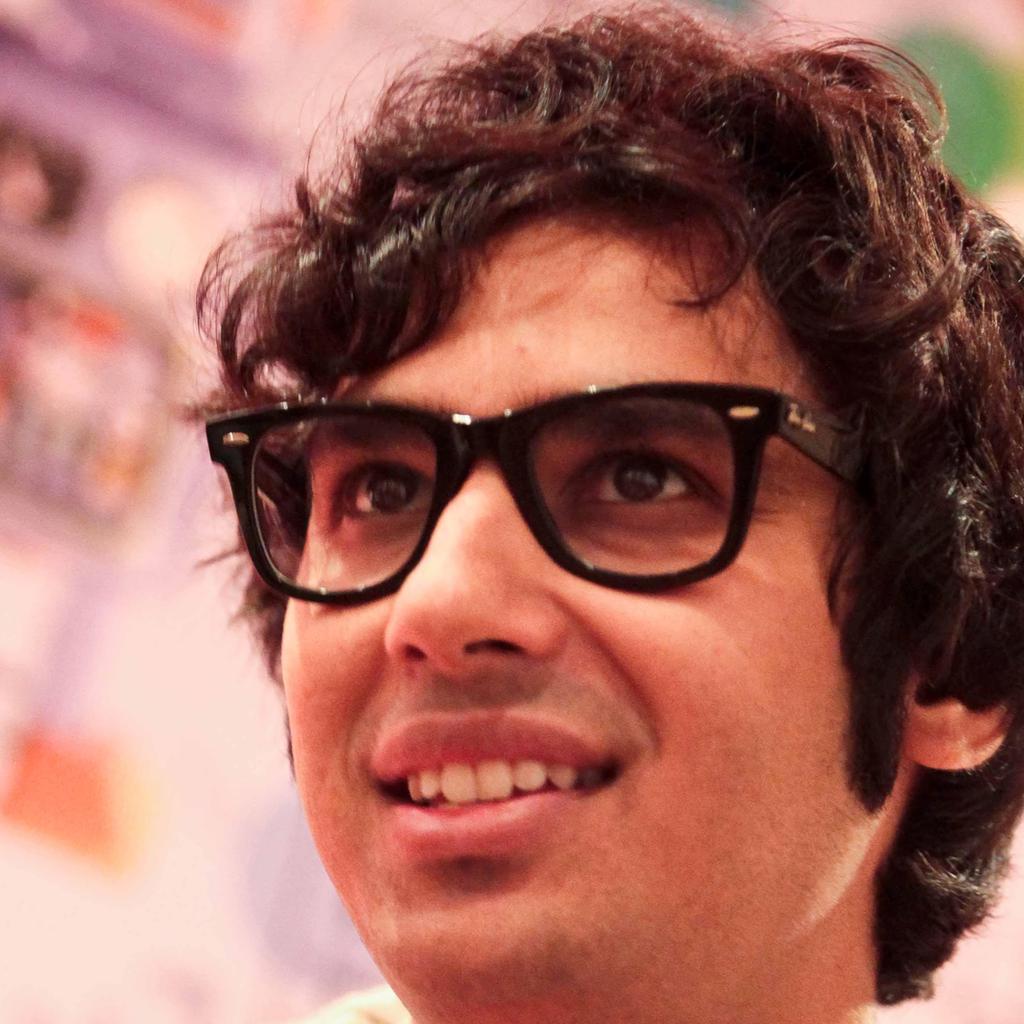}

 \tabularnewline
  \includegraphics[width=0.08\linewidth]{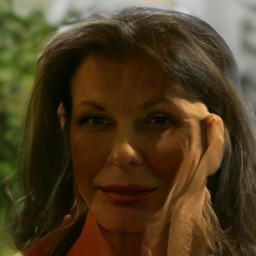}
  \includegraphics[width=0.08\linewidth]{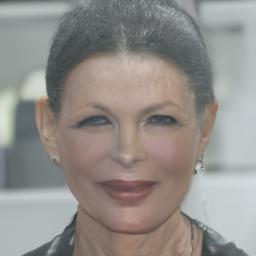}
  \includegraphics[width=0.08\linewidth]{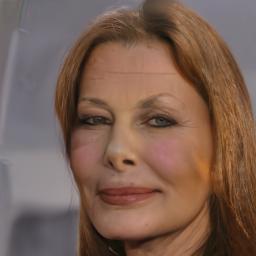}
  \includegraphics[width=0.08\linewidth]{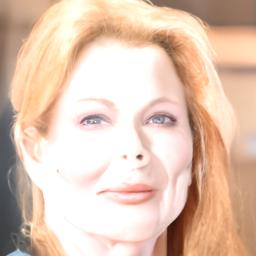}
  \includegraphics[width=0.08\linewidth]{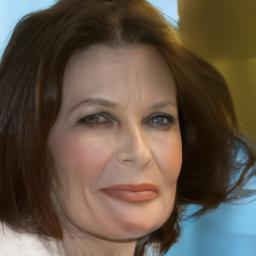}
    \hspace{5mm}
  \includegraphics[width=0.08\linewidth]{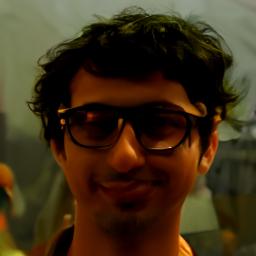}
  \includegraphics[width=0.08\linewidth]{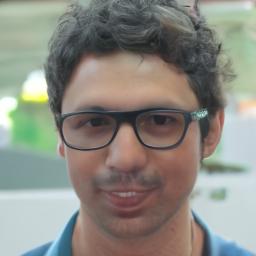}
  \includegraphics[width=0.08\linewidth]{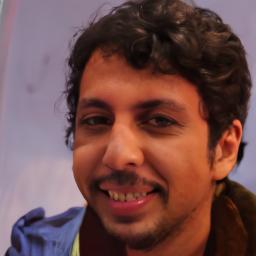}
  \includegraphics[width=0.08\linewidth]{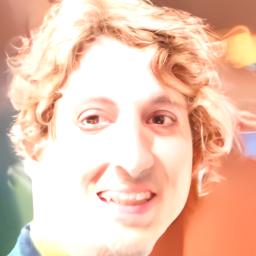}
  \includegraphics[width=0.08\linewidth]{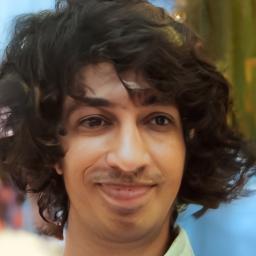}
\tabularnewline
\end{tabular}}
\vspace{-0.8cm}
\label{fig:sup1a}
\vspace{-16mm}
\end{figure*}%
\begin{figure*}[tb!]
    \centering
    \setlength{\tabcolsep}{0.5pt}
    {\small
    \renewcommand{\arraystretch}{0.5} 
    \begin{tabular}{c c c c c c c c c c c c c c }
    \captionsetup{type=figure, font=scriptsize}
  \includegraphics[width=0.08\linewidth]{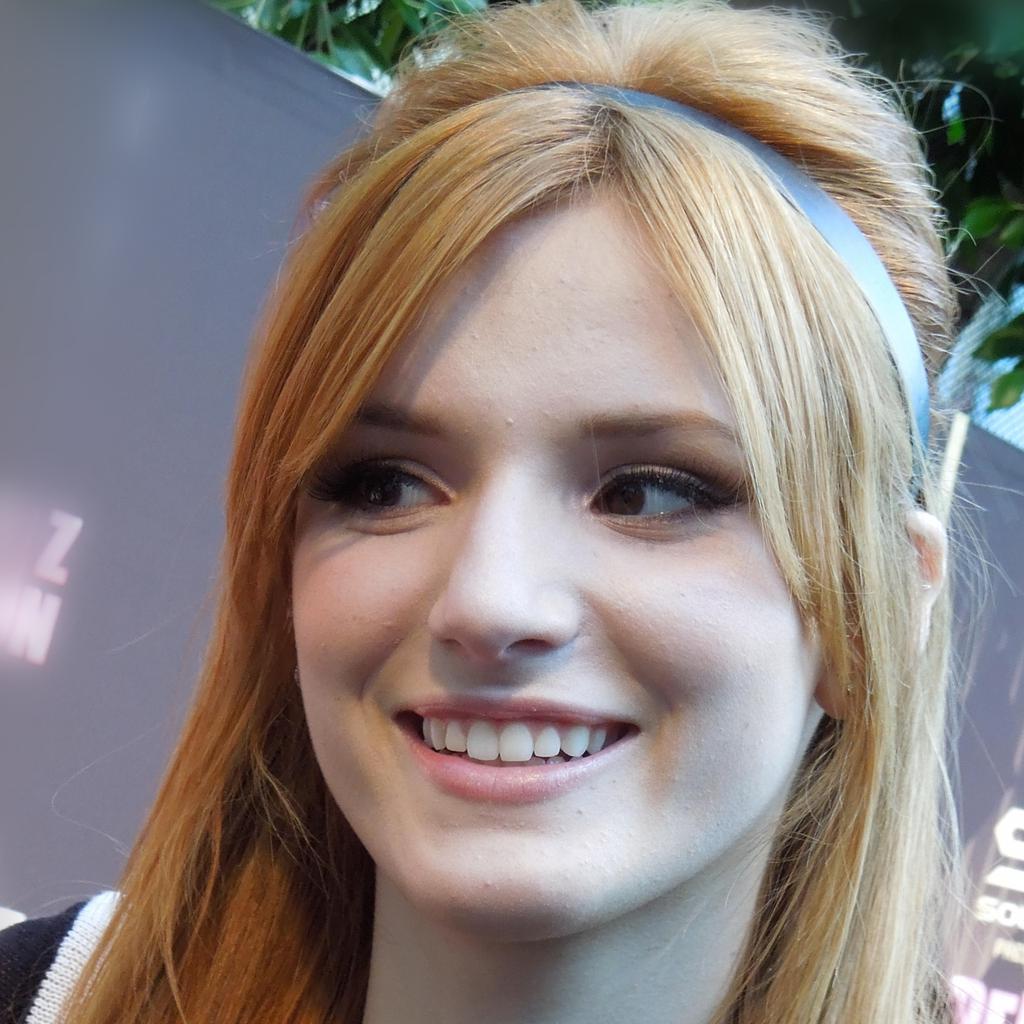}
  \includegraphics[width=0.08\linewidth]{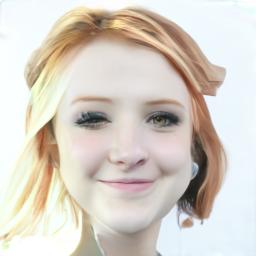}
  \includegraphics[width=0.08\linewidth]{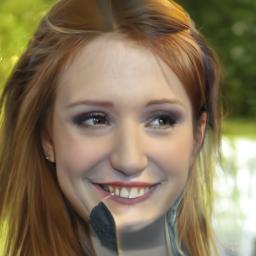}
  \includegraphics[width=0.08\linewidth]{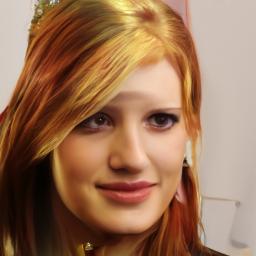}
  \includegraphics[width=0.08\linewidth]{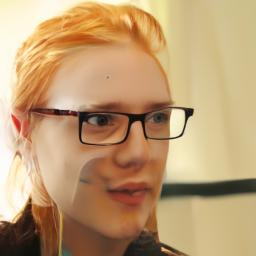}
  \includegraphics[width=0.08\linewidth]{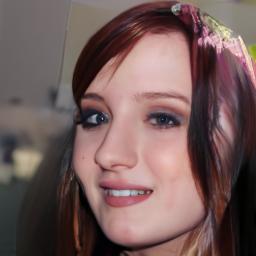}
    \hspace{5mm}
  \includegraphics[width=0.08\linewidth]{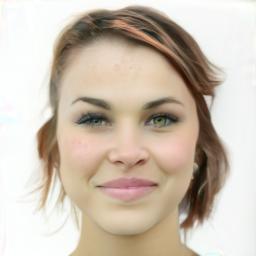}
  \includegraphics[width=0.08\linewidth]{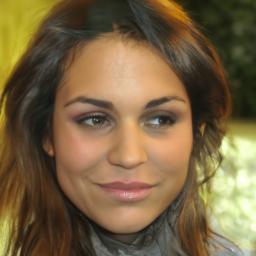}
  \includegraphics[width=0.08\linewidth]{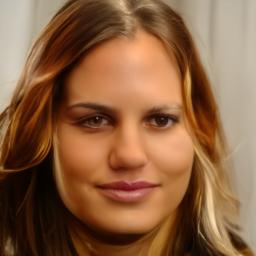}
  \includegraphics[width=0.08\linewidth]{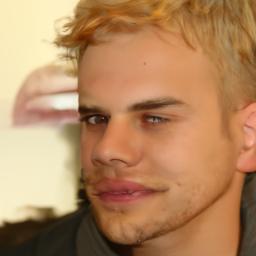}
  \includegraphics[width=0.08\linewidth]{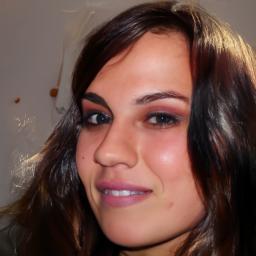}
  \includegraphics[width=0.08\linewidth]{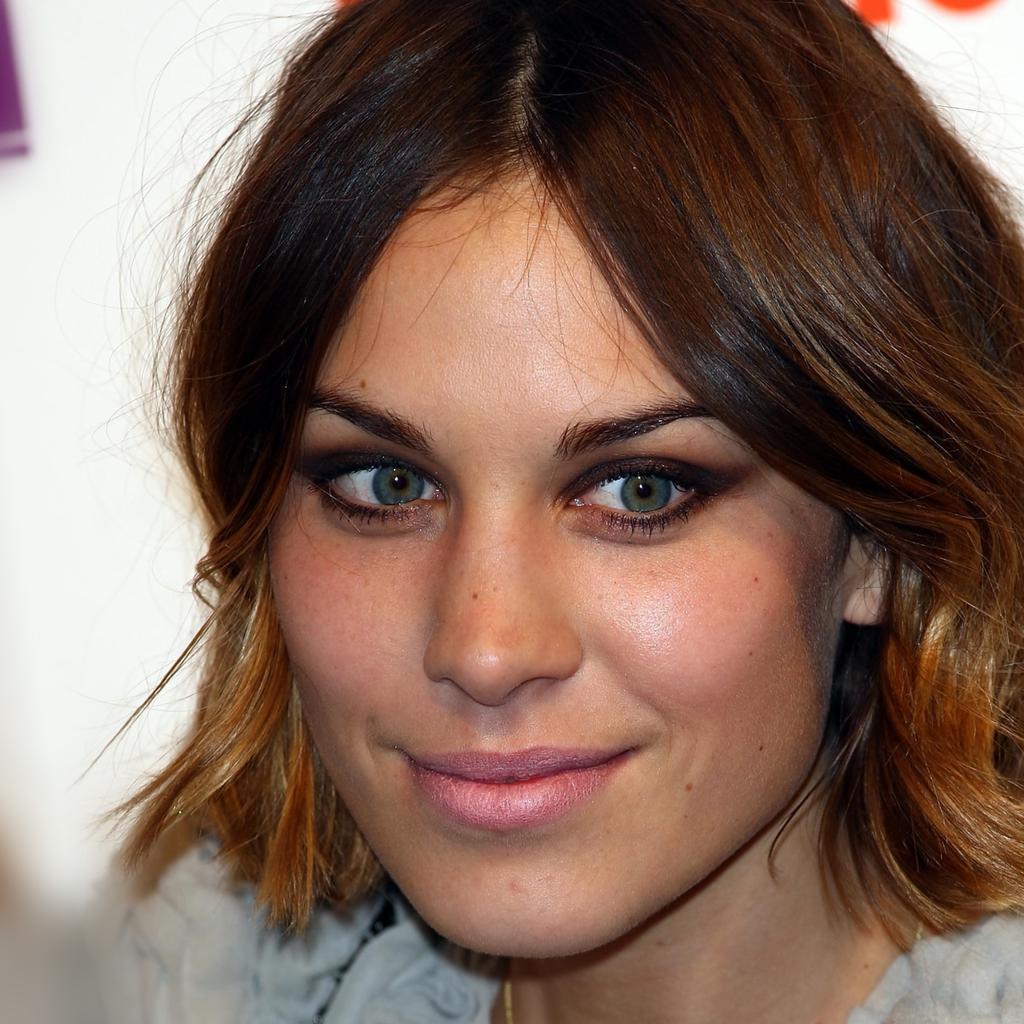}

 \tabularnewline
  \includegraphics[width=0.08\linewidth]{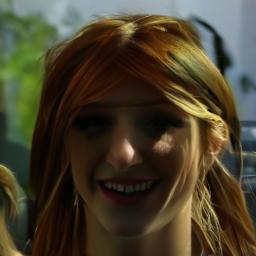}
  \includegraphics[width=0.08\linewidth]{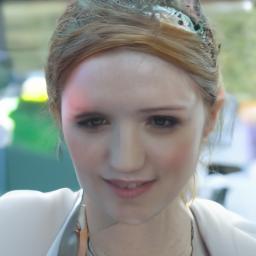}
  \includegraphics[width=0.08\linewidth]{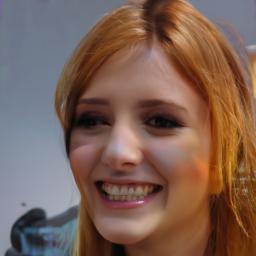}
  \includegraphics[width=0.08\linewidth]{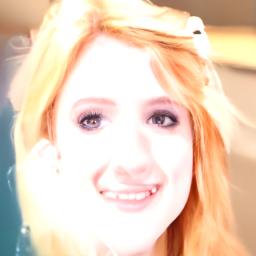}
  \includegraphics[width=0.08\linewidth]{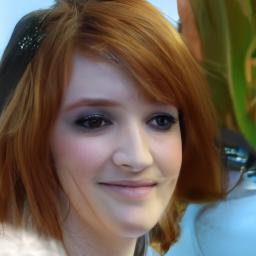}
    \hspace{5mm}
    
  \includegraphics[width=0.08\linewidth]{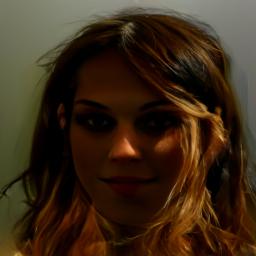}
  \includegraphics[width=0.08\linewidth]{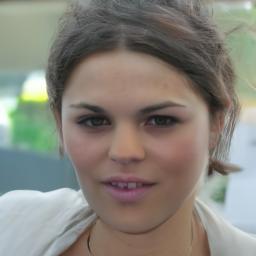}
  \includegraphics[width=0.08\linewidth]{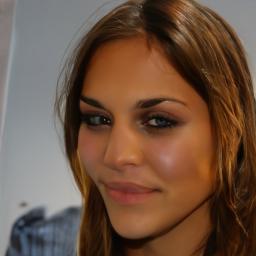}
  \includegraphics[width=0.08\linewidth]{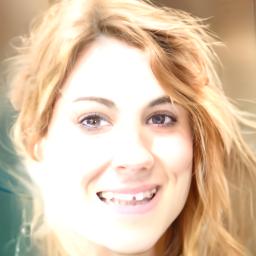}
  \includegraphics[width=0.08\linewidth]{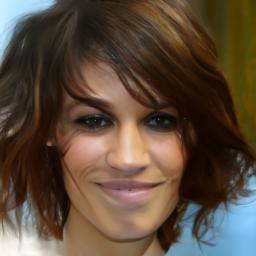}
\tabularnewline
\end{tabular}}
\vspace{-0.8cm}
\hspace{20pt}\captionof{figure}{Non-cherry-picked samples from our method corresponding to the same identity image. The identity image projects out from the rest of the images. Samples across different examples with same sampling location has same random seeds }\label{fig:sup1b}
\vspace{-8mm}
\end{figure*}%

\begin{figure*}[tb!]
    \centering
    \setlength{\tabcolsep}{0.5pt}
    {\small
    \renewcommand{\arraystretch}{0.5} 
    \begin{tabular}{c c c c c c c c c c c c c c }
    \captionsetup{type=figure, font=scriptsize}
  \includegraphics[width=0.08\linewidth]{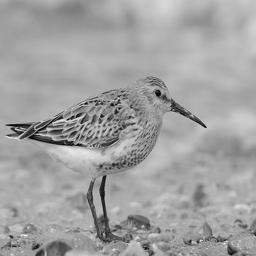}
  \includegraphics[width=0.08\linewidth]{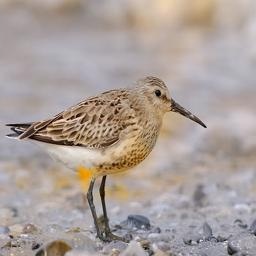}
  \includegraphics[width=0.08\linewidth]{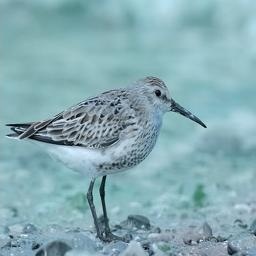}
  \includegraphics[width=0.08\linewidth]{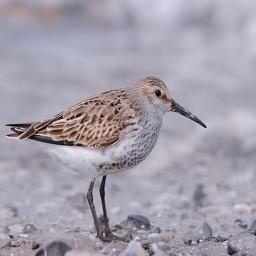}
  \includegraphics[width=0.08\linewidth]{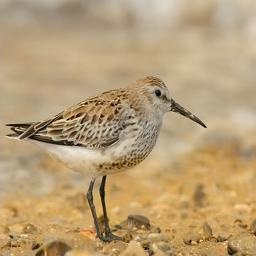}
  \includegraphics[width=0.08\linewidth]{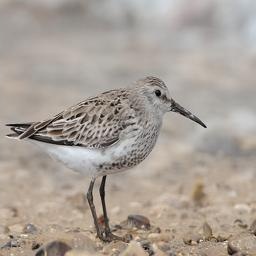}
    \hspace{5mm}
  \includegraphics[width=0.08\linewidth]{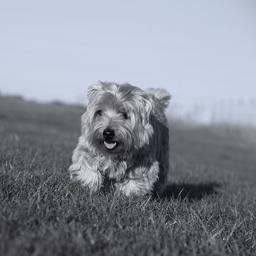}
  \includegraphics[width=0.08\linewidth]{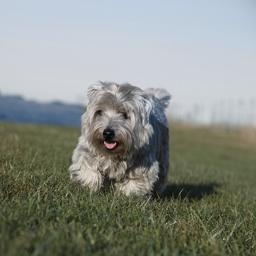}
  \includegraphics[width=0.08\linewidth]{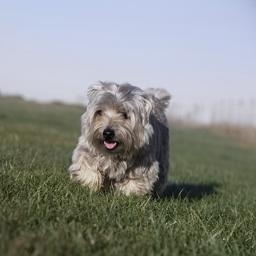}
  \includegraphics[width=0.08\linewidth]{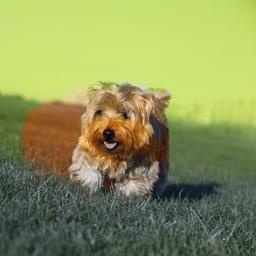}
  \includegraphics[width=0.08\linewidth]{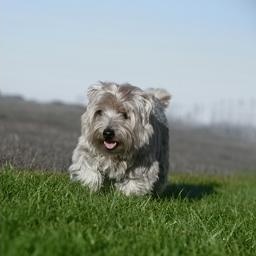}
  \includegraphics[width=0.08\linewidth]{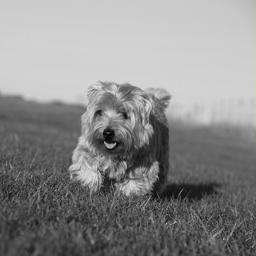}

 \tabularnewline
  \includegraphics[width=0.08\linewidth]{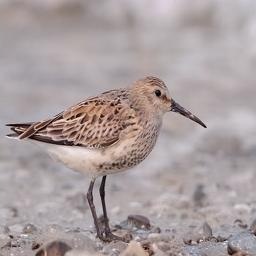}
  \includegraphics[width=0.08\linewidth]{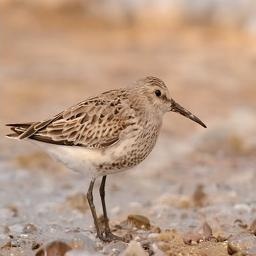}
  \includegraphics[width=0.08\linewidth]{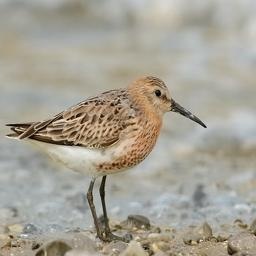}
  \includegraphics[width=0.08\linewidth]{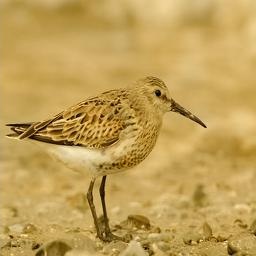}
  \includegraphics[width=0.08\linewidth]{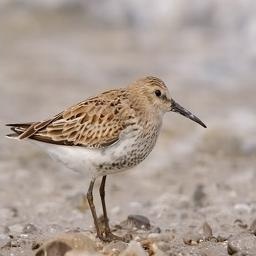}
    \hspace{5mm}
    
  \includegraphics[width=0.08\linewidth]{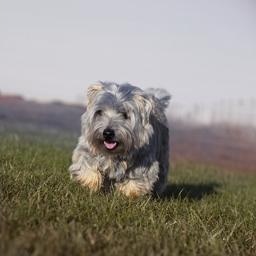}
  \includegraphics[width=0.08\linewidth]{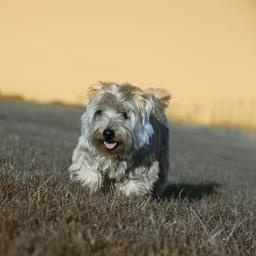}
  \includegraphics[width=0.08\linewidth]{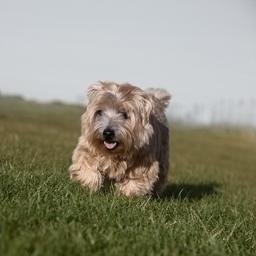}
  \includegraphics[width=0.08\linewidth]{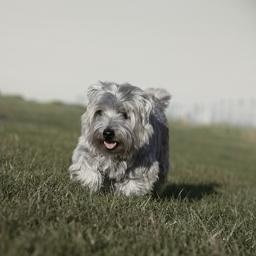}
  \includegraphics[width=0.08\linewidth]{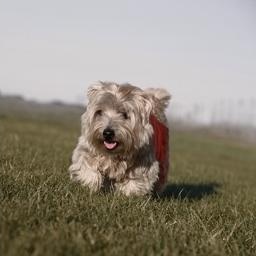}
\tabularnewline
\end{tabular}}
\vspace{-0.8cm}
\hspace{20pt}\captionof{figure}{Non-cherry-picked samples from our method corresponding to grayscale to RGB. The grayscale projects out from the rest of the images. Samples across different examples with same sampling location has same random seeds }\label{fig:sup1b}
\vspace{-8mm}
\end{figure*}%
\begin{figure*}[tb!]
    \centering
    \setlength{\tabcolsep}{0.5pt}
    {\small
    \renewcommand{\arraystretch}{0.5} 
    \begin{tabular}{c c c c c c c c c c c c c c }
    \captionsetup{type=figure, font=scriptsize}
  \includegraphics[width=0.08\linewidth]{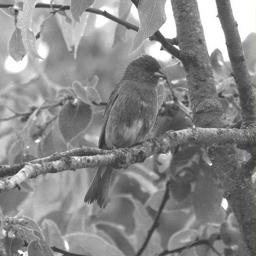}
  \includegraphics[width=0.08\linewidth]{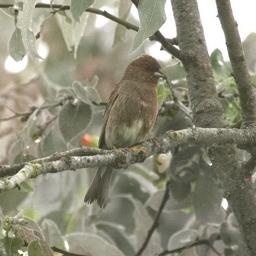}
  \includegraphics[width=0.08\linewidth]{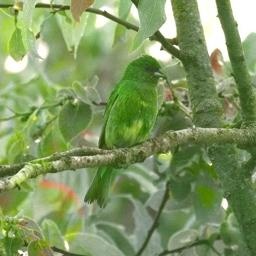}
  \includegraphics[width=0.08\linewidth]{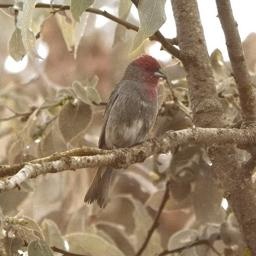}
  \includegraphics[width=0.08\linewidth]{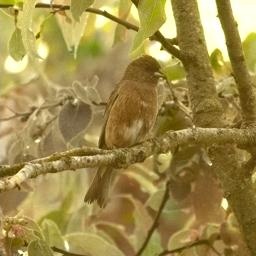}
  \includegraphics[width=0.08\linewidth]{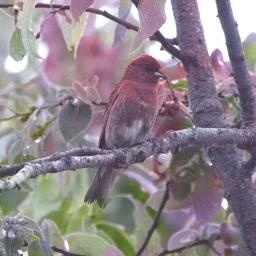}
    \hspace{5mm}
  \includegraphics[width=0.08\linewidth]{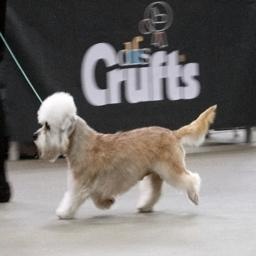}
  \includegraphics[width=0.08\linewidth]{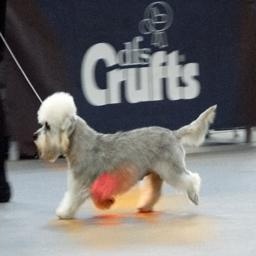}
  \includegraphics[width=0.08\linewidth]{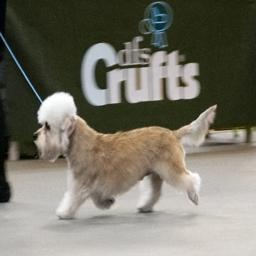}
  \includegraphics[width=0.08\linewidth]{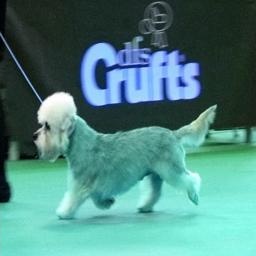}
  \includegraphics[width=0.08\linewidth]{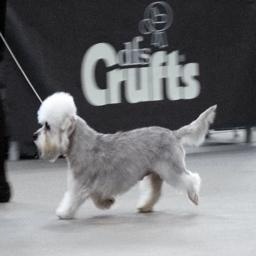}
  \includegraphics[width=0.08\linewidth]{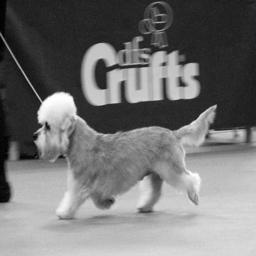}

 \tabularnewline
  \includegraphics[width=0.08\linewidth]{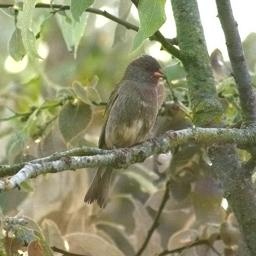}
  \includegraphics[width=0.08\linewidth]{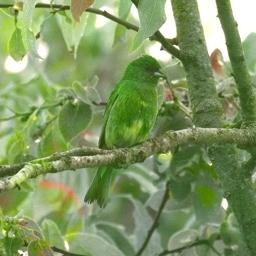}
  \includegraphics[width=0.08\linewidth]{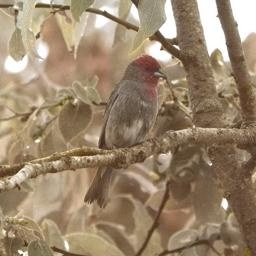}
  \includegraphics[width=0.08\linewidth]{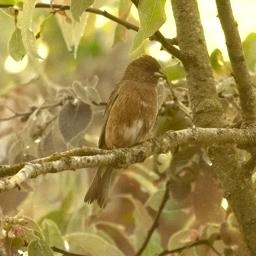}
  \includegraphics[width=0.08\linewidth]{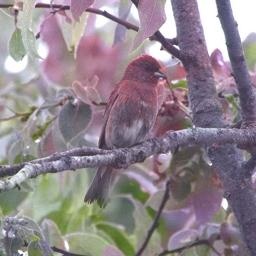}
    \hspace{5mm}
    
  \includegraphics[width=0.08\linewidth]{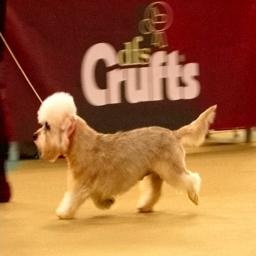}
  \includegraphics[width=0.08\linewidth]{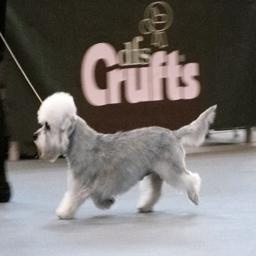}
  \includegraphics[width=0.08\linewidth]{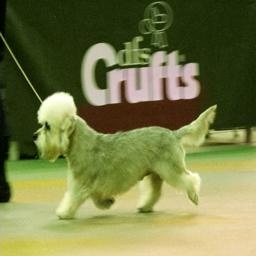}
  \includegraphics[width=0.08\linewidth]{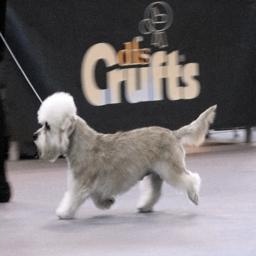}
  \includegraphics[width=0.08\linewidth]{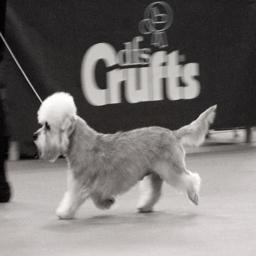}
\tabularnewline
\end{tabular}}
\vspace{-0.8cm}
\hspace{20pt}\captionof{figure}{Non-cherry-picked samples from our method corresponding to grayscale to RGB. The grayscale projects out from the rest of the images. Samples across different examples with same sampling location has same random seeds }\label{fig:sup1b}
\vspace{-8mm}
\end{figure*}%
\begin{figure*}[tb!]
    \centering
    \setlength{\tabcolsep}{0.5pt}
    {\small
    \renewcommand{\arraystretch}{0.5} 
    \begin{tabular}{c c c c c c c c c c c c c c }
    \captionsetup{type=figure, font=scriptsize}
  \includegraphics[width=0.08\linewidth]{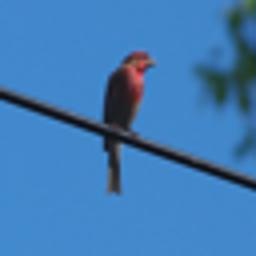}
  \includegraphics[width=0.08\linewidth]{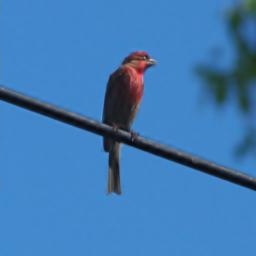}
  \includegraphics[width=0.08\linewidth]{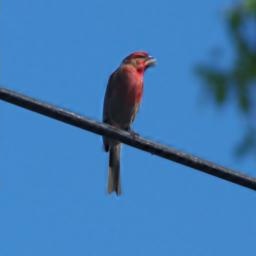}
  \includegraphics[width=0.08\linewidth]{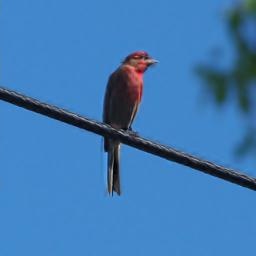}
  \includegraphics[width=0.08\linewidth]{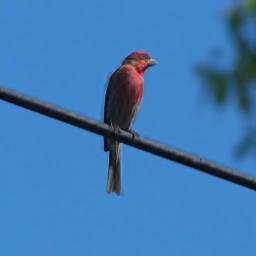}
  \includegraphics[width=0.08\linewidth]{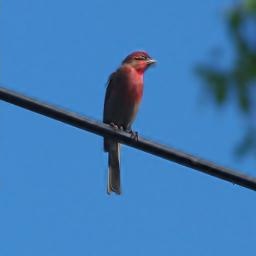}
    \hspace{5mm}
  \includegraphics[width=0.08\linewidth]{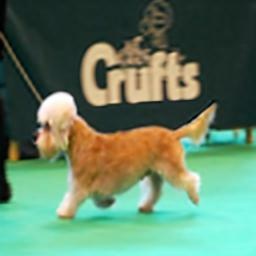}
  \includegraphics[width=0.08\linewidth]{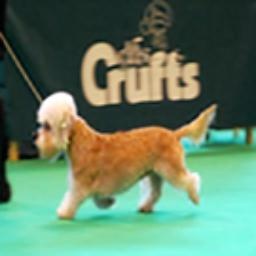}
  \includegraphics[width=0.08\linewidth]{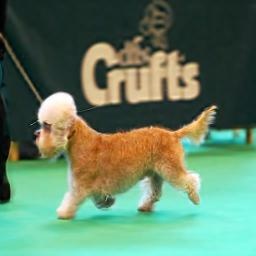}
  \includegraphics[width=0.08\linewidth]{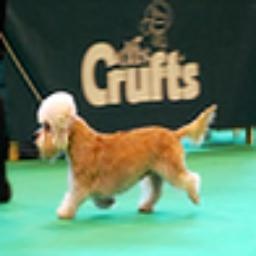}
  \includegraphics[width=0.08\linewidth]{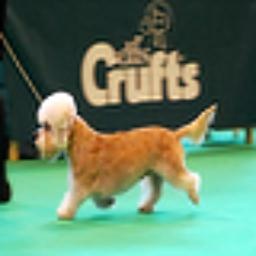}
  \includegraphics[width=0.08\linewidth]{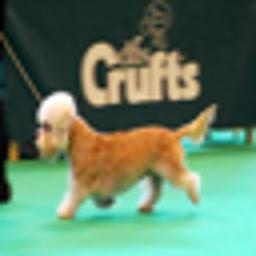}

 \tabularnewline
  \includegraphics[width=0.08\linewidth]{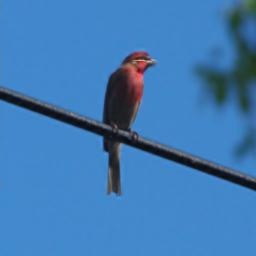}
  \includegraphics[width=0.08\linewidth]{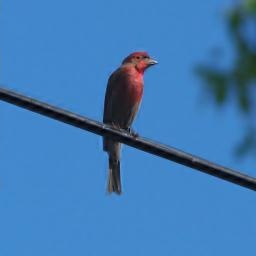}
  \includegraphics[width=0.08\linewidth]{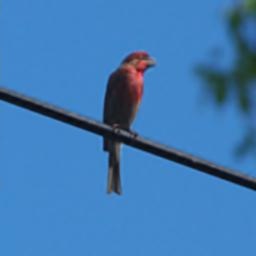}
  \includegraphics[width=0.08\linewidth]{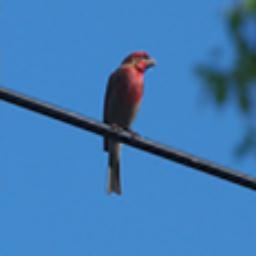}
  \includegraphics[width=0.08\linewidth]{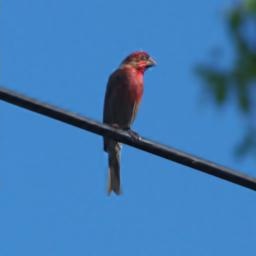}
    \hspace{5mm}
    
  \includegraphics[width=0.08\linewidth]{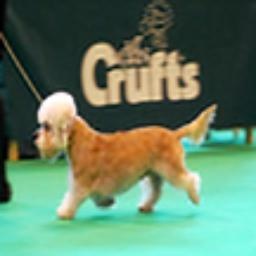}
  \includegraphics[width=0.08\linewidth]{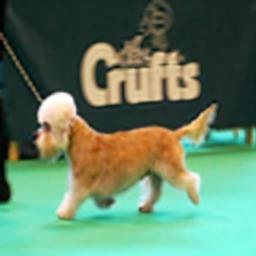}
  \includegraphics[width=0.08\linewidth]{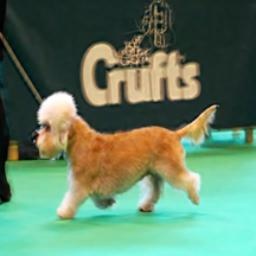}
  \includegraphics[width=0.08\linewidth]{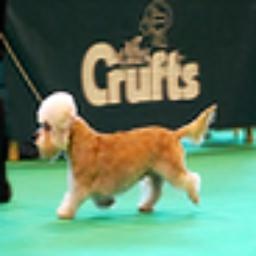}
  \includegraphics[width=0.08\linewidth]{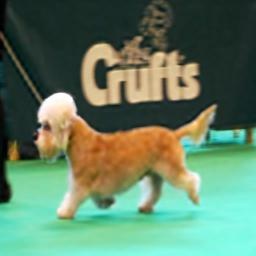}
\tabularnewline
\end{tabular}}
\vspace{-0.8cm}
\hspace{20pt}\captionof{figure}{Non-cherry-picked samples from our method corresponding to LR to SR. The grayscale projects out from the rest of the images. Samples across different examples with same sampling location has same random seeds }\label{fig:sup1b}
\vspace{-8mm}
\end{figure*}%
\begin{figure*}[tb!]
    \centering
    \setlength{\tabcolsep}{0.5pt}
    {\small
    \renewcommand{\arraystretch}{0.5} 
    \begin{tabular}{c c c c c c c c c c c c c c }
    \captionsetup{type=figure, font=scriptsize}
  \includegraphics[width=0.08\linewidth]{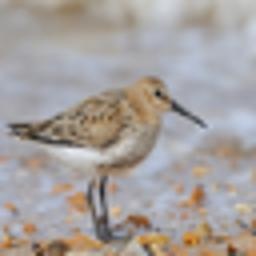}
  \includegraphics[width=0.08\linewidth]{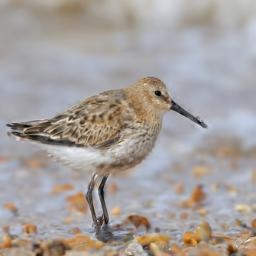}
  \includegraphics[width=0.08\linewidth]{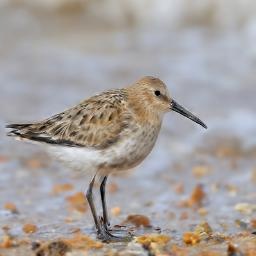}
  \includegraphics[width=0.08\linewidth]{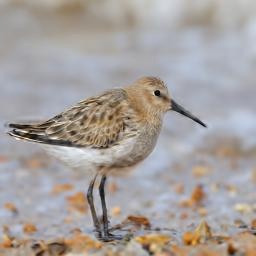}
  \includegraphics[width=0.08\linewidth]{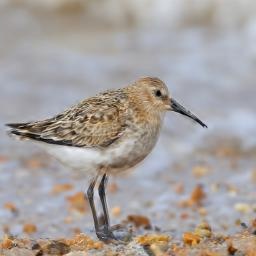}
  \includegraphics[width=0.08\linewidth]{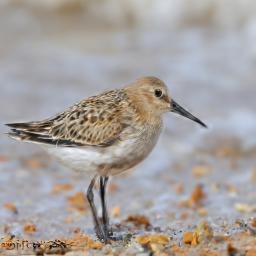}
    \hspace{5mm}
  \includegraphics[width=0.08\linewidth]{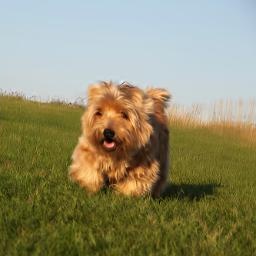}
  \includegraphics[width=0.08\linewidth]{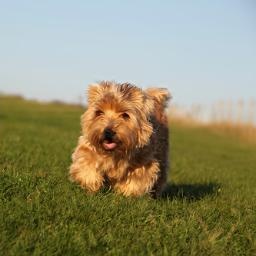}
  \includegraphics[width=0.08\linewidth]{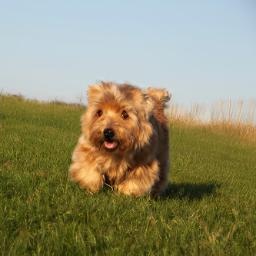}
  \includegraphics[width=0.08\linewidth]{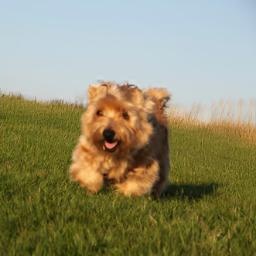}
  \includegraphics[width=0.08\linewidth]{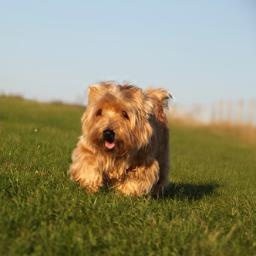}
  \includegraphics[width=0.08\linewidth]{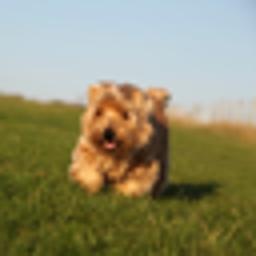}

 \tabularnewline
  \includegraphics[width=0.08\linewidth]{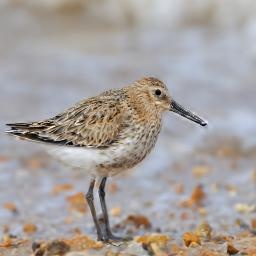}
  \includegraphics[width=0.08\linewidth]{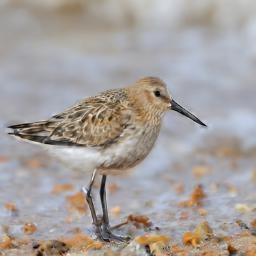}
  \includegraphics[width=0.08\linewidth]{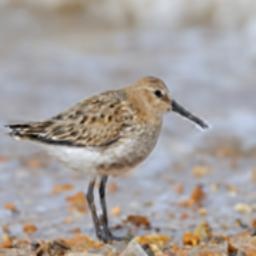}
  \includegraphics[width=0.08\linewidth]{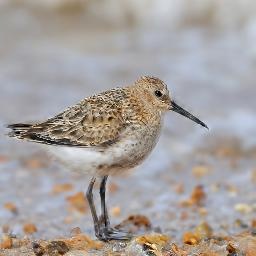}
  \includegraphics[width=0.08\linewidth]{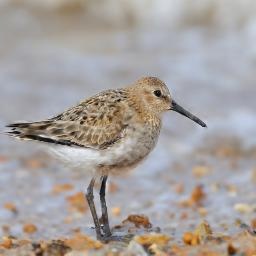}
    \hspace{5mm}
    
  \includegraphics[width=0.08\linewidth]{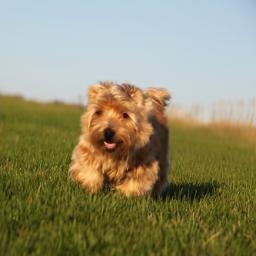}
  \includegraphics[width=0.08\linewidth]{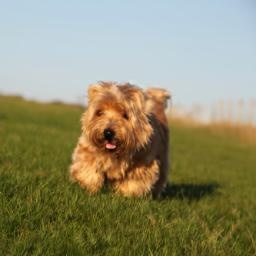}
  \includegraphics[width=0.08\linewidth]{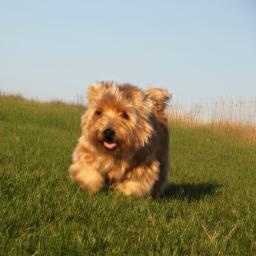}
  \includegraphics[width=0.08\linewidth]{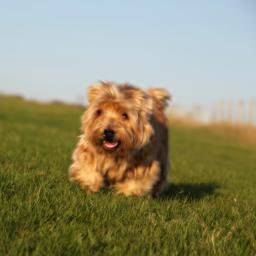}
  \includegraphics[width=0.08\linewidth]{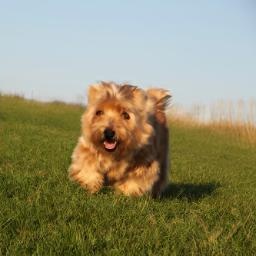}
\tabularnewline
\end{tabular}}
\vspace{-0.8cm}
\hspace{20pt}\captionof{figure}{Non-cherry-picked samples from our method corresponding to LR to SR. The grayscale projects out from the rest of the images. Samples across different examples with same sampling location has same random seeds }\label{fig:sup1b}
\vspace{-8mm}
\end{figure*}%
\end{appendices}

\end{document}